\documentclass{svproc}

\usepackage[utf8]{inputenc}
\usepackage[english]{babel}

\usepackage{url}

\usepackage{latexsym}
\usepackage{amssymb}
\usepackage{amsmath}

\usepackage{amsthm}

\usepackage{booktabs}
\usepackage{enumitem}
\usepackage{graphicx}
\usepackage{color}

\usepackage{algorithm}
\usepackage{algpseudocode}

\usepackage{times}
\usepackage{epsfig}
\usepackage{amssymb}
\usepackage{tikz}
\usepackage{comment}
\usepackage{color}
\usepackage{subcaption}
\usepackage{float}
\usepackage{placeins}
\usepackage{pdflscape}

\newcommand{\BibTeX}{B\kern-.05em{\sc i\kern-.025em b}\kern-.08em\TeX}

\begin{document}
\begin{frontmatter}
\mainmatter              
\title{Leveraging Lightweight Generators for Memory Efficient Continual Learning}
\titlerunning{Leveraging Lightweight Generators for Memory Efficient Continual Learning}  
%

\author{Christiaan Lamers \inst{1} \and Ahmed Nabil Belbachir \inst{1} \and Thomas Bäck \inst{2} \and Niki van Stein \inst{2}}
\authorrunning{Christiaan Lamers et al.} 

\tocauthor{Christiaan Lamers, Ahmed Nabil Belbachir, Thomas Bäck and Niki van Stein}

\institute{NORCE Norwegian Research Centre, Jon Lilletuns vei 9 H, 3. et,\\
4879 Grimstad, Norway,\\
\email{chla@norceresearch.no}
\and
LIACS, Leiden University, Einsteinweg 55, 2333 CC Leiden, The Netherlands 
}

\maketitle              

\begin{abstract}
Catastrophic forgetting can be trivially alleviated by keeping all data from previous tasks in memory. Therefore, minimizing the memory footprint while maximizing the amount of relevant information is crucial to the challenge of \textit{continual learning}. This paper aims to decrease required memory for \textit{memory-based continuous learning algorithms}. We explore the options of extracting a minimal amount of information, while maximally alleviating forgetting. We propose the usage of \textit{lightweight generators} based on Singular Value Decomposition to enhance existing continual learning methods, such as A-GEM and Experience Replay. These generators need a minimal amount of memory while being maximally effective. They require no training time, just a single linear-time fitting step, and can capture a distribution effectively from a small number of data samples. Depending on the dataset and network architecture, our results show a significant increase in average accuracy compared to the original methods. Our method shows great potential in minimizing the memory footprint of memory-based continual learning algorithms.
\end{abstract}


\end{frontmatter}


\section{Introduction}
Continual learning refers to the act of a learner being able to accumulate knowledge over a continual stream of experiences \cite{Thrun1998,parisi2019continual}. Such a learning process allows the learner to refine its abilities over time and integrate skills from different experiences. In practice this means that a learner should be able to learn from a continuous stream of data with a shifting distribution.\\
One of the main challenges in continual learning is overcoming \textit{catastrophic forgetting}. This refers to the declining performance on previous tasks when multiple tasks are presented to the learner in sequence.\\
Van de Ven et al. \cite{van2019three} mention three main scenarios in continual learning: task-incremental learning (Task-IL), domain incremental learning (Domain-IL) and class incremental learning (Class-IL). In both Task-IL and Domain-IL, every task has the same class labels, where in Class-IL, the class labels can change at every task. Task-IL and Domain-IL differ in the availability of task labels during test time. In Task-IL, task labels are available during test time, while in Domain-IL, they are not. Of these three scenarios, Task-IL and Domain-IL are most similar, and are the best fit for the methods we would like to propose. We consider Domain-IL to be the hardest challenge and thus the most interesting, therefore our experiments focus on Domain-IL.\\
The problem of catastrophic forgetting can be trivially solved by keeping all data from previous tasks in memory and simply training a neural network on this entire corpus of data. In practice, this is not feasible; thus selecting what to store and what not to store is a core challenge in memory-based continual learning algorithms. This challenge can be extended to the aim of using as little memory as possible, while maximizing the alleviation of catastrophic forgetting.  Besides this, Moore's Law is ending \cite{theis2017end}, so efficiency both in memory and computational cost is required to advance the state-of-the-art. Our aim is to minimize the amount of required memory by extracting only information on previous tasks that maximally alleviates forgetting.

\medskip

\textbf{Main contributions:}
A Singular Value Decomposition (SVD) \cite{golub1971singular} based method is developed that minimizes the footprint of continual learning algorithms. The benefits have been evaluated of storing abstractions of the data rather than storing the raw data for continual learning tasks. 

\medskip

The paper is organized as follows: \\
\textbf{Section 2} contains a \textit{literature review}, where the \textit{five trends} in continual learning are briefly discussed, with a focus on \textit{rehearsal-based approaches} and the use of \textit{generators}.\\
\textbf{Section 3} explains the \textit{methodology}. It elaborates on \textit{Domain Incremental Learning}, the mechanisms of the proposed \textit{Lightweight Generators} and how they are \textit{incorporated} in rehearsal-based algorithms in a memory efficient way.\\
\textbf{Section 4} demonstrates the \textit{effectiveness} of the proposed method through various experimental \textit{results}.\\
\textbf{Section 5} contains the \textit{conclusion} and \textit{future work}.

\section{Literature Review}
Parisi et al. \cite{parisi2019continual} describe three main trends in the alleviation of catastrophic forgetting: regularization, rehearsal and architectural strategies. Wang et al \cite{wang2024comprehensive} extend this list with two more trends: optimization-based and representation-based strategies.

\subsection{Five Trends}
\textit{Regularization} focuses on restricting changes to weight values in such a way that older tasks are not forgotten. \textit{Rehearsal} alleviates forgetting by mixing in samples of previous tasks with the current task. \textit{Architectural} strategies change the network's architecture during training time, freezing certain parts in order to not forget and adding new nodes in order to keep the ability to learn. The \textit{optimization-based} approach focuses on altering the optimization algorithm to negate catastrophic forgetting. The \textit{representation-based} approach focuses on representing the essence of the data in a efficient way through (semi-)supervised learning or meta learning. Rao et al. \cite{rao2019continual} represents tasks as clusters of datapoints, together with inter- and intra-cluster variation.

\subsection{Rehearsal}
In the \textit{rehearsal} approach, samples from previous tasks need to be kept in memory. Chaudry et al. \cite{chaudhry2019tiny} describe multiple strategies to accommodate this. \textit{Reservoir sampling} decides on adding samples to the memory by random chance. The \textit{ring buffer} strategy relies on the First In First Out principle. The \textit{K-Means} strategy decides on sample addition by measuring distances from centroids. The \textit{Mean of Features} strategy acts in a similar fashion, but measures distances from the mode in feature space. Lamers et al. \cite{Lamers_2023_ICCV} explore the possibilities of applying K-means on the input space as well as the weight space to select suitable samples, with the added challenge to do so in a task-agnostic setting.

\subsection{Generators}
Another approach for the \textit{rehearsal} methods is to generate synthetic samples. A multitude of methods exist that fall in this category named \textit{generative replay} \cite{shin2017continual,van2018generative,stoianov2022hippocampal,lesort2019generative}. These methods use a generator to synthesize data from previous tasks, which is then mixed with data from the current task and presented to the learner.\\
Van de Ven et al. \cite{van2020brain} train a variational autoencoder that is used to generate samples. While such a generator is well fitted to capture non-linear latent spaces, the amount of required memory is not as minimal as our approach. Another downside of variational autoencoders is that they need to be trained on a substantial amount of data, which requires a significant amount of time and resources due to the iterative application of back propagation.\\
Nagata et al. mention that the use of generators makes the learner more robust against overfitting versus storing raw samples \cite{nagata2024reducing}.\\
Alternatively to storing data that a learner would receive as input, it is also possible to store vectors in the weight space of the network. Orthogonal Gradient Descent (OGD) \cite{pmlr-v108-farajtabar20a} stores gradients of outputs of a neural network and projects gradient descent steps orthogonal to them. Saha et al. \cite{saha2021gradient} apply a similar strategy, but they aim to find structures in the weight space, by calculating base vectors for a subspace of most forgetting (thereby being able to avoid forgetting). They achieve this by applying Singular Value Decomposition (SVD) on network activations, after learning a task.\\
Our work focuses on Averaged Gradient Episodic Memory (A-GEM) \cite{chaudhry2018efficient} and Experience Replay (ER) \cite{rolnick2019experience,buzzega2021rethinking}, because these methods are relatively fast and their memory requirements are low. A-GEM is a combination of regularization and rehearsal. A-GEM does not directly regularize weights in the network, it does restrict gradient descent steps by means of projection. A-GEM can be considered to be a rehearsal based method, because it keeps a memory of samples from previous tasks, just like ER. A-GEM uses the samples for rehearsal in an indirect manner, through calculating directions of most forgetting in the weight space.\\
Both A-GEM and ER require a pool of samples. In our method we generate the samples, thus our method falls in the category of \textit{generative replay}. The novelty of our method lies in the \textit{reduced memory requirements} of our generators as well as a \textit{training time} that is negligibly short.

\section{Methodology}
To reiterate, catastrophic forgetting can be alleviated in a trivial fashion, by simply storing the entire stream of data and present this to a learner once all data is gathered. However, in practice this would result in unfeasible memory requirements. That does not mean that it is completely forbidden to store data for later use, but an effort should be made to minimize the amount of memory used. Rehearsal based methods like A-GEM and ER manage to successfully alleviate catastrophic forgetting by storing only a small subset of the entire data stream. Our aim is to reduce the amount of stored information further by retaining valuable information per \textit{task} and \textit{class} and discarding the less useful data.

\subsection{Problem Definition}
We focus on Domain-IL , where $m$ tasks ($T_0, T_1, ..., T_{m-1}$) are presented one after another. Each task has the same $c$ class labels $[0,c-1]$. The goal is to minimize the loss on task $T_k$ for all $k \in [0,m-1]$, while at the same minimizing the increase in loss on all tasks $T_l$, where $l \in [0,m-2]$, such that $l < k$. In other words, the goal is to learn a new task, while minimizing the forgetting on previous tasks. After training on task $T_k$ is finished, our memory-based algorithm takes $s$ randomly selected samples of task $T_k$ and stores them in a memory buffer. 

\subsection{Lightweight Generators}
\begin{figure}[!htbp]
\centering
\includegraphics[width=0.9\linewidth]{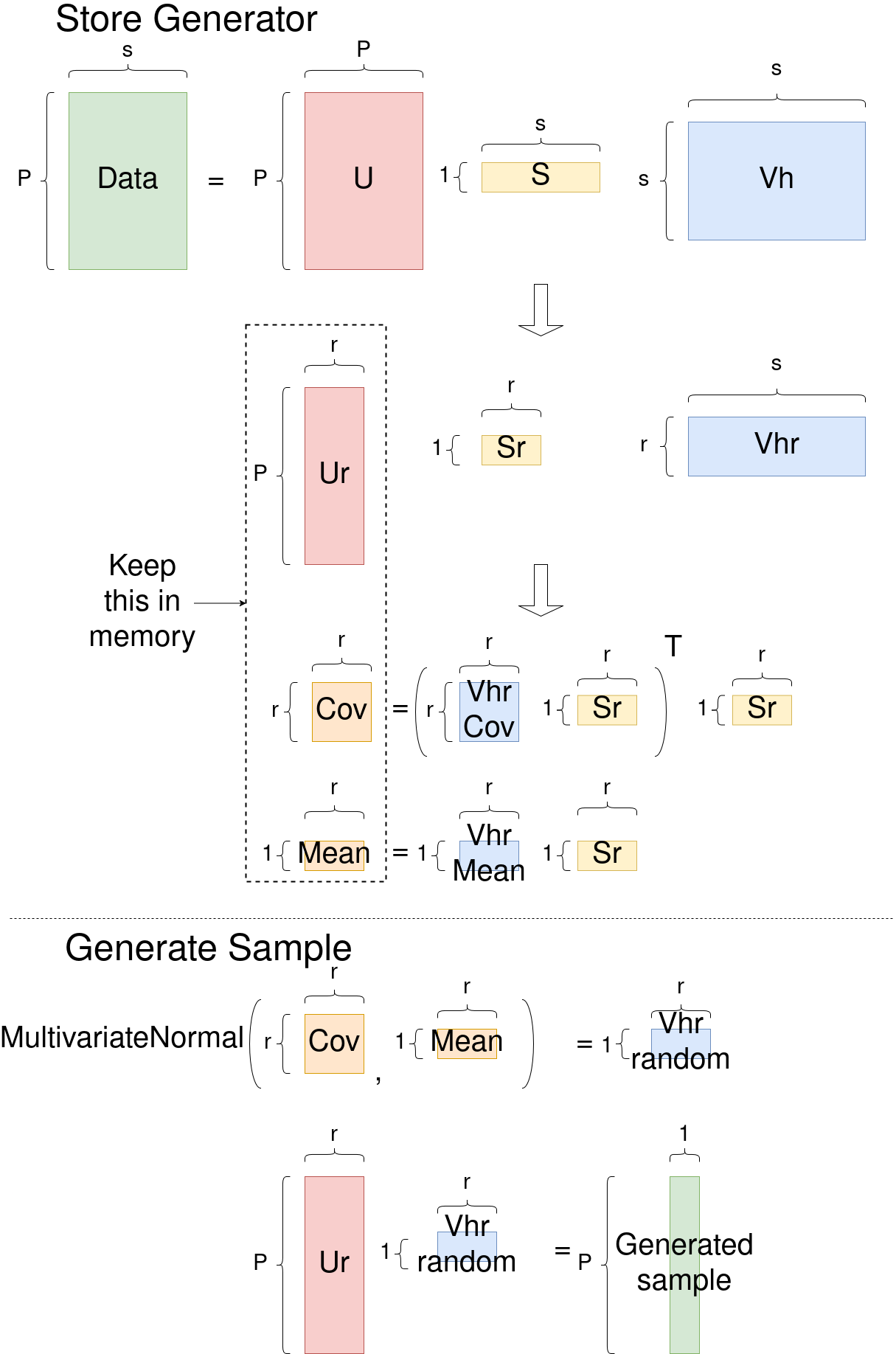}
\caption{SVD decomposes the data into the matrices $U$, $S$ and $Vh$. The only information stored for use by the generator is the rank-reduced matrix $Ur$ and the covariance matrix and mean of $Vhr$. The covariance and mean are used to draw a random sample from a multivariate normal, which in turn is combined with $Ur$ to generate a synthetic data sample.}
\label{fig:SVDgenerator}
\end{figure}

In our work, we investigate the possibility of using lightweight generators in memory-based continual learning algorithms. More specifically, we propose to replace the episodic memory of A-GEM with a lightweight generator based on SVD. We choose to use A-GEM and ER, since they are relatively fast and have a low memory footprint. However, the concept of lightweight SVD generators can be used in any memory-based continual learning algorithm.\\
In our method, SVD transforms the data to three matrices, namely $U$, $S$ and $Vh$, in linear time. Our method uses Truncated SVD with time complexity $\mathcal{O}(2sP^2 + P^3 + P + sP)$, for a $P \times s$ matrix , where $P$ is the number of pixels and $s$ is he number of data samples and $P \ll s$ \cite{li2019tutorial}. For a small constant $P$, this results in $\mathcal{O}(s)$, i.e. linear time, making it significantly cheaper to fit to a dataset than a VAE or a Generative Adversarial Neural Network (GANN)\\
The top part of figure \ref{fig:SVDgenerator} (Store Generator) shows how the information is created, that is used by a lightweight generator. It shows the data as a matrix of size $P \times s$, where $s$ is the number of data samples and $P$ is the length of each image flattened into a vector. The $U$ matrix represents the data as components ordered from biggest (most variance) to smallest (least variance). $S$ is depicted in a flattened form of size $1 \times s$, by keeping only the diagonal of the original $S$ matrix. This is an equivalent representation, since the original $S$ matrix contains only non zero entries on its diagonal. The combination of $S$ and the matrix $Vh$ of size $s \times s$ define how the components of $U$ can be recombined to get the original data back.\\
We consider the biggest components in $U$ to be the most useful information. Therefore we apply \textit{Truncated SVD}, meaning we choose to only keep the biggest components, while removing the smallest, thereby keeping the most useful information. We define a \textit{rank} $r$ ($r<s$) as the number of components to keep, resulting in the reduced matrices $Ur$ with size $P \times r$, $Sr$ with size $1 \times r$ and $Vhr$ with size $r \times P$. To reduce the memory footprint further, we choose to only keep the covariance of size $r \times r$ and the mean of size $1 \times r$ of the $Vhr$ matrix. We multiply these by $Sr$ and then store them. Note that the covariance matrix is multiplied by $Sr$, transposed, then again multiplied by $Sr$. This is needed to scale the covariance matrix with $Sr$ on both axes. After this process we only need to store $P*r + r^2 + r$ entries.
The bottom part of Figure \ref{fig:SVDgenerator} (Generate Sample) shows how this information is used to generate synthetic data samples. The covariance matrix and the mean are used to draw a random vector ($Vhr$ random) from a multivariate normal distribution. The dot product of $Ur$ and this random vector produces a synthetic generated sample.

\subsection{Generator Integration}
Algorithm \ref{alg:AGEMedit} shows our modified version of A-GEM. Task identities are represented by $k$, with $T_k$ denoting task $k$ and $T_{k_t}$ denoting a batch of data taken from $T_k$ at time step $t$. $L_{(x,y)}$ is the loss function that takes as arguments $x$, the data and $y$, the label. $L(w)$ is the calculated value of this loss function on the weight configuration $w$.\\
For the first task, i.e. $k = 0$, the gradient of $L(w)$ is calculated, denoted by $\nabla L(w)$, just like the original version of A-GEM, as can be seen in line 5-6 of Algorithm \ref{alg:AGEMedit}. After training on task 0 has finished, our algorithm starts to differ from A-GEM, as can be seen on line 24-33. For all $c$ classes, an empty data buffer is initialized for the current task $k$ (line 24-26). Then $s$ samples are taken from $T_k$ and put into their corresponding buffer, depending on their class $y$ (line 27-30). Finally, the data of buffer $\mathrm{data}_{k,y}$, task identity $k$ and class label $y$ are used to create a generator for each of the $c$ classes for task $k$ (line 31-33). Where the original A-GEM stores raw data samples, we create generators and store the necessary information for these generator.\\
For the subsequent tasks, i.e. $k > 0$, the generators are used to create synthetic data, on which reference gradients are calculated (line 8-20). First, an empty buffer $M$ is initialized (line 8). Then, $n$ samples are created and stored in this buffer (line 9-13), by selecting a task (line 10) and class (line 11) at random. A sample is generated for this task and class and added to $M$ (line 12). Just like in the original A-GEM, the content of $M$ is used to calculated a reference gradient $\nabla L_{ref}(w)$ (line 14-20). If the dot product between the gradient $\nabla L(w)$ and the reference gradient $\nabla L_{ref}(w)$ is greater or equal to $0$, $\nabla L(w)$ is used (line 16) to take a gradient descent step with step size $\eta$ (line 22). Otherwise, the gradient $\nabla L(w)$ is first projected onto the reference gradient $\nabla L_{ref}(w)$ (line 18) and then subtracted from $\nabla L(w)$, resulting in gradient $\tilde{g}$, which is orthogonal to $\nabla L_{ref}(w)$ (line 19). This orthogonalized gradient $\tilde{g}$ is then used to take a gradient descent step (line 22). Where the original A-GEM takes the samples in $M$ directly from the episodic memory, we use the generators to make synthetic samples. After the training process on task $k$ is finished, the generators are updated (line 24 - 33)\\
Algorithm \ref{alg:ERedit} shows how the lightweight generators are included in ER. First, $M$ is initialized as an empty set (line 4). If $k \not = 0$, $M$ is filled with generated samples (line 6 - 10). The loss $L(w)$ is calculated on the samples of both batch $T_{k_t}$ and the samples in $M$ (line 12). The gradient of the loss, $\nabla L(w)$, is used to update the weights with step size $\eta$ (line 13). After the training on task $k$ is finished, the generators are updated (line 15 - 24).\\
Algorithm \ref{alg:storegenerator} shows the procedure to store the necessary information for one generator. Each generator corresponds to one \textit{task} and one \textit{class}. The \textit{data} of one task and class is factorized into the matrices $U$, $S$ and $Vh$, using SVD (line 1). These matrices are truncated according to a defined \textit{rank}, by only keeping the \textit{rank} biggest components of $U$ (leftmost columns), the \textit{rank} first elements of $S$ and the \textit{rank} dimensions of each vector in $Vh$ (the topmost rows) (line 2). The \textit{mean} and \textit{covariance} matrix of $Vh$ are calculated and multiplied by $S$ (line 3-4). Finally, the $U$, the \textit{mean} and the \textit{covariance} matrix are stored and indexed by their \textit{task} and \textit{class} (line 5).\\
Algorithm \ref{alg:gensamp2} shows how a synthetic data sample is generated for a given \textit{task} and \textit{class} using  $U_{\mathrm{task},\mathrm{class}}$, $\mathrm{mean}_{\mathrm{task},\mathrm{class}}$, and $ \mathrm{cov}_{\mathrm{task},\mathrm{class}}$. A random vector $Vh_{\mathrm{random}}$ is generated from a multivariate normal distribution, using the stored $\mathrm{mean}_{\mathrm{task},\mathrm{class}}$, and $ \mathrm{cov}_{\mathrm{task},\mathrm{class}}$ (line 1). The dot product is taken from $U_{\mathrm{task},\mathrm{class}}$ and $Vh_{\mathrm{random}}$, resulting in a generated sample (line 2). This sample is returned together with the given class as a data-label pair (line 3).\\

\begin{algorithm}
\caption{Averaged Gradient Episodic Memory with SVD generator}\label{alg:AGEMedit}
\hspace*{\algorithmicindent} \textbf{Input:} Task sequence $T_0, T_1, T_2, ...$ learning rate $\eta$\\
\hspace*{\algorithmicindent} \textbf{Output:} Model
\begin{algorithmic}[1]
\State \textbf{Initialize} $M \gets \{\}; w \gets w_0$
\For{Task ID $k = 0, 1, 2, ...$}
    \Repeat
        \State $L(w) \gets \frac{1}{|T_{k_t}|}\sum_{(x,y)\in T_{k_t}} L_{(x,y)}(w)$
        \If{$k = 0$}
            \State $\tilde{g} \gets \nabla L(w)$
        \Else 
            \State $M \gets \{\}$
            \For{$i = \{1, ..., n\}$}
                \State $\mathrm{task} \gets \mathrm{RandomSelect}(\{0, ..., k - 1\})$
                \State $\mathrm{class} \gets \mathrm{RandomSelect}(\{0, ..., c-1\})$
                \State $M \gets M \cup \mathrm{GenerateSample}(\mathrm{task}, \mathrm{class})$
            \EndFor
            \State $L_{ref}(w) \gets \frac{1}{|M|}\sum_{(x,y)\in M} L_{(x,y)}(w)$
            \If{$\nabla L(w)^\top \nabla L_{ref}(w) \geq 0$}
                \State $\tilde{g} \gets \nabla L(w)$
            \Else
                \State $g_{proj} \gets \frac{\nabla L(w)^\top \nabla L_{ref}(w)}{\nabla L_{ref}(w)^\top \nabla L_{ref}(w)}\nabla L_{ref}(w)$
                \State $\tilde{g} \gets \nabla L(w) - g_{proj}$
            \EndIf
        \EndIf
        \State $w \gets w - \eta \tilde{g}$
    \Until{converge}
    \For{$y = \{0, ..., c-1\}$}
        \State $\mathrm{data}_{k,y} \gets \{\}$
    \EndFor
    \For{$i = \{1, ..., s\}$}
        \State $(\mathbf{x}, y) \sim T_k$
        \State $\mathrm{data}_{k,y} \gets \mathrm{data}_{k,y} \cup \mathbf{x}$
    \EndFor
    \For{$y = \{0, ..., c-1\}$}
        \State $\mathrm{StoreGenerator}(\mathrm{data}_{k,y}, k, y)$
    \EndFor
\EndFor
\end{algorithmic}
\end{algorithm}

\begin{algorithm}
\caption{Experience Replay with SVD generator}\label{alg:ERedit}
\hspace*{\algorithmicindent} \textbf{Input:} Task sequence $T_0, T_1, T_2, ...$ learning rate $\eta$\\
\hspace*{\algorithmicindent} \textbf{Output:} Model
\begin{algorithmic}[1]
\State \textbf{Initialize} $M \gets \{\}; w \gets w_0$
\For{Task ID $k = 0, 1, 2, ...$}
    \Repeat
        \State $M \gets \{\}$
        \If{$k \not = 0$}
            \For{$i = \{1, ..., n\}$}
                \State $\mathrm{task} \gets \mathrm{RandomSelect}(\{0, ..., k - 1\})$
                \State $\mathrm{class} \gets \mathrm{RandomSelect}(\{0, ..., c-1\})$
                \State $M \gets M \cup \mathrm{GenerateSample}(\mathrm{task}, \mathrm{class})$
            \EndFor
        \EndIf
        \State $L(w) \gets \frac{1}{|T_{k_t}| + |M|}\sum_{(x,y)\in T_{k_t} \cup M} L_{(x,y)}(w)$
        \State $w \gets w - \eta \nabla L(w)$
    \Until{converge}
    \For{$y = \{0, ..., c-1\}$}
        \State $\mathrm{data}_{k,y} \gets \{\}$
    \EndFor
    \For{$i = \{1, ..., s\}$}
        \State $(\mathbf{x}, y) \sim T_k$
        \State $\mathrm{data}_{k,y} \gets \mathrm{data}_{k,y} \cup \mathbf{x}$
    \EndFor
    \For{$y = \{0, ..., c-1\}$}
        \State $\mathrm{StoreGenerator}(\mathrm{data}_{k,y}, k, y)$
    \EndFor
\EndFor
\end{algorithmic}
\end{algorithm}

\begin{algorithm}
\caption{StoreGenerator}\label{alg:storegenerator}
\hspace*{\algorithmicindent} \textbf{Input:} data,task,class\\
\begin{algorithmic}[1]
\State $U, S,Vh \gets \mathrm{SVD}(\mathrm{data})$
\State $U, S,Vh \gets \mathrm{Truncate}(U, S,Vh)$
\State mean $\gets \frac{\Sigma_{\forall \mathbf{v} \in Vh} \mathbf{v}}{|Vh|} * S$
\State cov $\gets (\mathrm{Covariance}(Vh) * S)^T * S$
\State $U_{\mathrm{task},\mathrm{class}}, \mathrm{mean}_{\mathrm{task},\mathrm{class}}, \mathrm{cov}_{\mathrm{task},\mathrm{class}} \gets U, \mathrm{mean}, \mathrm{cov}$
\end{algorithmic}
\end{algorithm}

\begin{algorithm}
\caption{GenerateSample}\label{alg:gensamp2}
\hspace*{\algorithmicindent} \textbf{Input:} task,class\\
\hspace*{\algorithmicindent} \textbf{Output:} (sample, class)
\begin{algorithmic}[1]
\State $Vh_{\mathrm{random}} \gets \mathrm{MultivariateNormal}(\mathrm{mean}_{\mathrm{task},\mathrm{class}}, \mathrm{cov}_{\mathrm{task},\mathrm{class}})$
\State sample $\gets U_{\mathrm{task},\mathrm{class}} \cdot Vh_{\mathrm{random}}$
\State \Return (sample, class)
\end{algorithmic}
\end{algorithm}

\subsection{Memory Requirements}
By keeping only the biggest components after SVD, we can reduce the memory requirements. We refer to the number of kept components as the \textit{rank}. Equation \ref{eq:rankcompression} shows how we define the compression factor $F$ using the number of pixels per image $P$, the number of samples $s$ from the dataset, the number of classes $c$ and the rank $r$. 

\begin{equation}\label{eq:rankcompression}
    F = \frac{P * s}{c * (P * r + r^2+ r)}
\end{equation}

This equation divides the raw memory required ($P * s$) by the memory required by the lightweight generator ($c * (P * r + r^2+ r)$). For every class, the matrix $U$ is truncated by rank, resulting in an $P \times r$ matrix. The covariance matrix of the rank-truncated $Vh$ is a $r \times r$ matrix. The mean of the $Vh$ matrix, after rank truncation is a vector of length $r$. $S$ does not need to be stored, since the covariance matrix and the mean are multiplies by $S$ before storing them. Since a lightweight generator is kept for every class, the number is multiplied by the number of classes $c$.

\subsection{Experimental Setup}
\begin{figure}[H]
\centering
\includegraphics[width=1.0\linewidth]{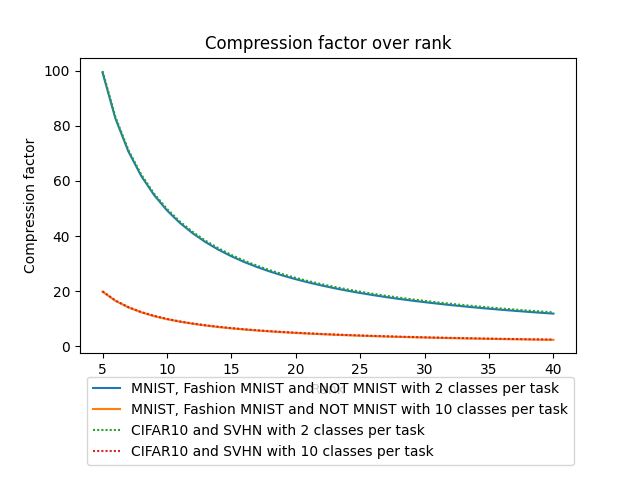}
\caption{Compression factor per the rank of SVD for 2 or 10 classes per task when taking 1000 data points.}
\label{fig:Compression}
\end{figure}

For Fashion MNIST \cite{xiao2017fashion}, MNIST \cite{deng2012mnist} and NOT MNIST \cite{notmnist}, $P = 28 * 28$. For a color image, each colored pixel has three channels, so in our definition a color pixel is equal to three pixels. Thus for CIFAR10 \cite{cifar10} and SVHN \cite{netzer2011reading}, $P = 3 * 32 * 32$. The effect of the rank over the compression factor can be seen in figure \ref{fig:Compression}.\\
For our experiments we set the \textit{rank} $r$ to 5 and the number of \textit{samples} $s$ to $1000$, because this gave satisfactory results in our early experiments. For Fashion MNIST, MNIST and NOT MNIST this gives us a compression factor $F$ of $19.85$ in the case of $10$ classes (the rotation experiments) and of $99.24$ in the case of $2$ classes (the class split experiments). For CIFAR10 and SVHN, the compression factor $F$ is $19.96$ in the case of $10$ classes (the rotation experiments) and of $99.81$ in the case of $2$ classes (the class split experiments).\\
We compare our method to the lower bound of Stochastic Gradient Descent (SGD), where nothing is done to alleviate the catastrophic forgetting. As an upper bound we compare to standard A-GEM and ER, where we take 1000 data samples per task and store them in memory. The lightweight generators are constructed with 1000 samples, the same number of samples as the standard A-GEM and ER use. Since the lightweight generators receive the same number of samples as the standard A-GEM, we do not expect that the lightweight generator experiments outperform the standard A-GEM. In order to have a fair comparison, we compare our method with A-GEM, where the number of stored samples per task is reduced to match the memory required for both approaches. We set the number of samples per task to $51$ for the rotation experiments and to $11$ for the task split experiments. The number of samples is determined by dividing $1000$ by the compression rate and rounding up. So for Fashion MNIST, MNIST and NOT MNIST, this is $1000/19.85 = 50.38$ for the rotation experiments and $1000/99.24 = 10.08$ for the class split experiments. These numbers are then rounded up to the nearest integer value, namely $51$ and $11$ samples. For CIFAR, this results in the same rounded values, namely $1000/19.96 = 50.10$ and $1000/99.81 = 10.01$, rounding up to $51$ and $11$. In this case we give A-GEM with reduced samples a slight advantage, where it can use slightly more memory than our proposed method (A-GEM with lightweight SVD generator). 

\newpage

\vspace{5mm}

For a continual learning setting, distinct tasks are required. We create tasks from standard datasets by means of \textit{rotation} and \textit{class splitting}. In case of rotation, each task contains the entire original dataset, but each entry is rotated clockwise by $d$ degrees, where $d \in \{0, 20, 40, 60, 80, 100, 120, 140, 160, 180\}$, thus creating 10 tasks, with 10 classes each. For the class splitting, each tasks consists of a subset of two classes from the original sample. The chosen subsets are $\{0, 1\},\{2, 3\},\{4, 5\}, \{6, 7\}, \{8, 9\}, \{0, 3\},$ $ \{2, 5\},\{4, 7\}, \{6, 9\}$ and $\{8, 1\}$, creating 10 tasks, with 2 classes each. After splitting the dataset into tasks, the labels of all tasks are cast to the same subspace, namely: $0$ for even numbered classes and $1$ for odd numbered classes.\\
We evaluate our method on three architectures: a Multi Layer Perceptron (MLP) \cite{popescu2009multilayer,almeida2020multilayer}, a Multi Layer Perceptron mixer (MLP mixer) \cite{tolstikhin2021mlpmixer} and ResNet18 \cite{he2016deep}. The MLP consists of two hidden layers, each of 200 nodes. The MLP mixer has a patch size of 4 by 4, a depth of 8, each block has 256 nodes. ResNet18 is a convolutional neural network with residual connections that is 18 layers deep.

\subsection{Evaluation Metric}
In order to measure the ability of a method to not forget and to learn new tasks, the \textit{Average Validation Accuracy} is used. This is calculated by equation \ref{eq:avgvalacc}, where the average validation accuracy $A$ on task $T_k$ is defined as the average of all $k$ validation accuracies on the validation sets $V_i$ of the tasks $T_i$, where $i \in [0,k]$, and the function $f$ gives the predicted label of neural network on a give data sample $\mathbf{x}$.

\begin{equation}\label{eq:avgvalacc}
    A = \frac{1}{k} \sum_{i=0}^{k} \frac{1}{|V_i|}\sum_{(\mathbf{x},y) \in V_i} \mathrm{if} f(\mathbf{x}) = y: 1 \, \mathrm{else} \, 0
\end{equation}

\section{Results and Discussion}
\begin{figure*}[!ht]
\centering
\begin{subfigure}{.32\textwidth}
    \includegraphics[width=1.0\textwidth]{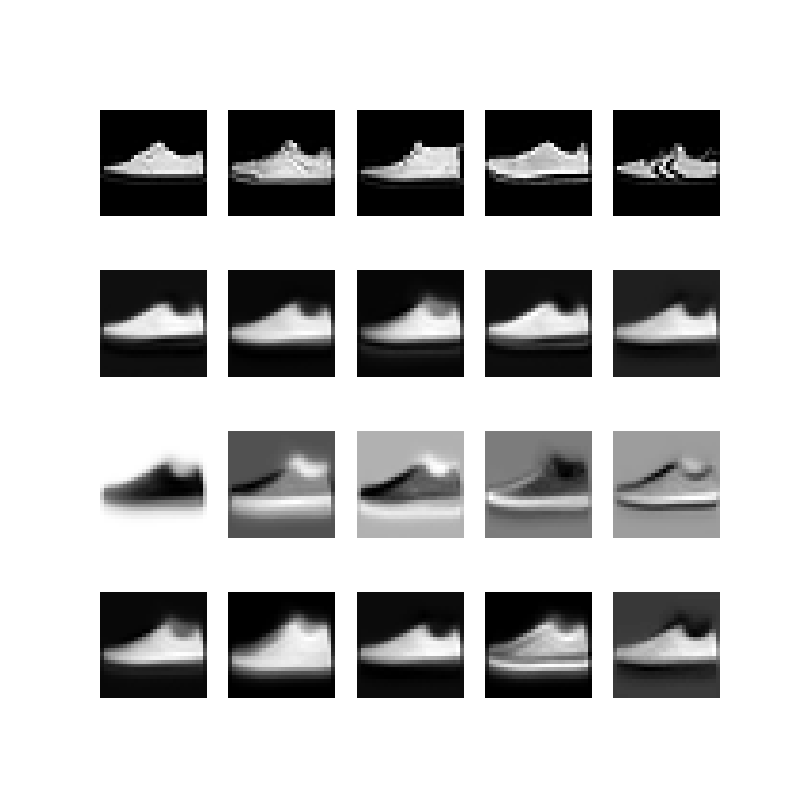}
    \caption{Fashion MNIST, class 7}
    \label{fig:GENpap0}
    \vspace*{2mm}
\end{subfigure}
\begin{subfigure}{.32\textwidth}
    \includegraphics[width=1.0\textwidth]{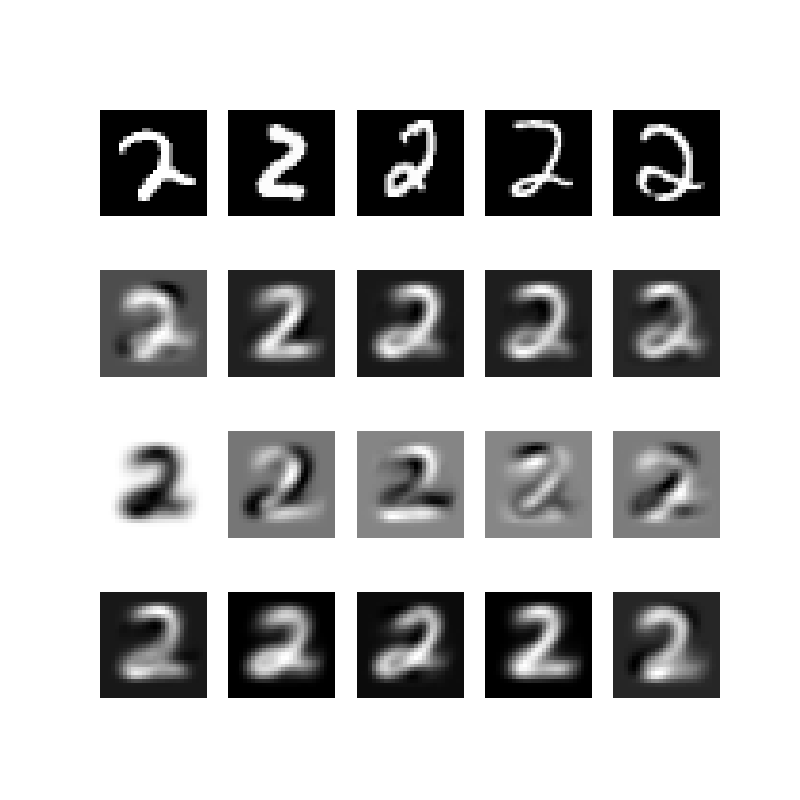}
    \caption{MNIST, class 2}
    \label{fig:GENpap1}
    \vspace*{2mm}
\end{subfigure}
\begin{subfigure}{.32\textwidth}
    \includegraphics[width=1.0\textwidth]{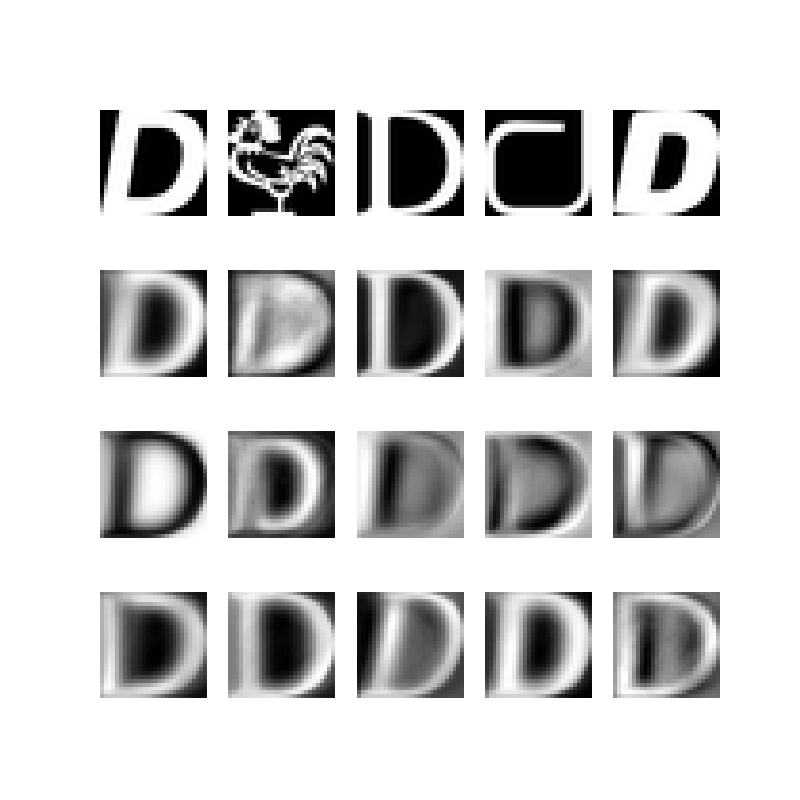}
    \caption{NOT MNIST, class 3}
    \label{fig:GENpap2}
    \vspace*{2mm}
\end{subfigure}\\
\begin{subfigure}{.32\textwidth}
    \includegraphics[width=1.0\textwidth]{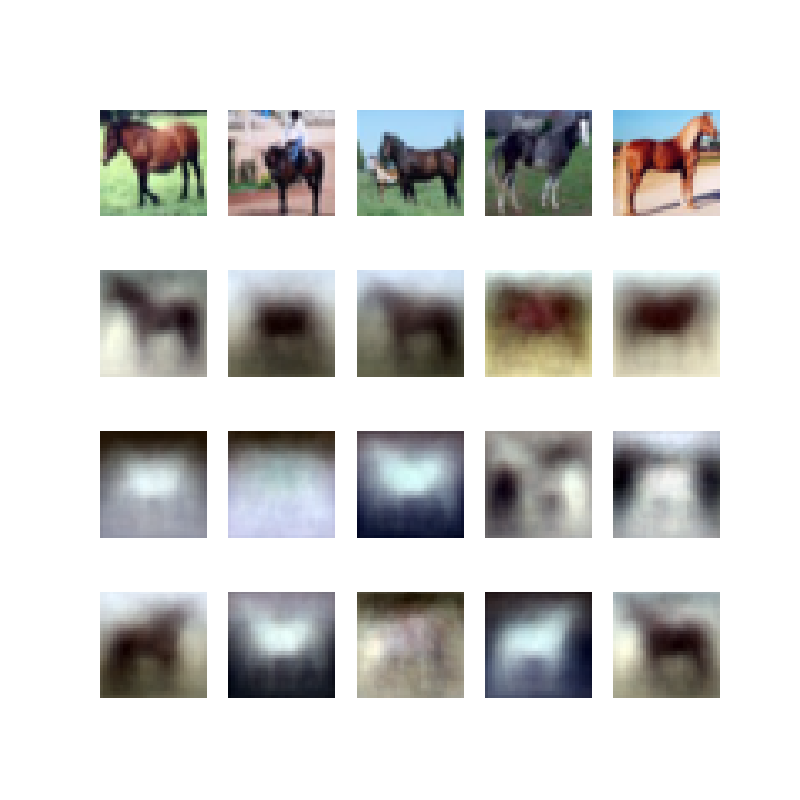}
    \caption{CIFAR10, class 7}
    \label{fig:GENpap3}
    \vspace*{2mm}
\end{subfigure}
\begin{subfigure}{.32\textwidth}
    \includegraphics[width=1.0\textwidth]{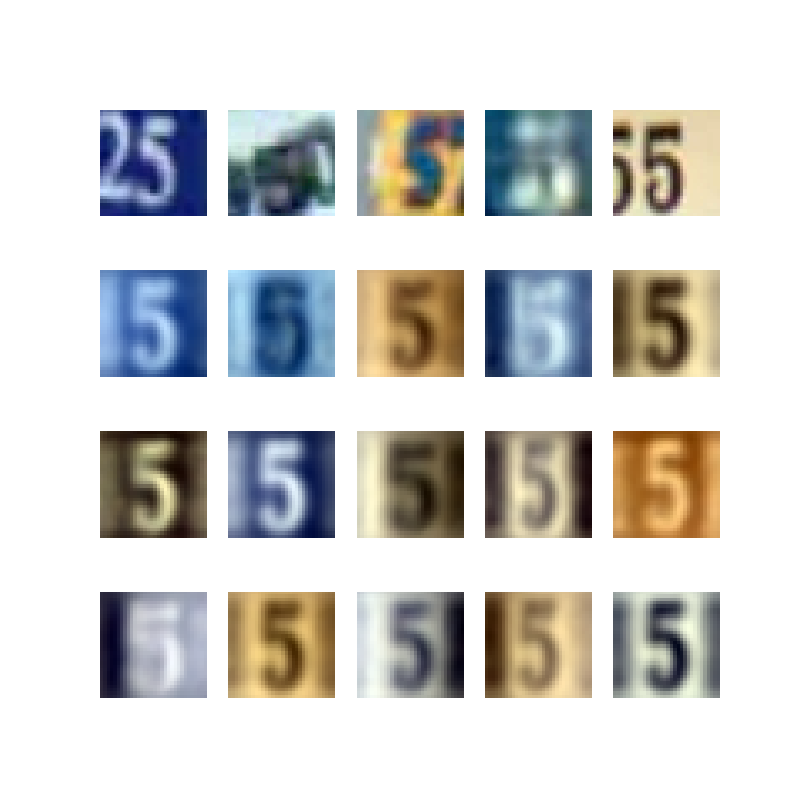}
    \caption{SVHN, class 5}
    \label{fig:GENpap4}
    \vspace*{2mm}
\end{subfigure}
\caption{Five examples for the different datasets of the lightweight generators. Per example, first row: Instances of one class. Second row: Recreation of instances using 5 components. Third row: The 5 components with the most variation, from big to small. Fourth row: Images synthesized by the lightweight generator, using 5 components.}
\label{fig:GENpap}
\vspace*{4mm}
\end{figure*}

\begin{figure*}[!ht]
\centering
\begin{subfigure}{1.0\textwidth}
    \includegraphics[width=.5\textwidth]{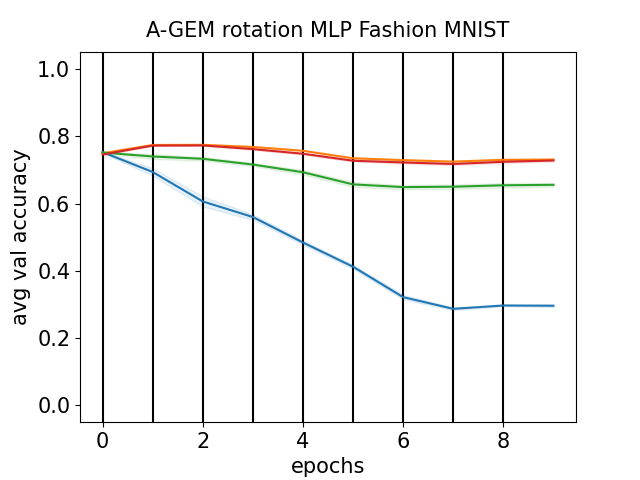}
    \includegraphics[width=.5\textwidth]{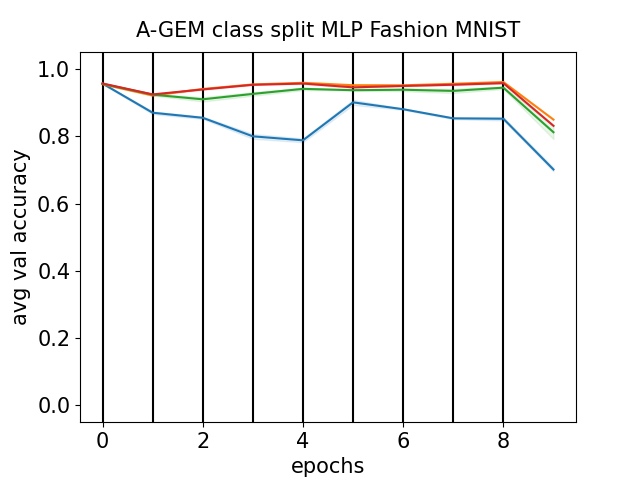}
    \caption{Average validation accuracy on Fashion MNIST}
    \label{fig:MLP_Fashion_MNIST}
    \vspace*{5mm}
\end{subfigure}
\begin{subfigure}{1.0\textwidth}
    \includegraphics[width=.5\textwidth]{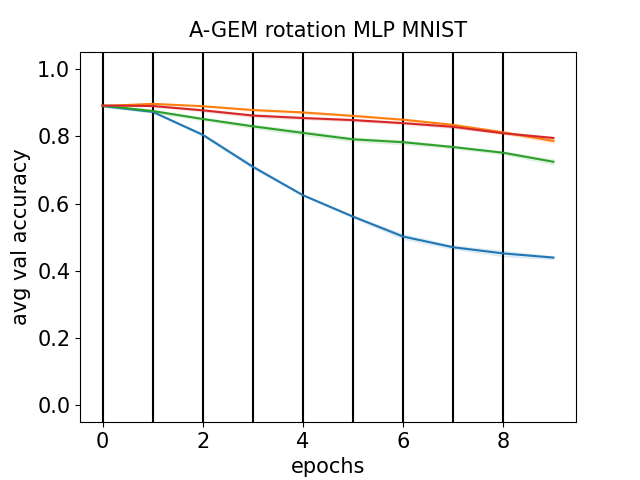}
    \includegraphics[width=.5\textwidth]{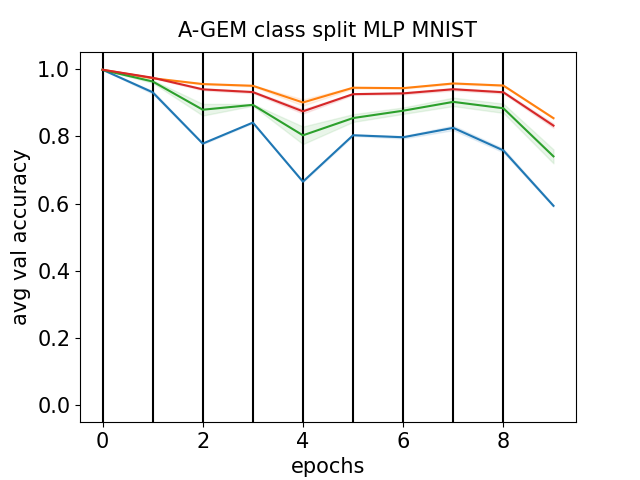}
    \caption{Average validation accuracy on MNIST}
    \label{fig:MLP_MNIST}
    \vspace*{5mm}
\end{subfigure}\\
\begin{subfigure}{1.0\textwidth}
    \includegraphics[width=.5\textwidth]{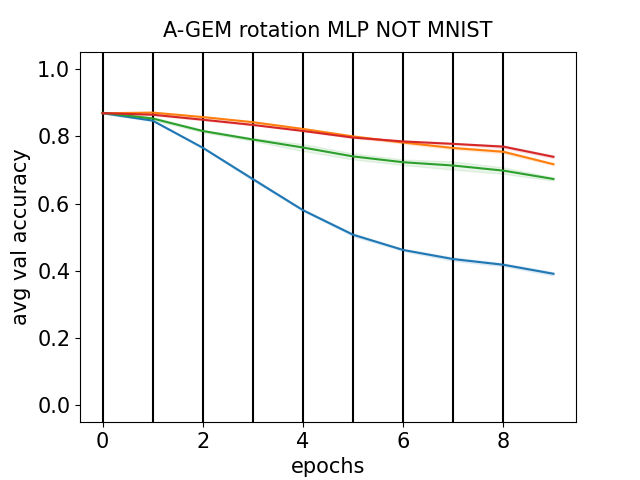}
    \includegraphics[width=.5\textwidth]{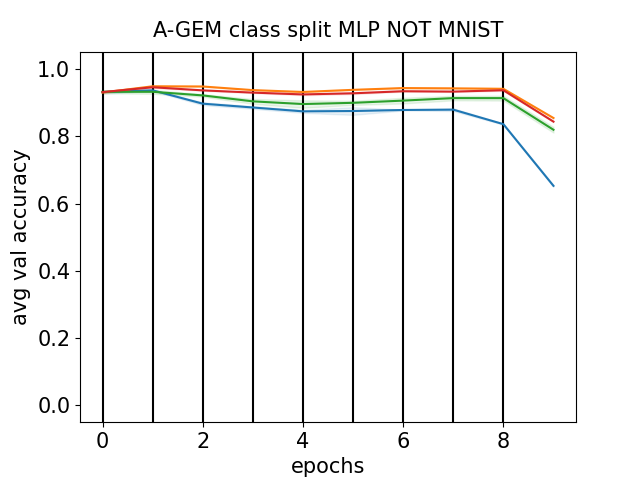}
    \caption{Average validation accuracy on NOT MNIST}
    \label{fig:MLP_NOT_MNIST}
    \vspace*{4mm}
\end{subfigure}\\
\begin{subfigure}{1.0\textwidth}
    \centering
    \includegraphics[width=1.0\textwidth]{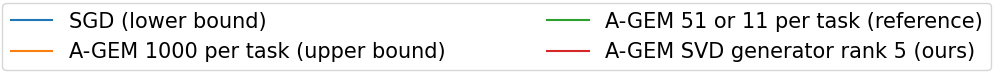}\hfill
    \label{fig:MLP_legend}
    \vspace*{4mm}
\end{subfigure}
\caption{These figures depict the average validation accuracy on all tasks trained on, using the \textbf{MLP} architecture and \textbf{A-GEM} method. The average of $5$ independent experiments is depicted as a line, surrounded by a shaded area of one standard deviation.}
\label{fig:MLP}
\vspace*{4mm}
\end{figure*}

Figure \ref{fig:GENpap} shows examples of what our lightweight generator uses to generate images, and what images it produces. Each sub-figure shows four rows. The first row shows samples of one class, the second row shows the same samples but reconstructed using the five biggest principle components. The third row shows what the five largest components look like, when plotted as an image. The fourth row shows samples made by the lightweight generator, using these five components.\\
The generated samples of Fashion MNIST (Figure \ref{fig:GENpap0} 4th row) show a great variety of shoe shapes and types, varying stripes on shoes and shoe shapes. When looking at the generated MNIST samples (Figure \ref{fig:GENpap1} 4th row), note that some samples have a clear curl in the left bottom corner, while some samples end in a sharp turn. Looking at the original samples of NOT MNIST and its reconstructions (Figure \ref{fig:GENpap2} 1st and 2nd row), note that the image of the rooster (2nd image 1st row) is reconstructed as the letter D. This shows the implicit property of SVD of capturing the most common global patterns first, hence the shape of the letter D. The most remarkable are the generated images of CIFAR10 (Figure \ref{fig:GENpap3} 4th row), which show a general horse shape. We did not expect our generator to capture such a pattern from such a highly irregular dataset as CIFAR10. The generated images from SVHN (Figure \ref{fig:GENpap3} 4th row), demonstrate that the generator manages to produce samples in a wide variety of color pallets.\\
In Figure \ref{fig:MLP} we can see the average validation accuracy on all tasks presented to the learner thus far. These are the results on the MLP architecture, where every experiment was repeated five times. The average of these five runs is depicted along with a shaded area of one standard deviation.\\
Note that for all datasets, the average validation accuracy of our method (A-GEM SVD generator with rank 5) is almost as high as our upper bound (A-GEM with 1000 samples per task), while requiring significantly less memory. Compared to a memory equivalent version of A-GEM (A-GEM 51 or 11 per task), our method outperforms it, with the exception of class split Fashion MNIST, where it only slightly outperforms it.\\
The figures belonging to the MLP mixer architecture and the ResNet18 architecture can be seen in the Supplementary material. \\

\begin{figure*}[!ht]
\centering
\begin{subfigure}{1.0\textwidth}
    \includegraphics[width=.5\textwidth]{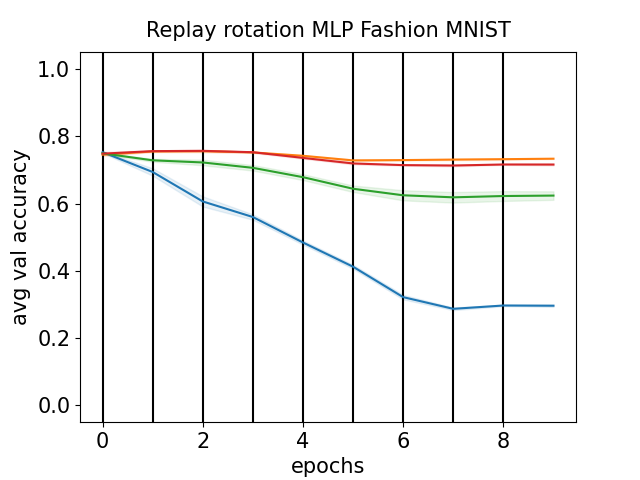}
    \includegraphics[width=.5\textwidth]{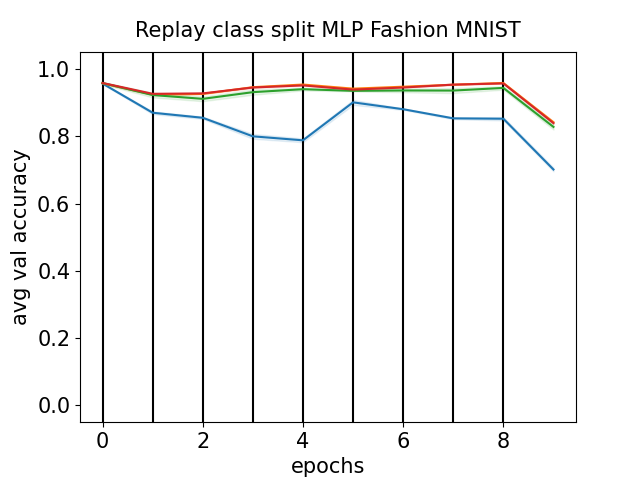}
    \caption{Average validation accuracy on Fashion MNIST}
    \label{fig:MLP_Fashion_MNIST_replay}
    \vspace*{5mm}
\end{subfigure}
\begin{subfigure}{1.0\textwidth}
    \includegraphics[width=.5\textwidth]{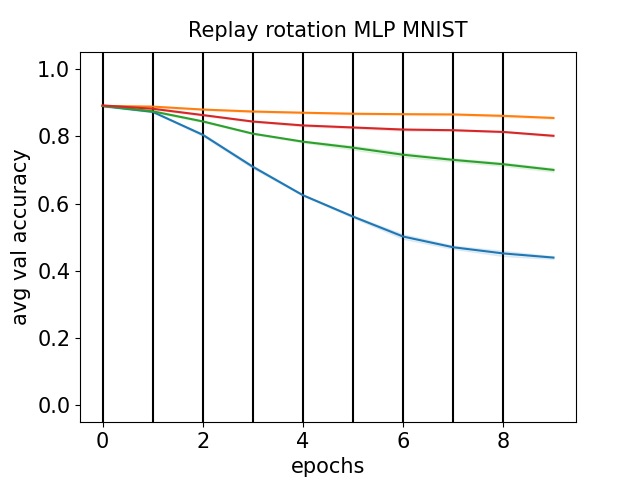}
    \includegraphics[width=.5\textwidth]{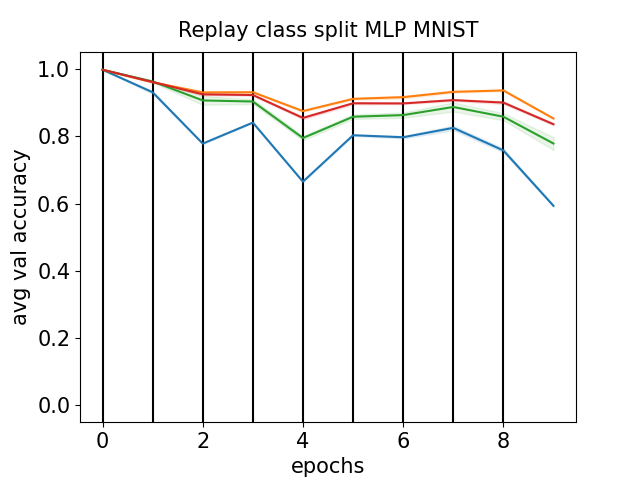}
    \caption{Average validation accuracy on MNIST}
    \label{fig:MLP_MNIST_replay}
    \vspace*{5mm}
\end{subfigure}\\
\begin{subfigure}{1.0\textwidth}
    \includegraphics[width=.5\textwidth]{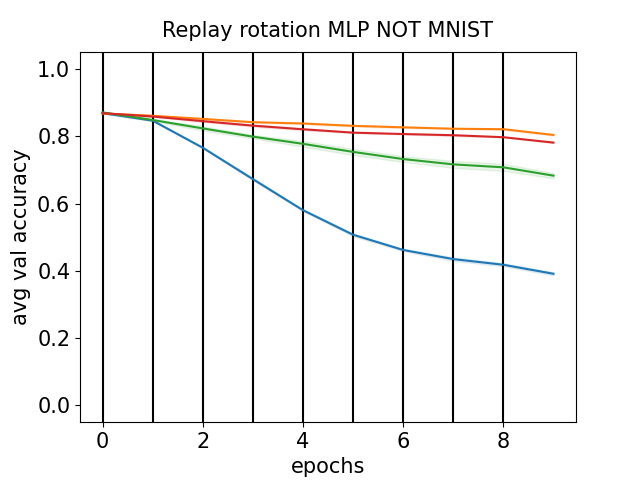}
    \includegraphics[width=.5\textwidth]{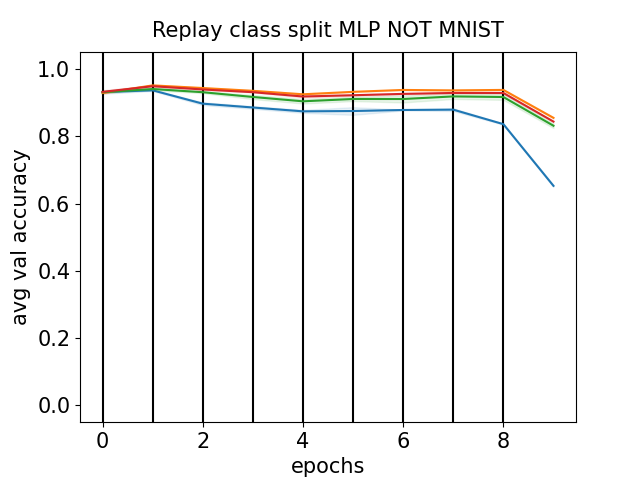}
    \caption{Average validation accuracy on NOT MNIST}
    \label{fig:MLP_NOT_MNIST_replay}
    \vspace*{4mm}
\end{subfigure}\\
\begin{subfigure}{1.0\textwidth}
    \centering
    \includegraphics[width=1.0\textwidth]{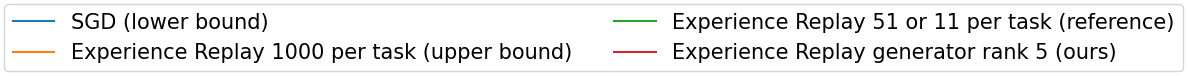}\hfill
    \label{fig:MLP_legend_replay}
    \vspace*{4mm}
\end{subfigure}
\caption{These figures depict the average validation accuracy on all tasks trained on, using the \textbf{MLP} architecture and \textbf{ER} method. The average of $5$ independent experiments is depicted as a line, surrounded by a shaded area of one standard deviation.}
\label{fig:MLP_replay}
\vspace*{4mm}
\end{figure*}

\begin{table*}[!h]
\centering
\caption{Average accuracy, averaged over all epochs, average of 5 runs on the \textit{MLP} architecture, using A-GEM and Experience Replay (ER). Tasks are created from the datasets by rotation (rot) and class split (class) protocols. Highest (significant best with p-value threshold of 0.01) values are denoted in \textbf{bold}. Values that are lower but not significantly lower are also denoted in \textbf{bold}.}
\label{tab:resultsMLP}
\scalebox{0.85}{
\begin{tabular}{c|cc|cc|cc}
    & \multicolumn{2}{c|}{Fashion MNIST} & \multicolumn{2}{c|}{MNIST} & \multicolumn{2}{c}{NOT MNIST}\\
    & rot & class & rot & class & rot & class\\
    \hline
    SGD & 0.471 $\pm$ 0.005 & 0.846 $\pm$ 0.002 & 0.632 $\pm$ 0.003 & 0.799 $\pm$ 0.001 & 0.595 $\pm$ 0.002 & 0.865 $\pm$ 0.002\\
    \hline
    A-GEM 51 / 11 & 0.690 $\pm$ 0.003 & 0.923 $\pm$ 0.003 & 0.807 $\pm$ 0.003 & 0.879 $\pm$ 0.008 & 0.764 $\pm$ 0.004 & 0.904 $\pm$ 0.004\\
    A-GEM gen (ours) & \textbf{0.742} $\pm$ 0.002 & \textbf{0.937} $\pm$ 0.001 & \textbf{0.850} $\pm$ 0.002 & \textbf{0.927} $\pm$ 0.001 & \textbf{0.810} $\pm$ 0.001 & \textbf{0.925} $\pm$ 0.001\\
    \hline
    \color{gray}{A-GEM 1000} & \color{gray}{0.747} $\pm$ 0.001 & \color{gray}{0.940} $\pm$ 0.001 & \color{gray}{0.857} $\pm$ 0.002 & \color{gray}{0.943} $\pm$ 0.002 & \color{gray}{0.808} $\pm$ 0.001 & \color{gray}{0.932} $\pm$ 0.000\\
    \hline
    \hline
    ER 51 / 11 & 0.672 $\pm$ 0.009 & 0.924 $\pm$ 0.003 & 0.786 $\pm$ 0.002 & 0.881 $\pm$ 0.006 & 0.771 $\pm$ 0.005 & 0.911 $\pm$ 0.004\\
    ER gen (ours) & \textbf{0.733} $\pm$ 0.002 & \textbf{0.934} $\pm$ 0.001 & \textbf{0.839} $\pm$ 0.001 & \textbf{0.910} $\pm$ 0.002 & \textbf{0.823} $\pm$ 0.001& \textbf{0.922} $\pm$ 0.002\\
    \hline
    \color{gray}{ER 1000} & \color{gray}{0.740} $\pm$ 0.001 & \color{gray}{0.935} $\pm$ 0.001 & \color{gray}{0.872} $\pm$ 0.002 & \color{gray}{0.925} $\pm$ 0.002 & \color{gray}{0.837} $\pm$ 0.001 & \color{gray}{0.928} $\pm$ 0.001\\
\end{tabular}
}

\end{table*}

\begin{landscape}

\begin{table*}[!ht]
\centering
\caption{Average accuracy, averaged over all epochs, average of 5 runs on the \textit{MLP mixer} architecture, using A-GEM and Experience Replay (ER). Tasks are created from the datasets by rotation (rot) and class split (class) protocols. Highest (significant best with p-value threshold of 0.01) values are denoted in \textbf{bold}. Values that are lower but not significantly lower are also denoted in \textbf{bold}.}
\label{tab:resultsMLPmixer}
\scalebox{0.85}{
\begin{tabular}{c|cc|cc|cc|cc|cc}
    & \multicolumn{2}{c|}{Fashion MNIST} & \multicolumn{2}{c|}{MNIST} & \multicolumn{2}{c|}{NOT MNIST} & \multicolumn{2}{c|}{CIFAR10} & \multicolumn{2}{c}{SVHN}\\
    & rot & class & rot & class & rot & class & rot & class & rot & class\\
    \hline
    SGD & 0.458 $\pm$ 0.008 & 0.788 $\pm$ 0.007 & 0.662 $\pm$ 0.004 & 0.755 $\pm$ 0.005 & 0.579 $\pm$ 0.003 & 0.786 $\pm$ 0.017 & 0.516 $\pm$ 0.002 & 0.696 $\pm$ 0.002 & 0.493 $\pm$ 0.003 & 0.684 $\pm$ 0.002\\
    \hline
    A-GEM 51 / 11 & 0.731 $\pm$ 0.006 & \textbf{0.890} $\pm$ 0.008 & 0.821 $\pm$ 0.005 & 0.853 $\pm$ 0.012 & 0.776 $\pm$ 0.006 & 0.890 $\pm$ 0.008 & \textbf{0.532} $\pm$ 0.002 & \textbf{0.702} $\pm$ 0.002 & 0.599 $\pm$ 0.007 & 0.715 $\pm$ 0.006\\
    A-GEM gen (ours) & \textbf{0.768} $\pm$ 0.002 & \textbf{0.897} $\pm$ 0.002 & \textbf{0.846} $\pm$ 0.003 & \textbf{0.882} $\pm$ 0.010& \textbf{0.806} $\pm$ 0.004 & \textbf{0.914} $\pm$ 0.004 & 0.526 $\pm$ 0.002 & \textbf{0.705} $\pm$ 0.001 & \textbf{0.630} $\pm$ 0.002 & \textbf{0.750} $\pm$ 0.002\\
    \hline
    \color{gray}{A-GEM 1000} & \color{gray}{0.798} $\pm$ 0.002 & \color{gray}{0.941} $\pm$ 0.002 & \color{gray}{0.892} $\pm$ 0.005 & \color{gray}{0.931} $\pm$ 0.002 & \color{gray}{0.852} $\pm$ 0.004 & \color{gray}{0.948} $\pm$ 0.001 & \color{gray}{0.572} $\pm$ 0.002 & \color{gray}{0.756} $\pm$ 0.002 & \color{gray}{0.703} $\pm$ 0.002 &\color{gray}{0.843} $\pm$ 0.003\\
    \hline
    \hline
    ER 51 / 11 & 0.743 $\pm$ 0.009 & \textbf{0.911} $\pm$ 0.007 & 0.845 $\pm$ 0.005 & 0.89 $\pm$ 0.004 & 0.774 $\pm$ 0.008 & 0.899 $\pm$ 0.011 & \textbf{0.541} $\pm$ 0.002 & \textbf{0.711} $\pm$ 0.002 & 0.592 $\pm$ 0.003 & 0.717 $\pm$ 0.005\\
    ER gen (ours) & \textbf{0.793} $\pm$ 0.002 & \textbf{0.918} $\pm$ 0.006 & \textbf{0.892} $\pm$ 0.003 & \textbf{0.913} $\pm$ 0.008 & \textbf{0.825} $\pm$ 0.004 & \textbf{0.918} $\pm$ 0.003 & \textbf{0.536} $\pm$ 0.002 & \textbf{0.712} $\pm$ 0.001 & \textbf{0.667} $\pm$ 0.002 & \textbf{0.762} $\pm$ 0.004\\
    \hline
    \color{gray}{ER 1000} & \color{gray}{0.823} $\pm$ 0.002 & \color{gray}{0.957} $\pm$ 0.001 & \color{gray}{0.946} $\pm$ 0.001 & \color{gray}{0.97} $\pm$ 0.004 & \color{gray}{0.886} $\pm$ 0.001 & \color{gray}{0.950} $\pm$ 0.002 & \color{gray}{0.597} $\pm$ 0.002 & \color{gray}{0.776} $\pm$ 0.002 & \color{gray}{0.765} $\pm$ 0.001 &\color{gray}{0.874} $\pm$ 0.002\\
\end{tabular}
}
\end{table*}

\begin{table*}[!ht]
\centering
\caption{Average accuracy, averaged over all epochs, average of 5 runs on the \textit{ResNet18} architecture, using A-GEM and Experience Replay (ER). Tasks are created from the datasets by rotation (rot) and class split (class) protocols. Highest (significant best with p-value threshold of 0.01) values are denoted in \textbf{bold}. Values that are lower but not significantly lower are also denoted in \textbf{bold}.}
\label{tab:resultsResNet18}
\scalebox{0.85}{
\begin{tabular}{c|cc|cc|cc|cc|cc}
    & \multicolumn{2}{c|}{Fashion MNIST} & \multicolumn{2}{c|}{MNIST} & \multicolumn{2}{c|}{NOT MNIST} & \multicolumn{2}{c|}{CIFAR10} & \multicolumn{2}{c}{SVHN}\\
    & rot & class & rot & class & rot & class & rot & class & rot & class\\
    \hline
    SGD & 0.436 $\pm$ 0.002 & 0.802 $\pm$ 0.007 & 0.690 $\pm$ 0.004 & 0.769 $\pm$ 0.007 & 0.569 $\pm$ 0.003 & 0.777 $\pm$ 0.016 & 0.500 $\pm$ 0.002 & 0.700 $\pm$ 0.002 & 0.557 $\pm$ 0.002 & 0.666 $\pm$ 0.002\\
    \hline
    A-GEM 51 / 11  & \textbf{0.730} $\pm$ 0.006 & \textbf{0.891} $\pm$ 0.010 & \textbf{0.880} $\pm$ 0.002 & \textbf{0.878} $\pm$ 0.007 & \textbf{0.815} $\pm$ 0.010 & \textbf{0.897} $\pm$ 0.007 & \textbf{0.518} $\pm$ 0.002 & \textbf{0.716} $\pm$ 0.001 & 0.671 $\pm$ 0.002 & \textbf{0.679} $\pm$ 0.070\\
    A-GEM gen (ours) & \textbf{0.706} $\pm$ 0.013 & \textbf{0.888} $\pm$ 0.007 & \textbf{0.889} $\pm$ 0.008 & \textbf{0.889} $\pm$ 0.014 & 0.720 $\pm$ 0.018 & \textbf{0.888} $\pm$ 0.007 & 0.502 $\pm$ 0.003 & 0.705 $\pm$ 0.003 & \textbf{0.707} $\pm$ 0.004 & \textbf{0.737} $\pm$ 0.002\\
    \hline
    \color{gray}{A-GEM 1000} & \color{gray}{0.783} $\pm$ 0.002 & \color{gray}{0.942} $\pm$ 0.002 & \color{gray}{0.917} $\pm$ 0.007 & \color{gray}{0.932} $\pm$ 0.004 & \color{gray}{0.878} $\pm$ 0.004 & \color{gray}{0.945} $\pm$ 0.005 & \color{gray}{0.556} $\pm$ 0.002 & \color{gray}{0.780} $\pm$ 0.001 & \color{gray}{0.782} $\pm$ 0.002 & \color{gray}{0.852} $\pm$ 0.003\\
    \hline
    \hline
    ER 51 / 11  & \textbf{0.748} $\pm$ 0.005 & \textbf{0.916} $\pm$ 0.007 & \textbf{0.924} $\pm$ 0.002 & 0.912 $\pm$ 0.01 & \textbf{0.838} $\pm$ 0.005 & \textbf{0.904} $\pm$ 0.006 & \textbf{0.529} $\pm$ 0.001 & \textbf{0.723} $\pm$ 0.002 & 0.705 $\pm$ 0.005 & 0.733 $\pm$ 0.006\\
    ER gen (ours) & \textbf{0.748} $\pm$ 0.008 & \textbf{0.912} $\pm$ 0.006 & \textbf{0.929} $\pm$ 0.006 & \textbf{0.936} $\pm$ 0.004 & 0.768 $\pm$ 0.010 & \textbf{0.901} $\pm$ 0.006 & 0.515 $\pm$ 0.001 & 0.707 $\pm$ 0.002 & \textbf{0.756} $\pm$ 0.002 & \textbf{0.766} $\pm$ 0.003\\
    \hline
    \color{gray}{ER 1000} & \color{gray}{0.825} $\pm$ 0.003 & \color{gray}{0.954} $\pm$ 0.001 & \color{gray}{0.971} $\pm$ 0.001 & \color{gray}{0.976} $\pm$ 0.002 & \color{gray}{0.914} $\pm$ 0.001 & \color{gray}{0.952} $\pm$ 0.001 & \color{gray}{0.615} $\pm$ 0.001 & \color{gray}{0.811} $\pm$ 0.002 & \color{gray}{0.845} $\pm$ 0.001 & \color{gray}{0.906} $\pm$ 0.001\\
\end{tabular}
}
\end{table*}

\end{landscape}

Table \ref{tab:resultsMLP}, \ref{tab:resultsMLPmixer} and \ref{tab:resultsResNet18} denote the averaged accuracy, averaged over all epochs, then averaged over five independent repetitions. The highest scores are highlighted in bold. In case a lower score is not significantly different from the highest with a 99 \% confidence (T test with a p-value of 0.01), it is also highlighted in bold.\\
Table \ref{tab:resultsMLP} shows that our lightweight generator significantly outperforms A-GEM 51 / 11 and ER 51 /11 when using the MLP architecture, on Fashion MNIST, MNIST and NOT MNIST. It is thus significantly more effective at counteracting catastrophic forgetting, while, it uses a similar amount of memory.\\
Table \ref{tab:resultsMLPmixer} shows our results are similar for the MLP mixer architecture, with a few exceptions. For the rotation case of CIFAR10, our method performs worse than A-GEM 51 / 11, while the it is on par with the ER 51 / 11 experiment. For the class split case of Fashion MNIST and CIFAR10, our method's performance is not significantly different from A-GEM 51 / 11.\\
Table \ref{tab:resultsResNet18} shows that our method is not that effective on the ResNet18 architecture. It is at best on par with A-GEM 51 / 11 of ER 51 / 11, and in some case worse (NOT MNIST rotation and CIFAR10). This can be explained by the fact that convolutional neural networks look at local patterns, while SVD mainly captures global patterns in the data, thus being a less effective generator to capture the patterns a convolutional neural network needs for learning and memorizing different tasks.\\
One of the reason our method is successful on Fashion MNIST, MNIST, NOT MNIST and SVHN, could be the regularity of these datasets. The complexity of these datasets is low enough, that most of its data samples can be expressed as linear combinations of only a small number of components. For more complex datasets like CIFAR10, using only five components cannot successfully capture the essence of the dataset.\\
Table \ref{tab:rank80} shows the results, when the number of components are increased from five to $80$. Our method outperformed the raw sampling method in the class split scenario, when using the MLP mixer architecture. This was not the case for the rotation scenario when using the MLP mixer architecture and for both scenarios, when using the ResNet18 architecture.

\begin{table*}[!ht]
\centering
\caption{Average accuracy, averaged over all epochs, average of 5 runs on the \textit{MLP mixer} and \textit{ResNet18} architectures, using A-GEM and Experience Replay (ER). Tasks are created from the datasets by rotation (rot) and class split (class) protocols. Highest (significant best with p-value threshold of 0.01) values are denoted in \textbf{bold}. Values that are lower but not significantly lower are also denoted in \textbf{bold}.}
\label{tab:rank80}
\scalebox{1.07}{
\begin{tabular}{c|cc|cc}
    & \multicolumn{2}{c|}{MLP mixer CIFAR10} & \multicolumn{2}{c}{ResNet18 CIFAR10}\\
    & rot & class & rot & class\\
    \hline
    SGD & 0.516 $\pm$ 0.002 & 0.696 $\pm$ 0.002 & 0.500 $\pm$ 0.002 & 0.700 $\pm$ 0.002\\
    \hline
    A-GEM 822 / 165 & \textbf{0.568} $\pm$ 0.002 & 0.734 $\pm$ 0.002 & \textbf{0.553} $\pm$ 0.002 & \textbf{0.757} $\pm$ 0.002\\
    A-GEM gen rank 80 (ours) & 0.550 $\pm$ 0.000 & \textbf{0.752} $\pm$ 0.003 & 0.529 $\pm$ 0.003 & 0.738 $\pm$ 0.005\\
    \hline
    \color{gray}{A-GEM 1000} & \color{gray}{0.572} $\pm$ 0.002 & \color{gray}{0.756} $\pm$ 0.002 & \color{gray}{0.556} $\pm$ 0.002 & \color{gray}{0.780} $\pm$ 0.001\\
    \hline
    \hline
    ER 822 / 165 & \textbf{0.590} $\pm$ 0.002 & 0.745 $\pm$ 0.002 & \textbf{0.608} $\pm$ 0.001 & \textbf{0.776} $\pm$ 0.002\\
    ER gen rank 80 (ours) & 0.556 $\pm$ 0.003 & \textbf{0.755} $\pm$ 0.003 & 0.553 $\pm$ 0.001 & 0.736 $\pm$ 0.001\\
    \hline
    \color{gray}{ER 1000} & \color{gray}{0.597} $\pm$ 0.002 & \color{gray}{0.776} $\pm$ 0.002 & \color{gray}{0.615} $\pm$ 0.001 & \color{gray}{0.811} $\pm$ 0.002\\
\end{tabular}
}
\end{table*}

\FloatBarrier
\section{Conclusion}
To alleviate catastrophic forgetting with low-memory constraints, we propose the use of lightweight SVD generators in replay based methods such as A-GEM and ER.
The addition of the lightweight SVD generator allows A-GEM and ER to alleviate the effects of catastrophic forgetting more effectively than simply taking raw data samples with a comparable memory footprint. The effectiveness of this method is highly dependent on the dataset and the architecture of the neural network. Our method is more effective on less complex datasets like Fashion MNIST, MNIST, NOT MNIST and SVHN, as compared to more complex datasets like CIFAR10. Our method works better on MLP or MLP mixer architectures, that rely more on global patterns than on ResNet18, which rely on local patterns. The performance of our method can improve when increasing the number of generator components. Our results show the effectiveness of our method in two different scenarios, on several datasets and architectures.\\
Our lightweight SVD generators have an extremely low memory footprint while requiring virtually no training time, making them both highly efficient in terms of memory as well as in computational effort.\\
Future work could include comparing different compression methods such as jpeg compression, resolution downscaling, or non-lossy compression. Other generators like variational autoencoders have been proven to be successful \cite{van2020brain} but require more memory and computational effort. It would be interesting to explore the effect of minimizing the memory requirements for such generators. In our current method, only one generator is used for every class per task, relying on linearity in the latent space. Increasing the number of generators per class increases the potential to effectively capture the manifold of the latent space, since manifolds are locally linear. Such a manifold can be represented by patches of linear subspaces, where each patch is represented by one generator.



\section*{Acknowledgements}
This work is supported by the project ULEARN ``Unsupervised Lifelong Learning'' and co-funded under the grant number 316080 of the Research Council of Norway.

%
%
\FloatBarrier

\bibliographystyle{ieeetr}
\bibliography{mybibfile}

\clearpage

\mainmatter              
\title{Supplementary Material for: Leveraging Lightweight Generators for Memory Efficient Continual Learning}
\titlerunning{Supplementary Material for: Leveraging Lightweight Generators for Memory Efficient Continual Learning}  
%
\author{Christiaan Lamers \inst{1} \and Ahmed Nabil Belbachir \inst{1} \and Thomas Bäck \inst{2} \and Niki van Stein \inst{2}}

\authorrunning{Christiaan Lamers et al.} 

\tocauthor{Christiaan Lamers, Ahmed Nabil Belbachir, Thomas Bäck and Niki van Stein}

\institute{NORCE Norwegian Research Centre, Jon Lilletuns vei 9 H, 3. et,\\
4879 Grimstad, Norway,\\
\email{chla@norceresearch.no}
\and
LIACS, Leiden University, Einsteinweg 55, 2333 CC Leiden, The Netherlands 
}

\maketitle              

\appendix

This supplementary material contains additional figures regarding the experiments of the paper ``Leveraging Lightweight Generators for Memory Efficient Continual Learning". No additional experiments are added in this appendix. All results presented here belong to experiments described in the main paper.

\section{SVD components and lightweight generator output}
\begin{figure*}
\centering
\begin{subfigure}{0.4\textwidth}
    \includegraphics[width=1.0\textwidth]{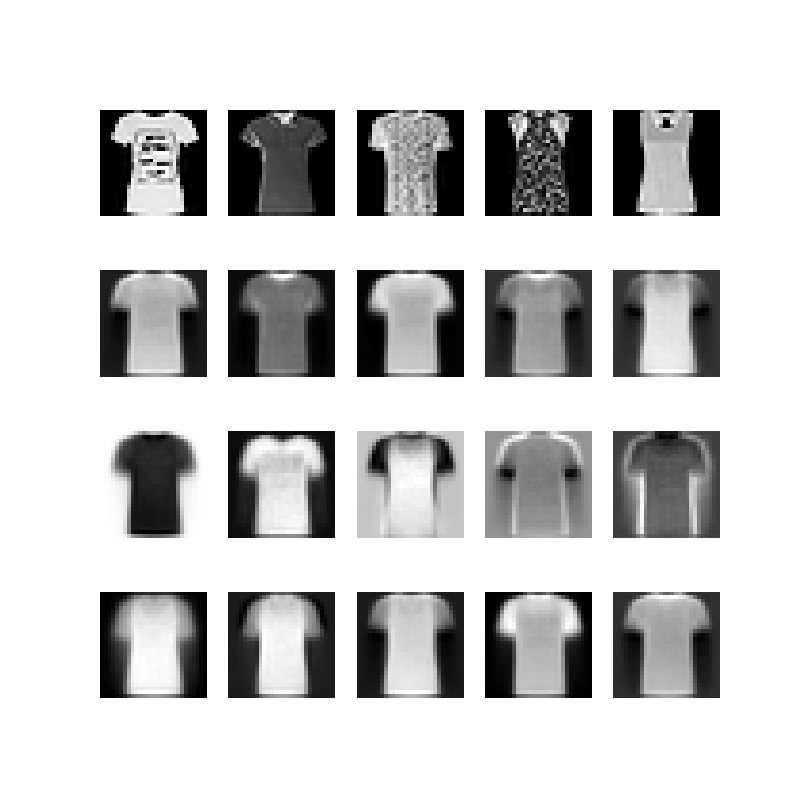}
    \caption{Fashion MNIST, class 0}
    \label{fig:GEN00}
    \vspace*{2mm}
\end{subfigure}
\hfill
\begin{subfigure}{0.4\textwidth}
    \includegraphics[width=1.0\textwidth]{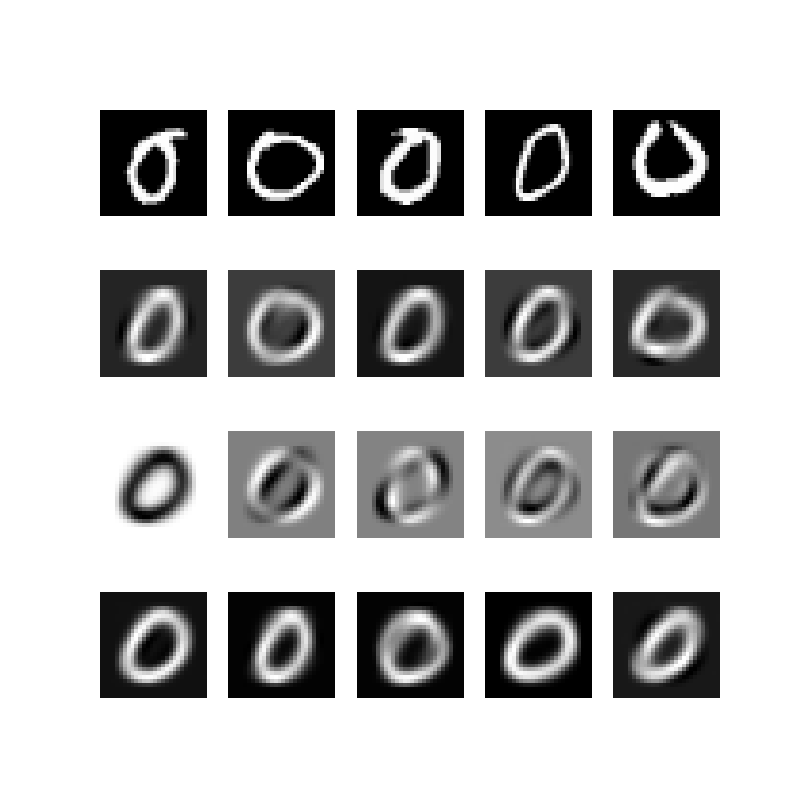}
    \caption{MNIST, class 0 }
    \label{fig:GEN01}
    \vspace*{2mm}
\end{subfigure}\\
\begin{subfigure}{0.4\textwidth}
    \includegraphics[width=1.0\textwidth]{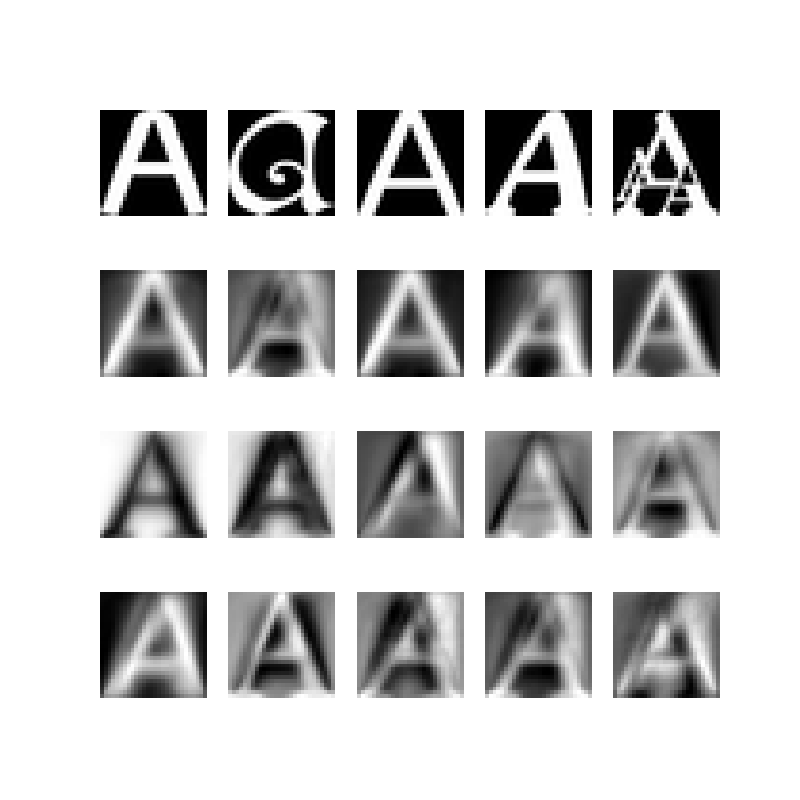}
    \caption{NOT MNIST, class 0}
    \label{fig:GEN02}
    \vspace*{2mm}
\end{subfigure}
\hfill
\begin{subfigure}{0.4\textwidth}
    \includegraphics[width=1.0\textwidth]{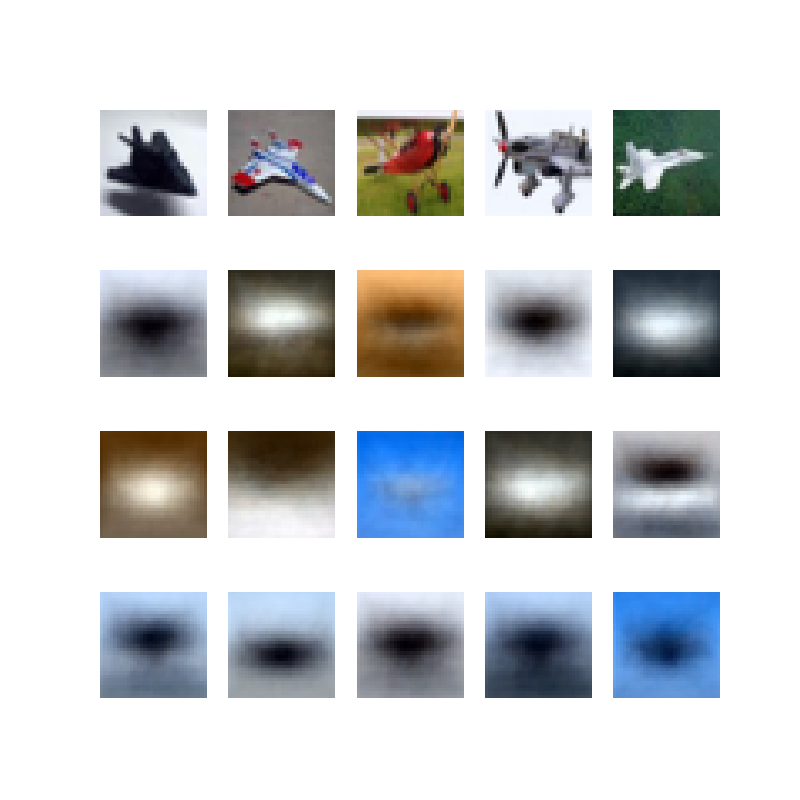}
    \caption{CIFAR10, class 0}
    \label{fig:GEN03}
    \vspace*{2mm}
\end{subfigure}\\
\begin{subfigure}{0.4\textwidth}
    \includegraphics[width=1.0\textwidth]{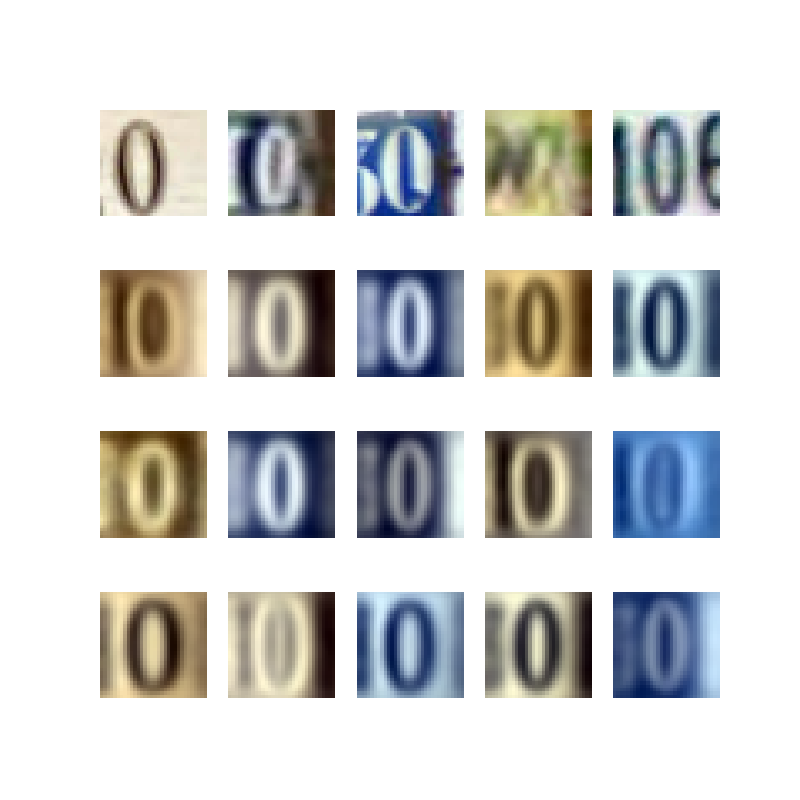}
    \caption{SVHN, class 0}
    \label{fig:GEN04}
    \vspace*{2mm}
\end{subfigure}
\caption{First row: Instances of one class. Second row: Recreation of instances using 5 components. Third row: The biggest 5 components, from big to small. Fourth row: Images synthesized by the lightweight generator, using 5 components.}
\label{fig:GEN0}
\end{figure*}

\begin{figure*}
\centering
\begin{subfigure}{0.4\textwidth}
    \includegraphics[width=1.0\textwidth]{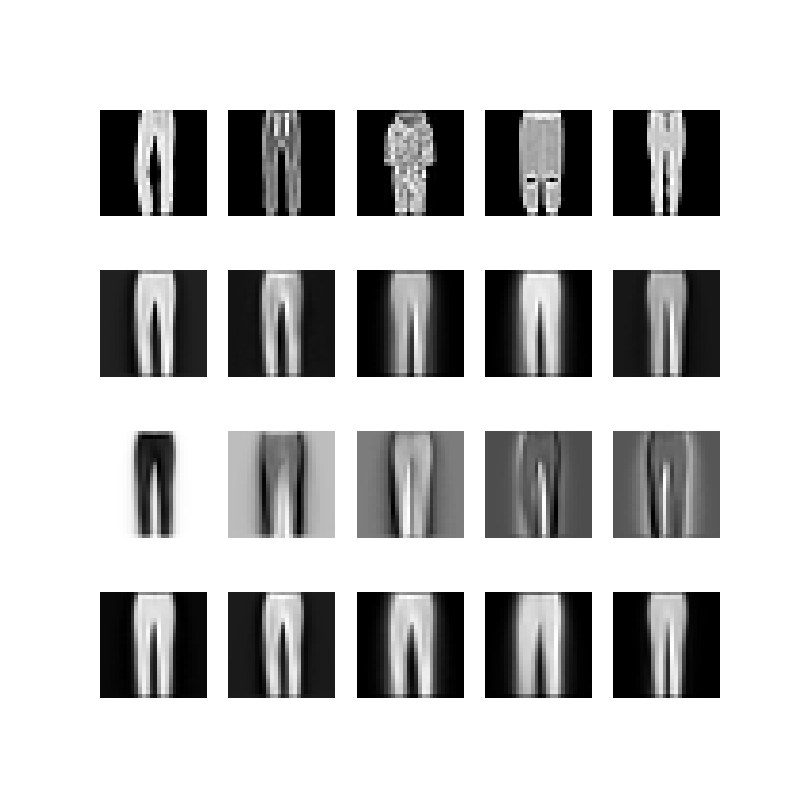}
    \caption{Fashion MNIST, class 1}
    \label{fig:GEN10}
    \vspace*{2mm}
\end{subfigure}
\hfill
\begin{subfigure}{0.4\textwidth}
    \includegraphics[width=1.0\textwidth]{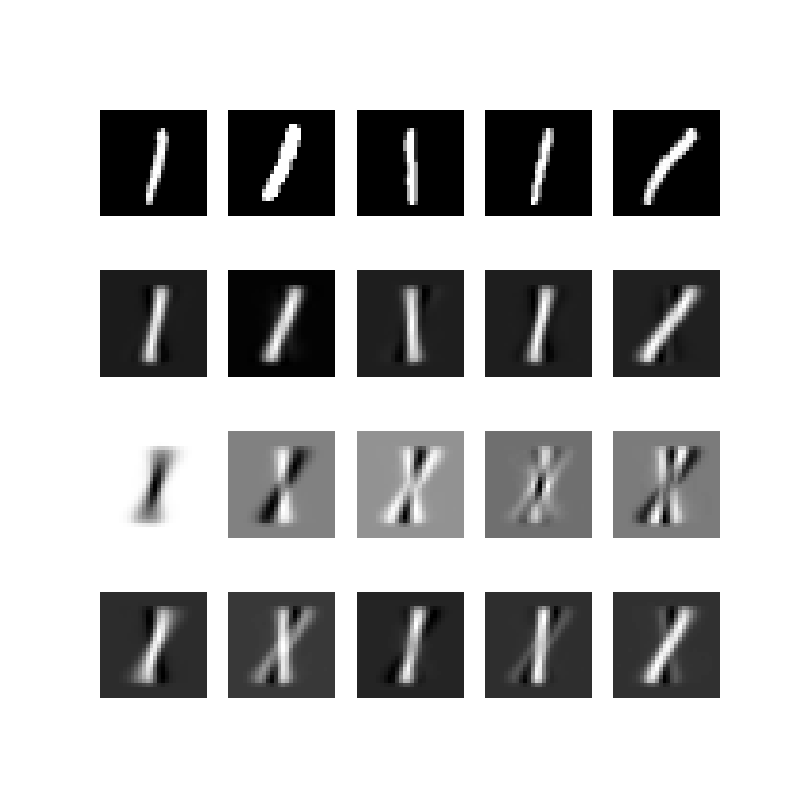}
    \caption{MNIST, class 1}
    \label{fig:GEN11}
    \vspace*{2mm}
\end{subfigure}\\
\begin{subfigure}{.4\textwidth}
    \includegraphics[width=1.0\textwidth]{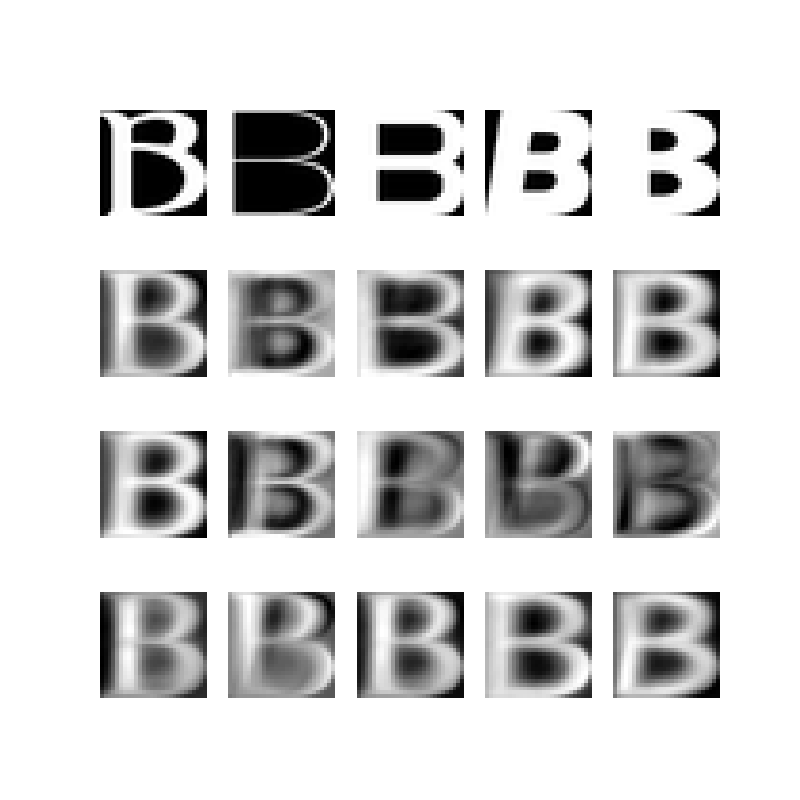}
    \caption{NOT MNIST, class 1}
    \label{fig:GEN12}
    \vspace*{2mm}
\end{subfigure}
\hfill
\begin{subfigure}{.4\textwidth}
    \includegraphics[width=1.0\textwidth]{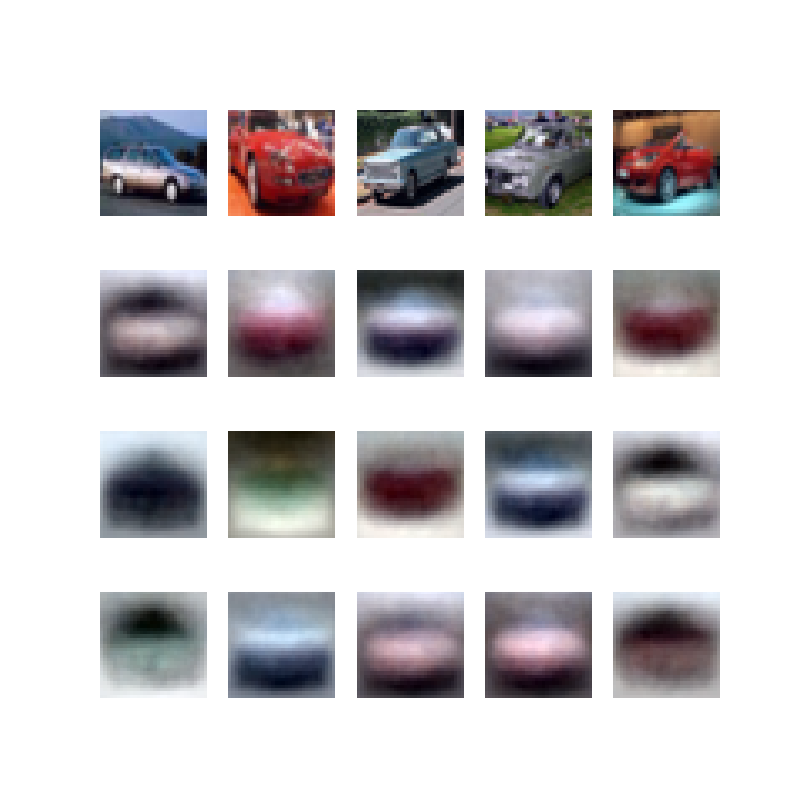}
    \caption{CIFAR10, class 1}
    \label{fig:GEN13}
    \vspace*{2mm}
\end{subfigure}\\
\begin{subfigure}{.4\textwidth}
    \includegraphics[width=1.0\textwidth]{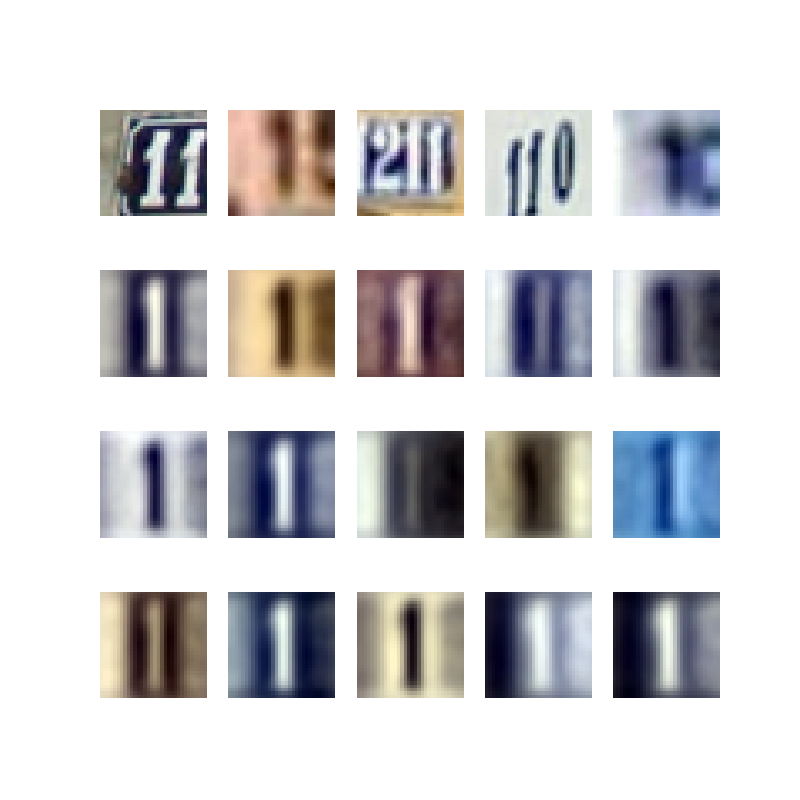}
    \caption{SVHN, class 1}
    \label{fig:GEN14}
    \vspace*{2mm}
\end{subfigure}
\caption{First row: Instances of one class. Second row: Recreation of instances using 5 components. Third row: The biggest 5 components, from big to small. Fourth row: Images synthesized by the lightweight generator, using 5 components.}
\label{fig:GEN1}
\end{figure*}

\begin{figure*}
\centering
\begin{subfigure}{.4\textwidth}
    \includegraphics[width=1.0\textwidth]{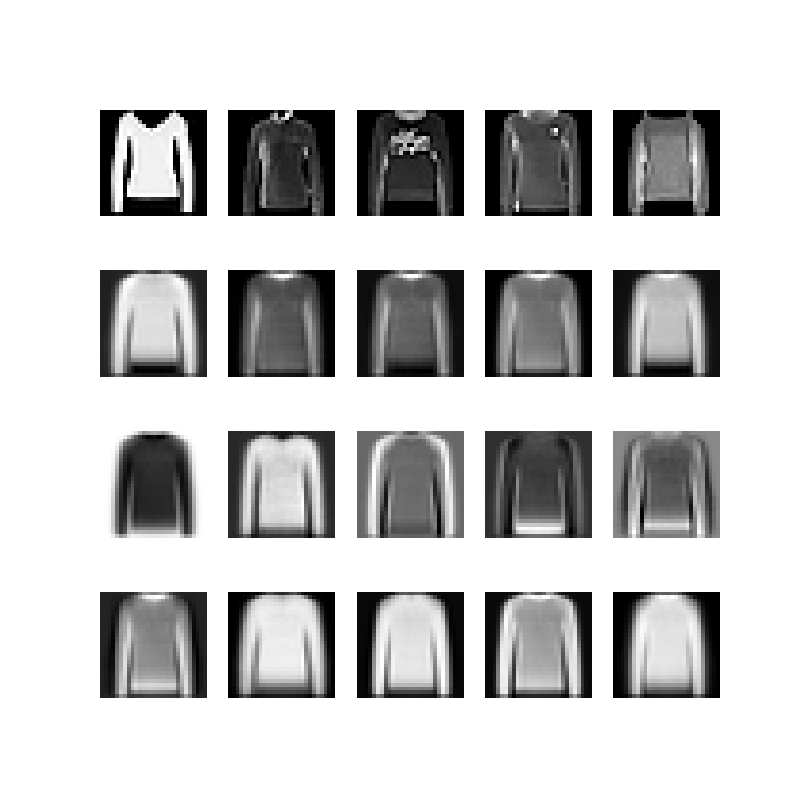}
    \caption{Fashion MNIST, class 2}
    \label{fig:GEN20}
    \vspace*{2mm}
\end{subfigure}
\hfill
\begin{subfigure}{.4\textwidth}
    \includegraphics[width=1.0\textwidth]{figures/Generator/MNIST_class_2_comp_5.png}
    \caption{MNIST, class 2}
    \label{fig:GEN21}
    \vspace*{2mm}
\end{subfigure}\\
\begin{subfigure}{.4\textwidth}
    \includegraphics[width=1.0\textwidth]{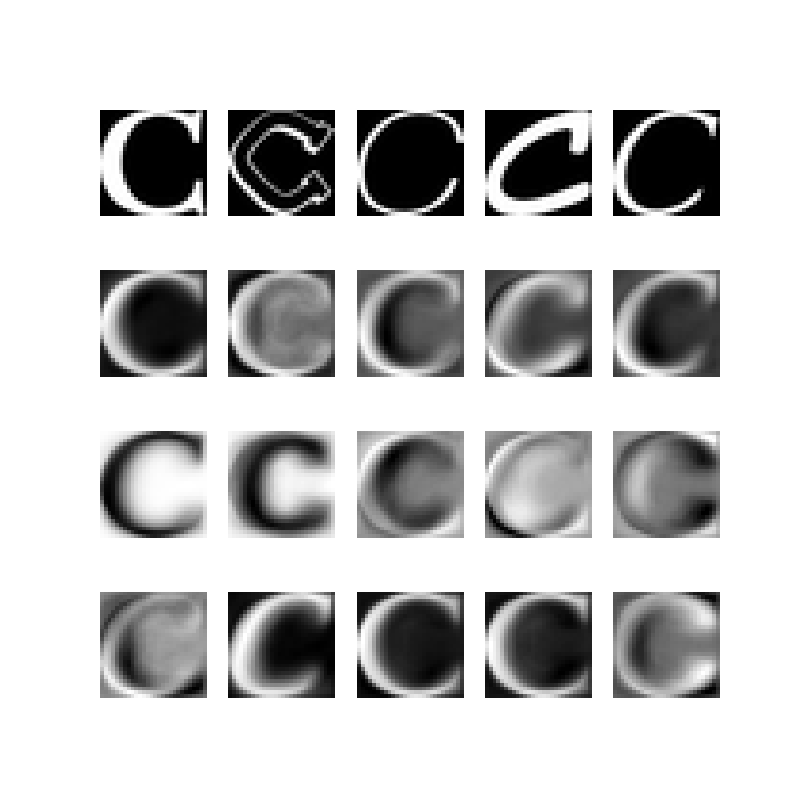}
    \caption{NOT MNIST, class 2}
    \label{fig:GEN22}
    \vspace*{2mm}
\end{subfigure}
\hfill
\begin{subfigure}{.4\textwidth}
    \includegraphics[width=1.0\textwidth]{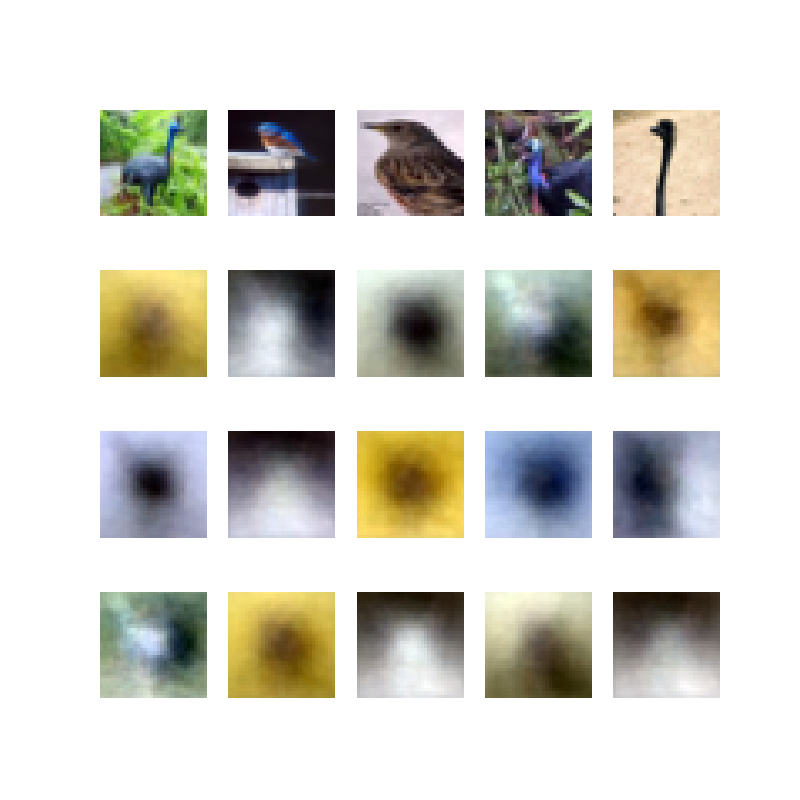}
    \caption{CIFAR10, class 2}
    \label{fig:GEN23}
    \vspace*{2mm}
\end{subfigure}\\
\begin{subfigure}{.4\textwidth}
    \includegraphics[width=1.0\textwidth]{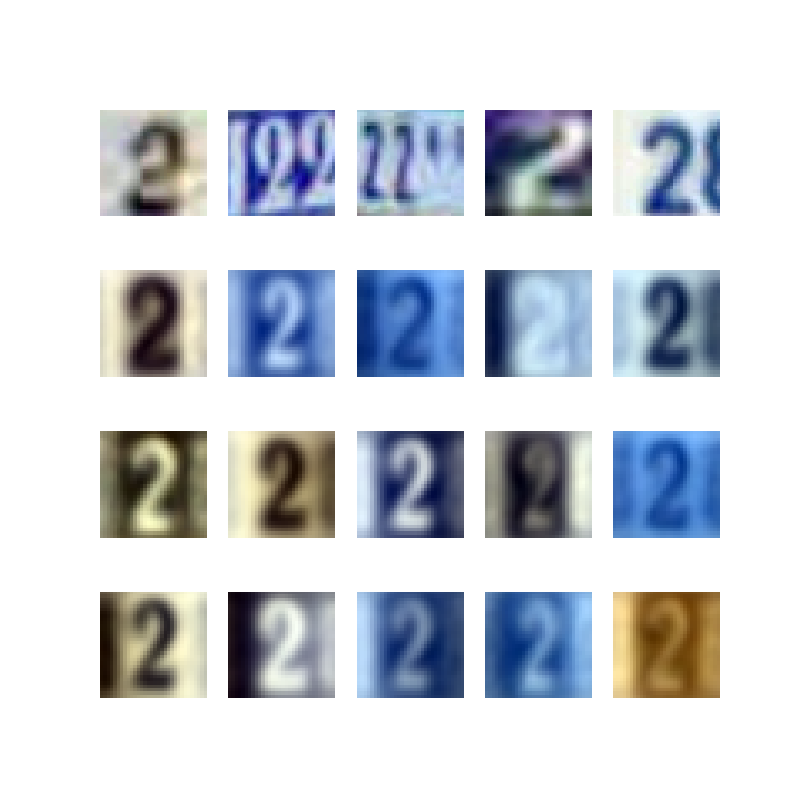}
    \caption{SVHN, class 2}
    \label{fig:GEN24}
    \vspace*{2mm}
\end{subfigure}
\caption{First row: Instances of one class. Second row: Recreation of instances using 5 components. Third row: The biggest 5 components, from big to small. Fourth row: Images synthesized by the lightweight generator, using 5 components.}
\label{fig:GEN2}
\end{figure*}

\begin{figure*}
\centering
\begin{subfigure}{.4\textwidth}
    \includegraphics[width=1.0\textwidth]{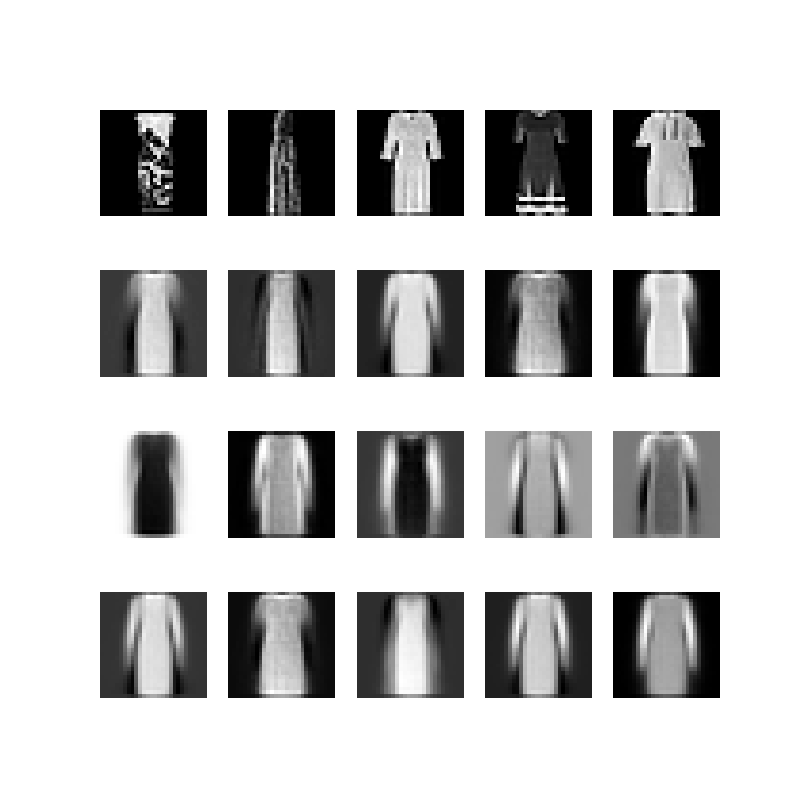}
    \caption{Fashion MNIST, class 3}
    \label{fig:GEN30}
    \vspace*{2mm}
\end{subfigure}
\hfill
\begin{subfigure}{.4\textwidth}
    \includegraphics[width=1.0\textwidth]{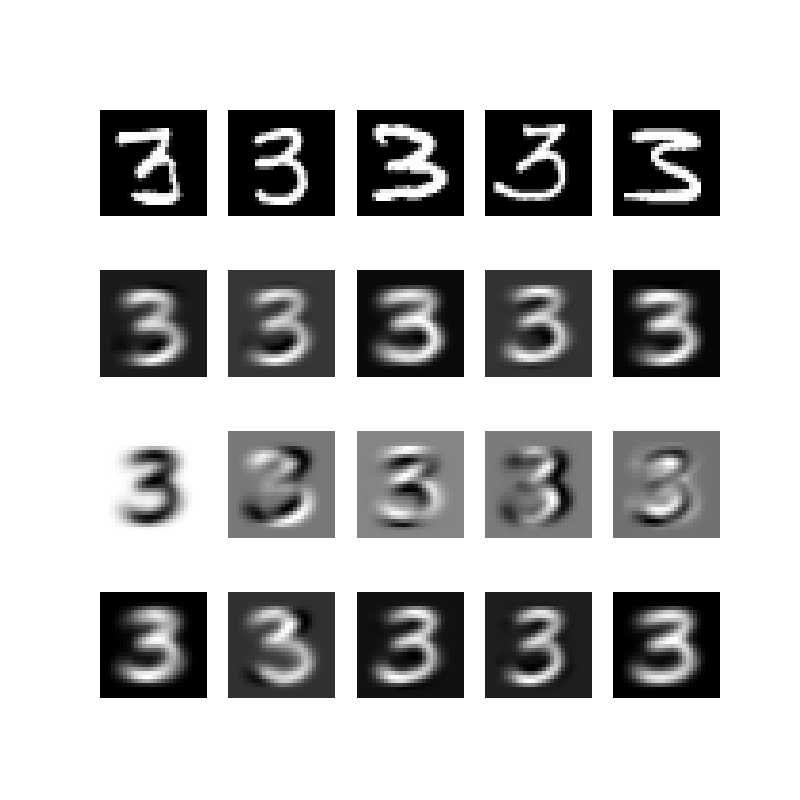}
    \caption{MNIST, class 3}
    \label{fig:GEN31}
    \vspace*{2mm}
\end{subfigure}\\
\begin{subfigure}{.4\textwidth}
    \includegraphics[width=1.0\textwidth]{figures/Generator/NOT_MNIST_class_3_comp_5.png}
    \caption{NOT MNIST, class 3}
    \label{fig:GEN32}
    \vspace*{2mm}
\end{subfigure}
\hfill
\begin{subfigure}{.4\textwidth}
    \includegraphics[width=1.0\textwidth]{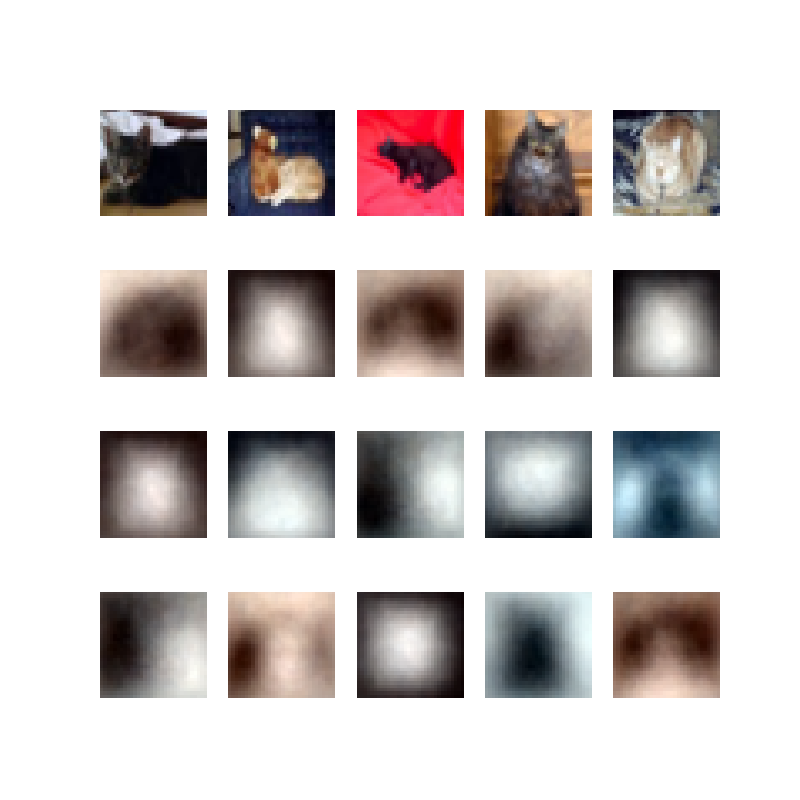}
    \caption{CIFAR10, class 3}
    \label{fig:GEN33}
    \vspace*{2mm}
\end{subfigure}\\
\begin{subfigure}{.4\textwidth}
    \includegraphics[width=1.0\textwidth]{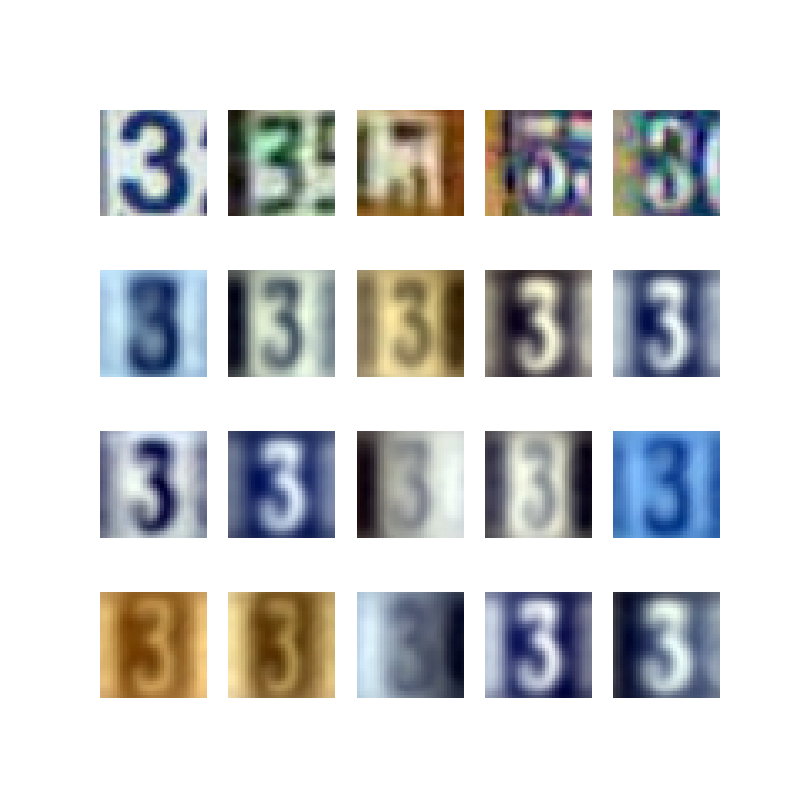}
    \caption{SVHN, class 3}
    \label{fig:GEN34}
    \vspace*{2mm}
\end{subfigure}
\caption{First row: Instances of one class. Second row: Recreation of instances using 5 components. Third row: The biggest 5 components, from big to small. Fourth row: Images synthesized by the lightweight generator, using 5 components.}
\label{fig:GEN3}
\end{figure*}

\begin{figure*}
\centering
\begin{subfigure}{.4\textwidth}
    \includegraphics[width=1.0\textwidth]{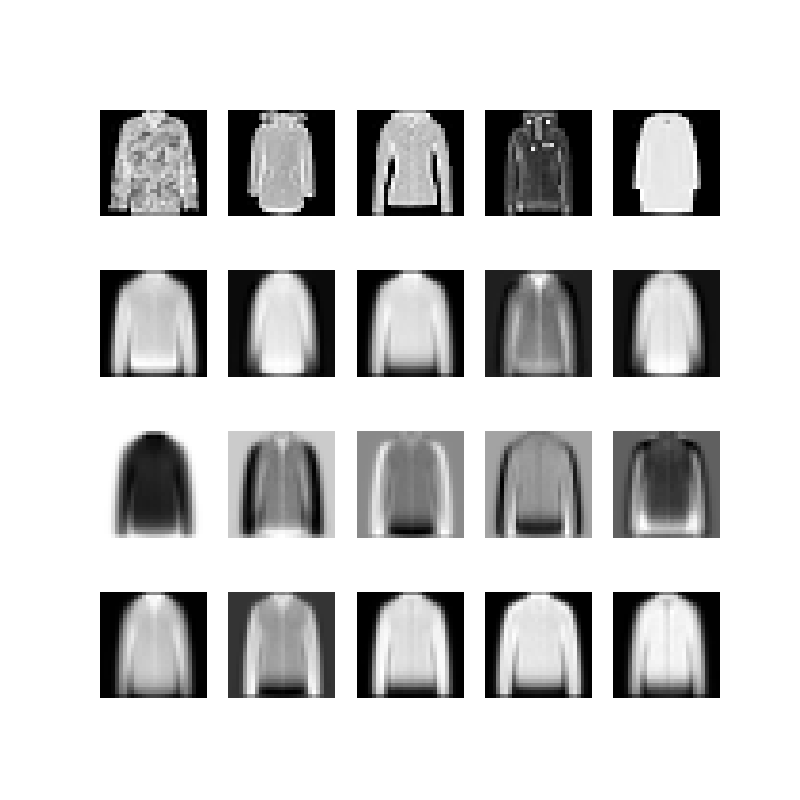}
    \caption{Fashion MNIST, class 4}
    \label{fig:GEN40}
    \vspace*{2mm}
\end{subfigure}
\hfill
\begin{subfigure}{.4\textwidth}
    \includegraphics[width=1.0\textwidth]{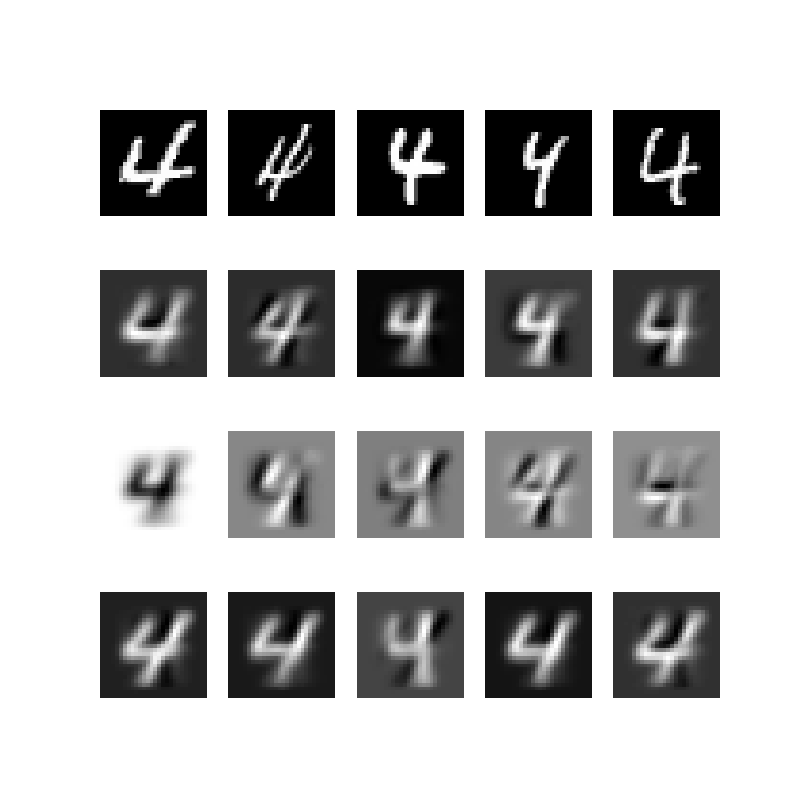}
    \caption{MNIST, class 4}
    \label{fig:GEN41}
    \vspace*{2mm}
\end{subfigure}\\
\begin{subfigure}{.4\textwidth}
    \includegraphics[width=1.0\textwidth]{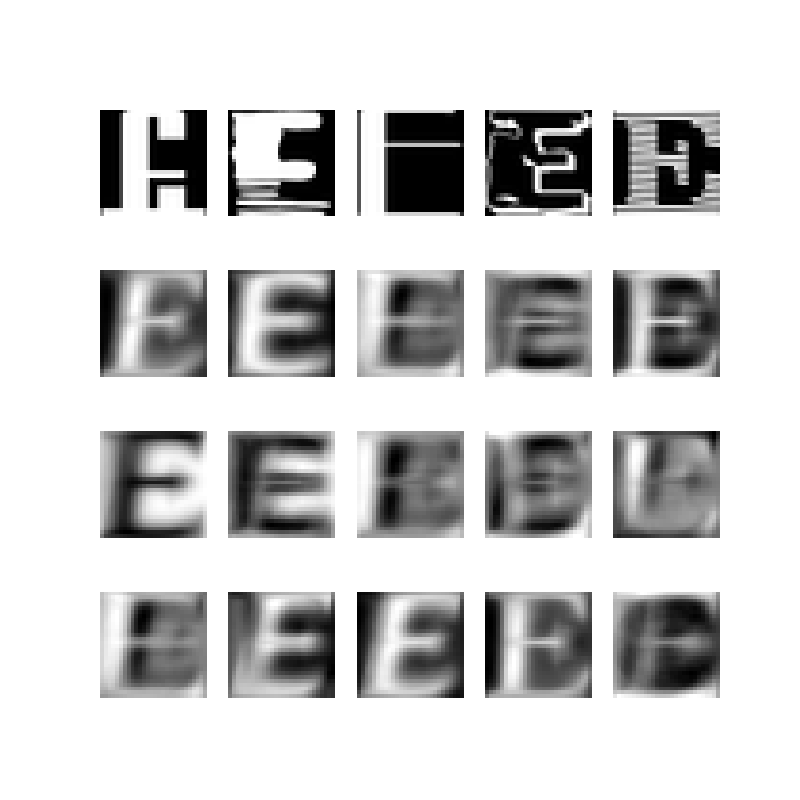}
    \caption{NOT MNIST, class 4}
    \label{fig:GEN42}
    \vspace*{2mm}
\end{subfigure}
\hfill
\begin{subfigure}{.4\textwidth}
    \includegraphics[width=1.0\textwidth]{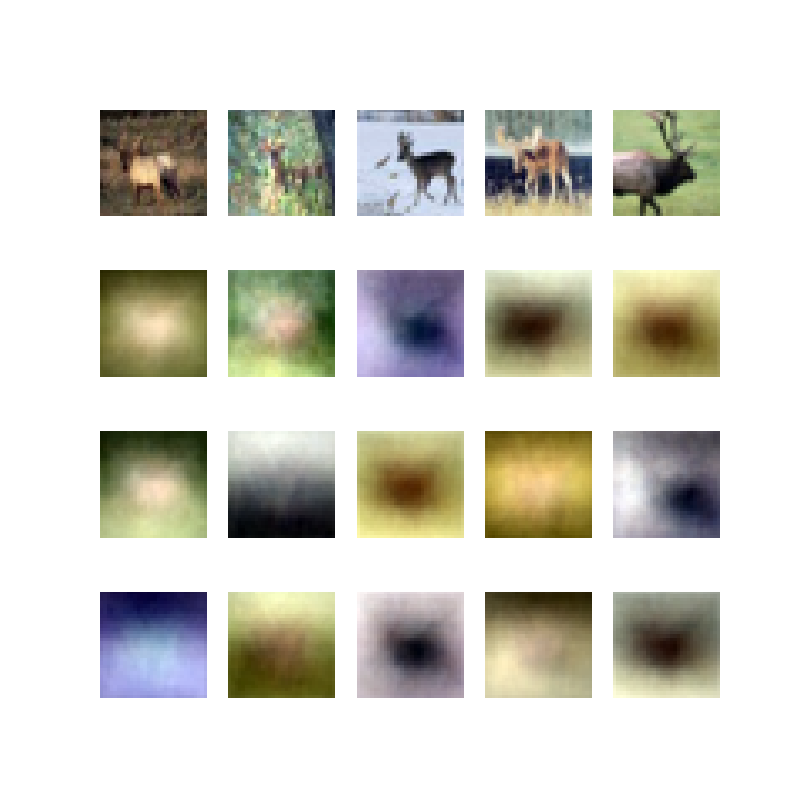}
    \caption{CIFAR10, class 4}
    \label{fig:GEN43}
    \vspace*{2mm}
\end{subfigure}\\
\begin{subfigure}{.4\textwidth}
    \includegraphics[width=1.0\textwidth]{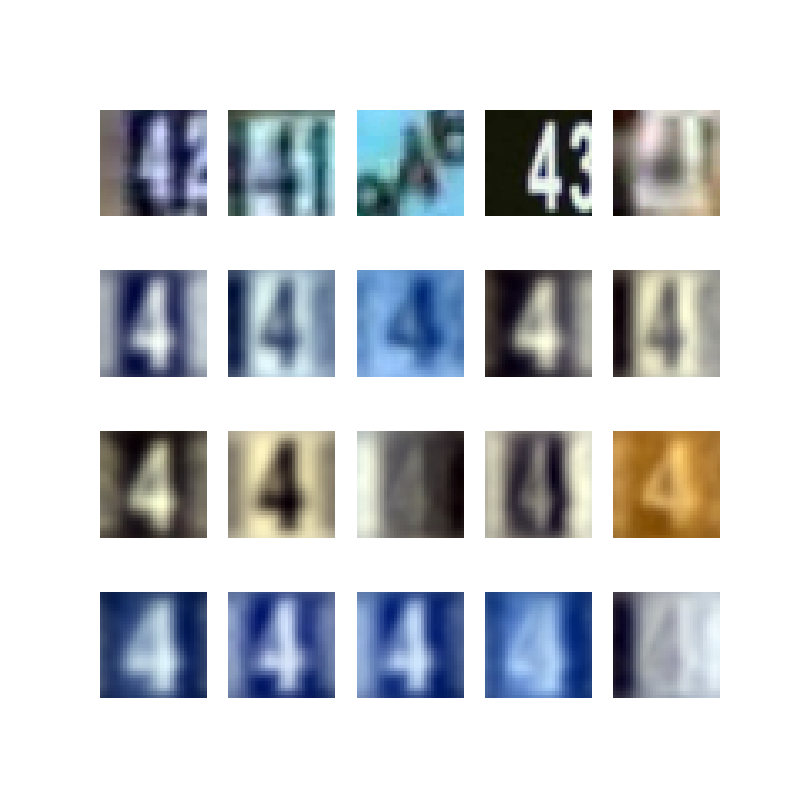}
    \caption{SVHN, class 4}
    \label{fig:GEN44}
    \vspace*{2mm}
\end{subfigure}
\caption{First row: Instances of one class. Second row: Recreation of instances using 5 components. Third row: The biggest 5 components, from big to small. Fourth row: Images synthesized by the lightweight generator, using 5 components.}
\label{fig:GEN4}
\end{figure*}

\begin{figure*}
\centering
\begin{subfigure}{.4\textwidth}
    \includegraphics[width=1.0\textwidth]{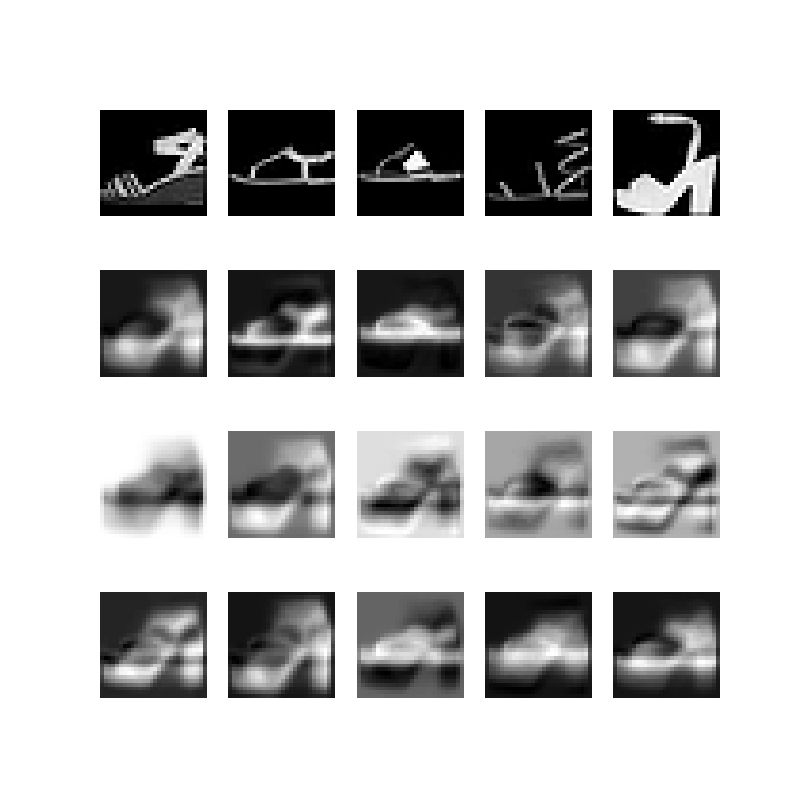}
    \caption{Fashion MNIST, class 5}
    \label{fig:GEN50}
    \vspace*{2mm}
\end{subfigure}
\hfill
\begin{subfigure}{.4\textwidth}
    \includegraphics[width=1.0\textwidth]{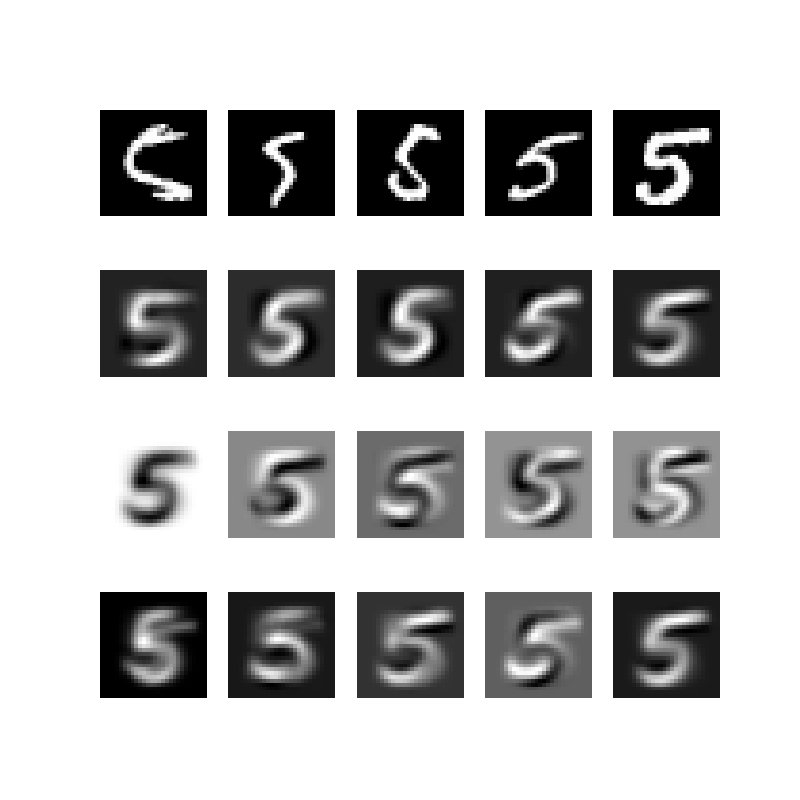}
    \caption{MNIST, class 5}
    \label{fig:GEN51}
    \vspace*{2mm}
\end{subfigure}\\
\begin{subfigure}{.4\textwidth}
    \includegraphics[width=1.0\textwidth]{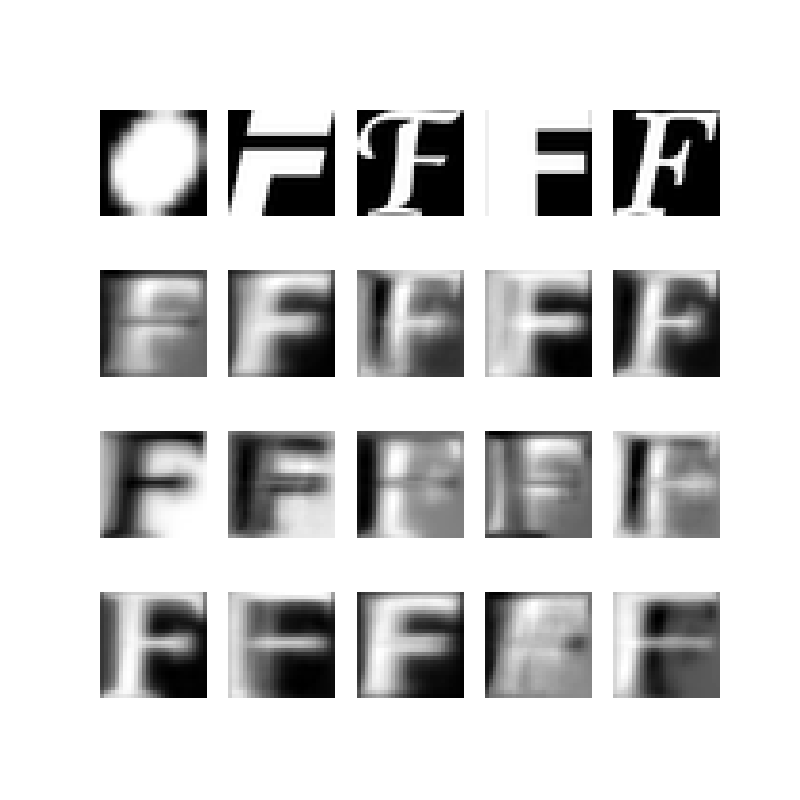}
    \caption{NOT MNIST, class 5}
    \label{fig:GEN52}
    \vspace*{2mm}
\end{subfigure}
\hfill
\begin{subfigure}{.4\textwidth}
    \includegraphics[width=1.0\textwidth]{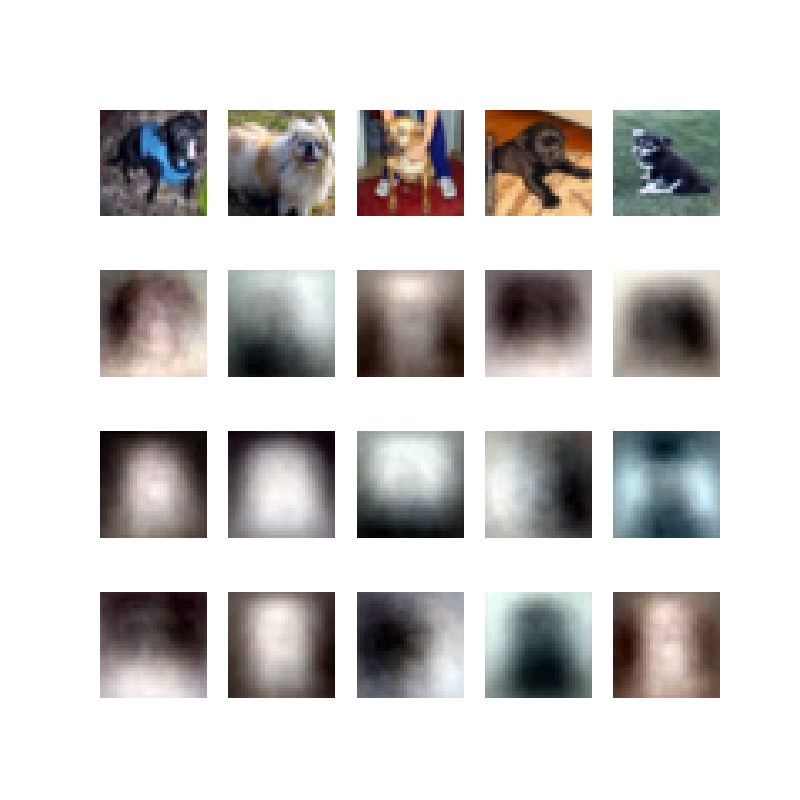}
    \caption{CIFAR10, class 5}
    \label{fig:GEN53}
    \vspace*{2mm}
\end{subfigure}\\
\begin{subfigure}{.4\textwidth}
    \includegraphics[width=1.0\textwidth]{figures/Generator/SVHN_class_5_comp_5.png}
    \caption{SVHN, class 5}
    \label{fig:GEN54}
    \vspace*{2mm}
\end{subfigure}
\caption{First row: Instances of one class. Second row: Recreation of instances using 5 components. Third row: The biggest 5 components, from big to small. Fourth row: Images synthesized by the lightweight generator, using 5 components.}
\label{fig:GEN5}
\end{figure*}

\begin{figure*}
\centering
\begin{subfigure}{.4\textwidth}
    \includegraphics[width=1.0\textwidth]{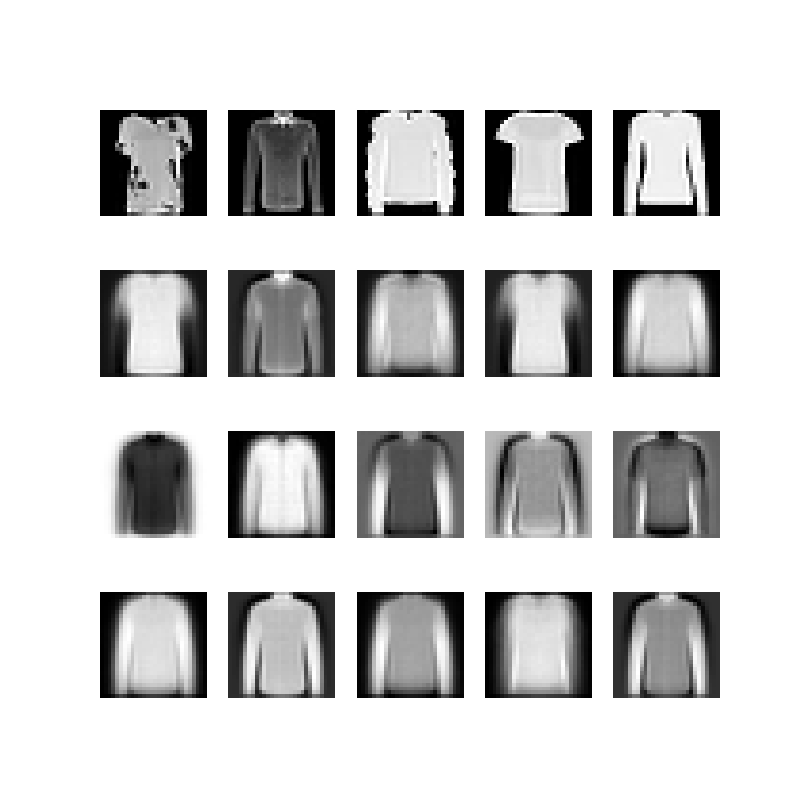}
    \caption{Fashion MNIST, class 6}
    \label{fig:GEN60}
    \vspace*{2mm}
\end{subfigure}
\hfill
\begin{subfigure}{.4\textwidth}
    \includegraphics[width=1.0\textwidth]{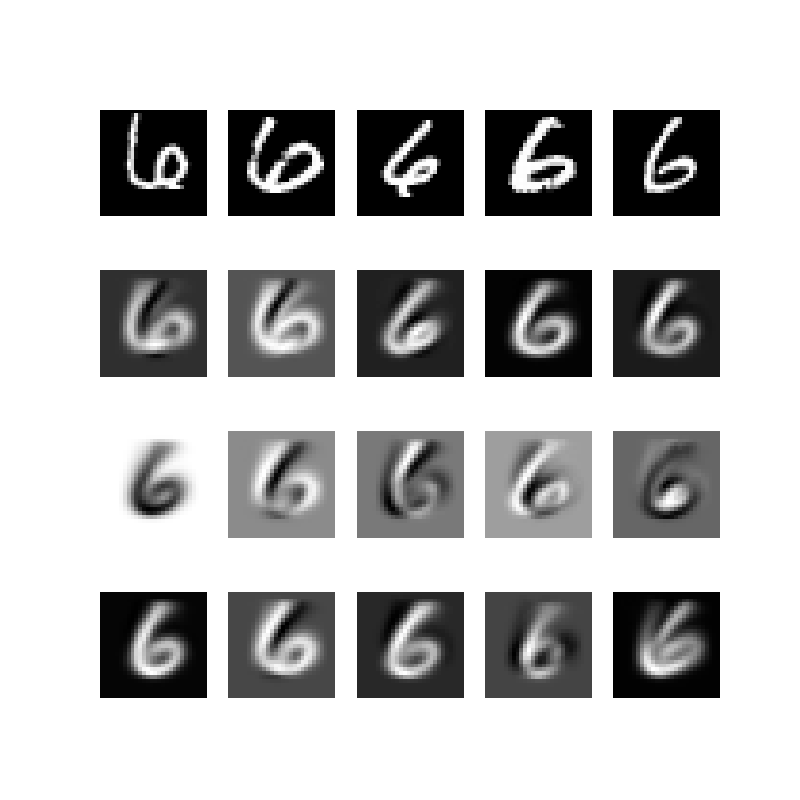}
    \caption{MNIST, class 6}
    \label{fig:GEN61}
    \vspace*{2mm}
\end{subfigure}\\
\begin{subfigure}{.4\textwidth}
    \includegraphics[width=1.0\textwidth]{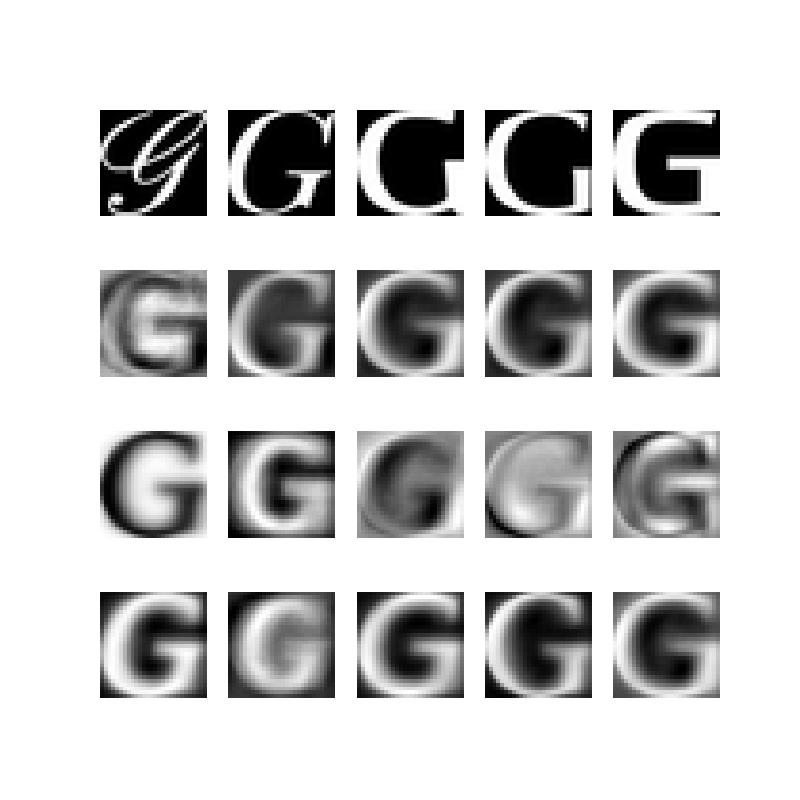}
    \caption{NOT MNIST, class 6}
    \label{fig:GEN62}
    \vspace*{2mm}
\end{subfigure}
\hfill
\begin{subfigure}{.4\textwidth}
    \includegraphics[width=1.0\textwidth]{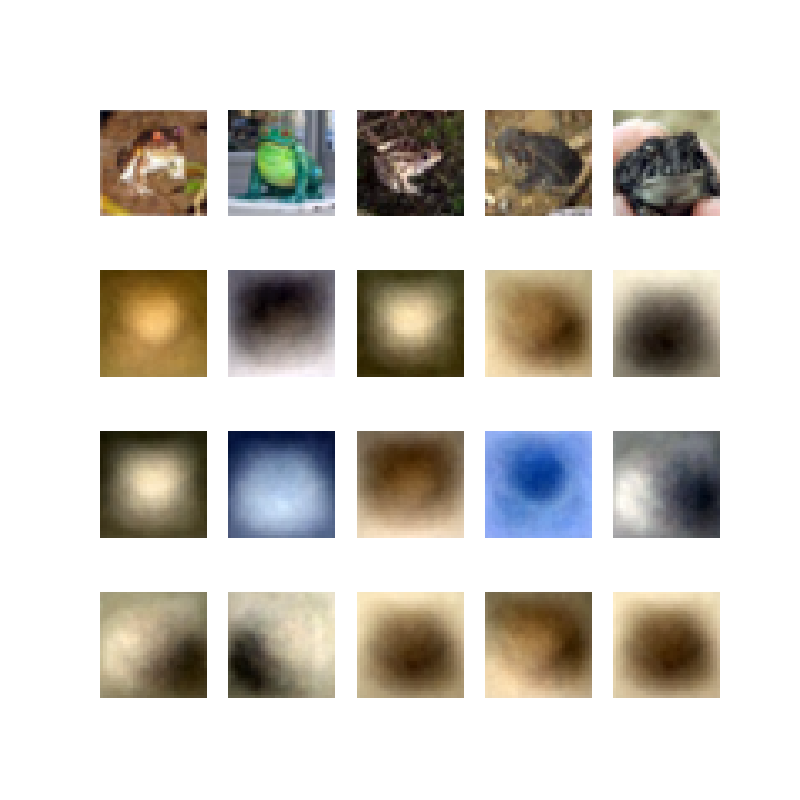}
    \caption{CIFAR10, class 6}
    \label{fig:GEN63}
    \vspace*{2mm}
\end{subfigure}\\
\begin{subfigure}{.4\textwidth}
    \includegraphics[width=1.0\textwidth]{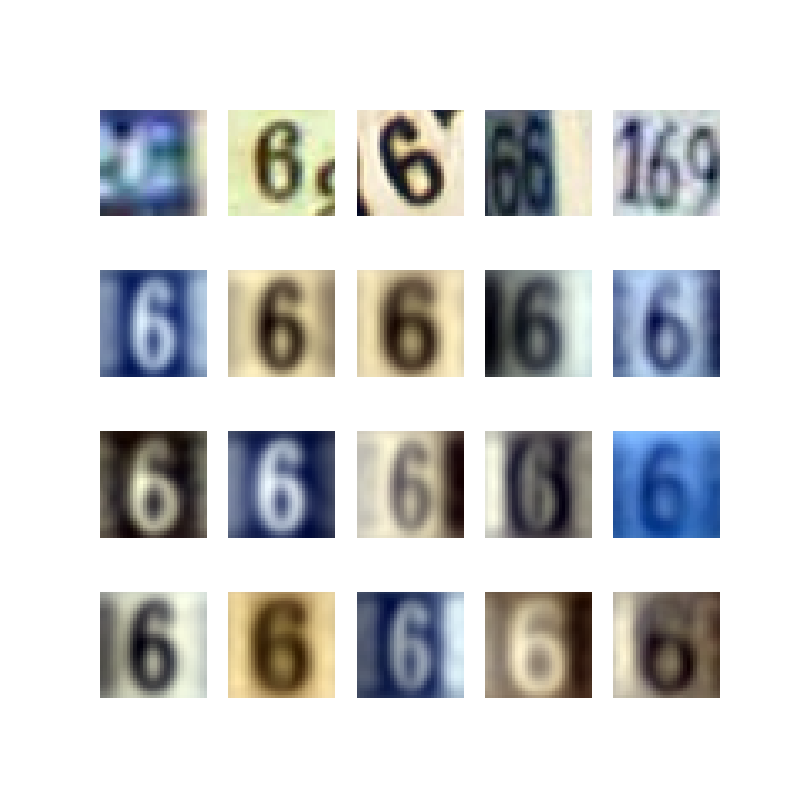}
    \caption{SVHN, class 6}
    \label{fig:GEN64}
    \vspace*{2mm}
\end{subfigure}
\caption{First row: Instances of one class. Second row: Recreation of instances using 5 components. Third row: The biggest 5 components, from big to small. Fourth row: Images synthesized by the lightweight generator, using 5 components.}
\label{fig:GEN6}
\end{figure*}

\begin{figure*}
\centering
\begin{subfigure}{.4\textwidth}
    \includegraphics[width=1.0\textwidth]{figures/Generator/Fashion_MNIST_class_7_comp_5.png}
    \caption{Fashion MNIST, class 7}
    \label{fig:GEN70}
    \vspace*{2mm}
\end{subfigure}
\hfill
\begin{subfigure}{.4\textwidth}
    \includegraphics[width=1.0\textwidth]{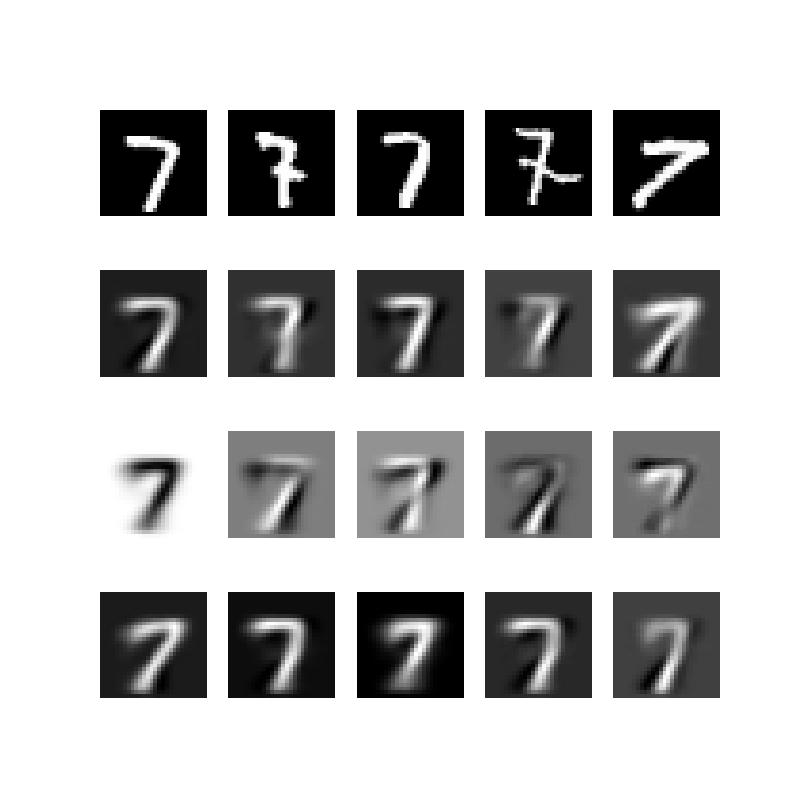}
    \caption{MNIST, class 7}
    \label{fig:GEN71}
    \vspace*{2mm}
\end{subfigure}\\
\begin{subfigure}{.4\textwidth}
    \includegraphics[width=1.0\textwidth]{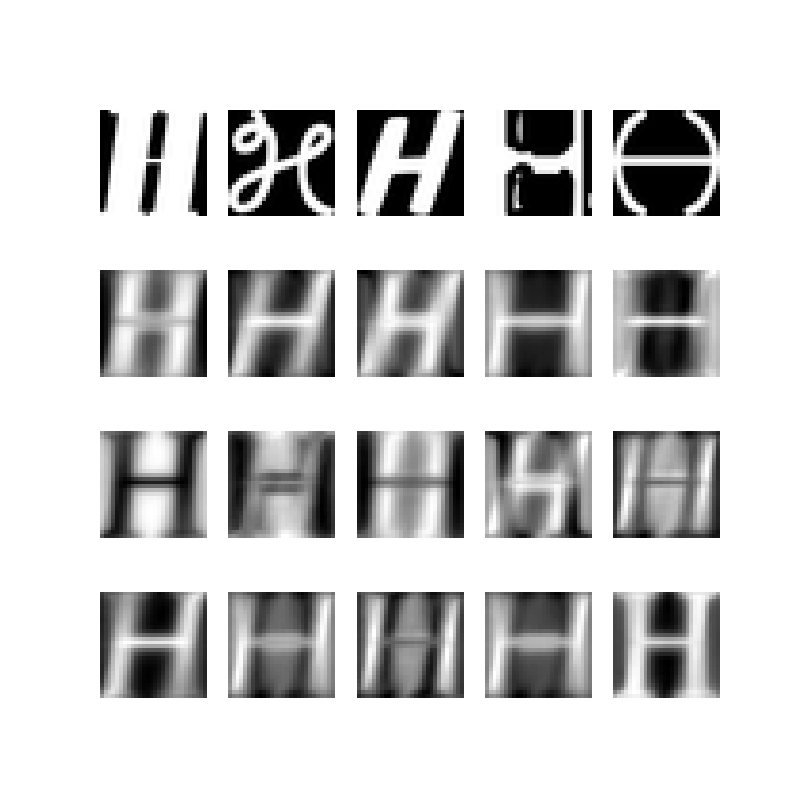}
    \caption{NOT MNIST, class 7}
    \label{fig:GEN72}
    \vspace*{2mm}
\end{subfigure}
\hfill
\begin{subfigure}{.4\textwidth}
    \includegraphics[width=1.0\textwidth]{figures/Generator/CIFAR10_class_7_comp_5.png}
    \caption{CIFAR10, class 7}
    \label{fig:GEN73}
    \vspace*{2mm}
\end{subfigure}\\
\begin{subfigure}{.4\textwidth}
    \includegraphics[width=1.0\textwidth]{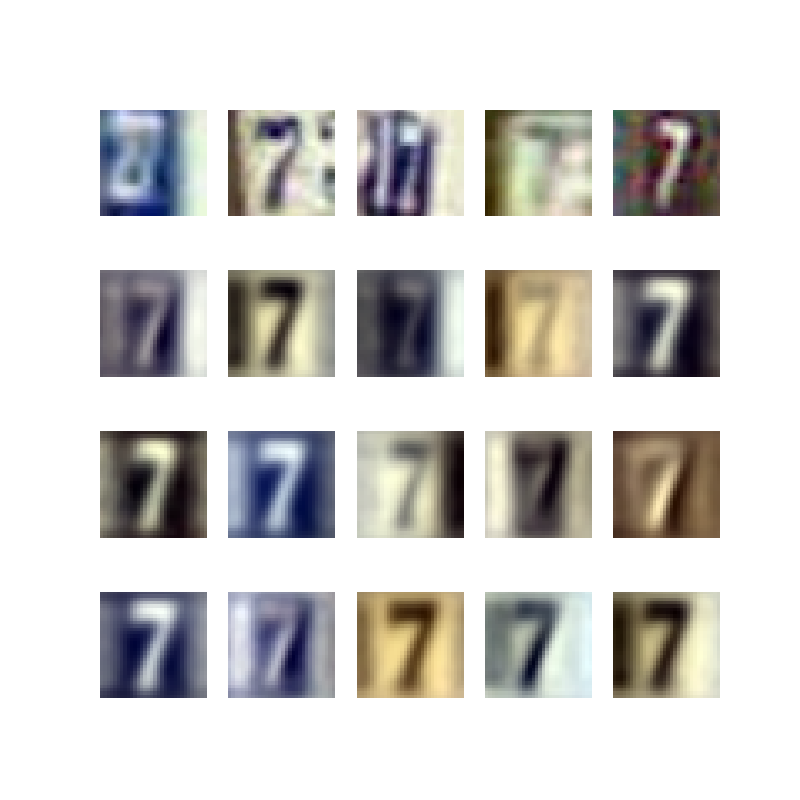}
    \caption{SVHN, class 7}
    \label{fig:GEN74}
    \vspace*{2mm}
\end{subfigure}
\caption{First row: Instances of one class. Second row: Recreation of instances using 5 components. Third row: The biggest 5 components, from big to small. Fourth row: Images synthesized by the lightweight generator, using 5 components.}
\label{fig:GEN7}
\end{figure*}

\begin{figure*}
\centering
\begin{subfigure}{.4\textwidth}
    \includegraphics[width=1.0\textwidth]{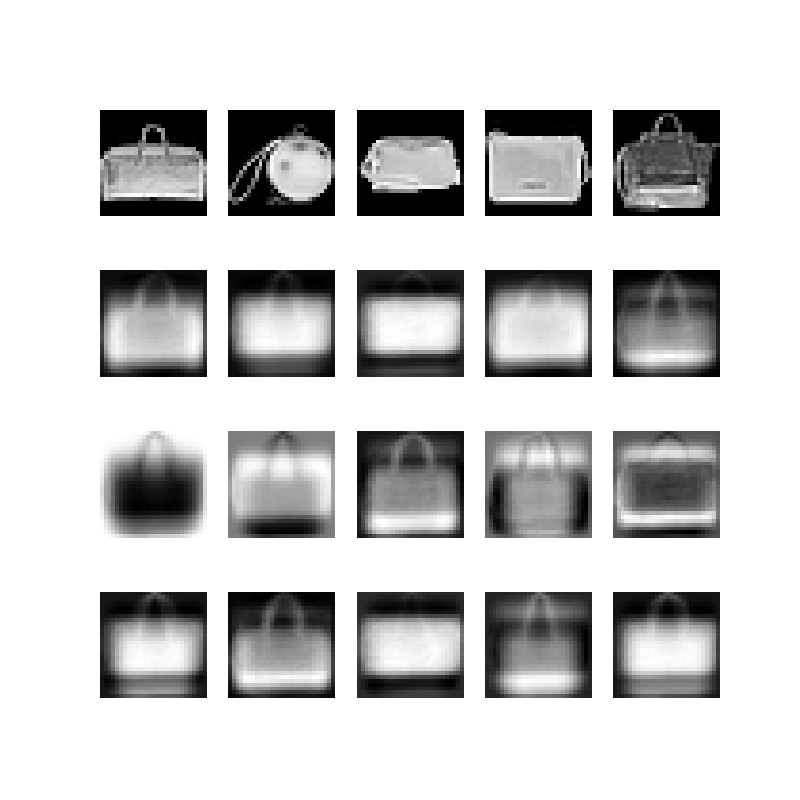}
    \caption{Fashion MNIST, class 8}
    \label{fig:GEN80}
    \vspace*{2mm}
\end{subfigure}
\hfill
\begin{subfigure}{.4\textwidth}
    \includegraphics[width=1.0\textwidth]{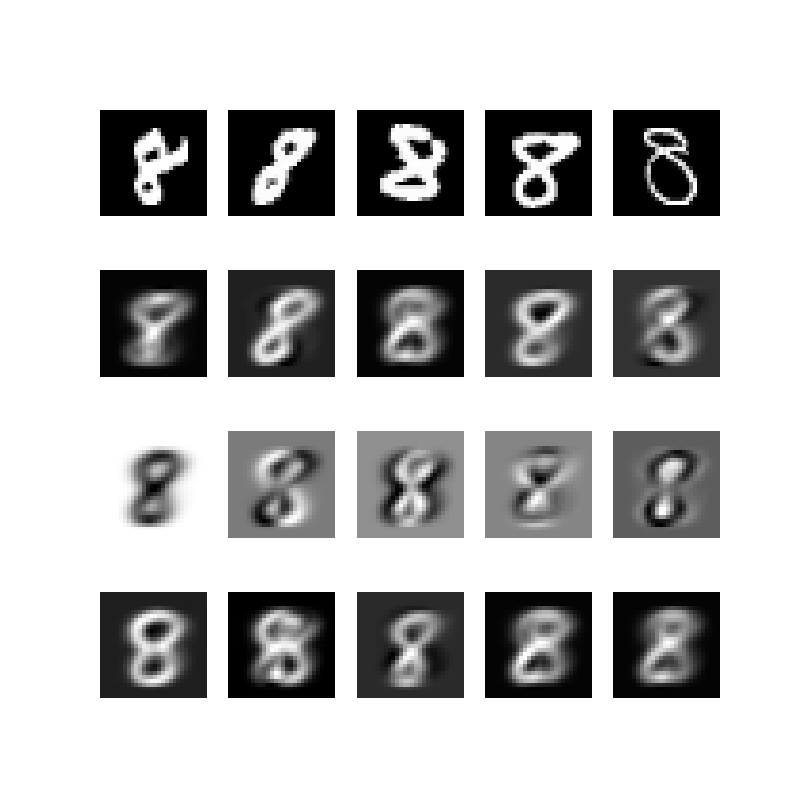}
    \caption{MNIST, class 8}
    \label{fig:GEN81}
    \vspace*{2mm}
\end{subfigure}\\
\begin{subfigure}{.4\textwidth}
    \includegraphics[width=1.0\textwidth]{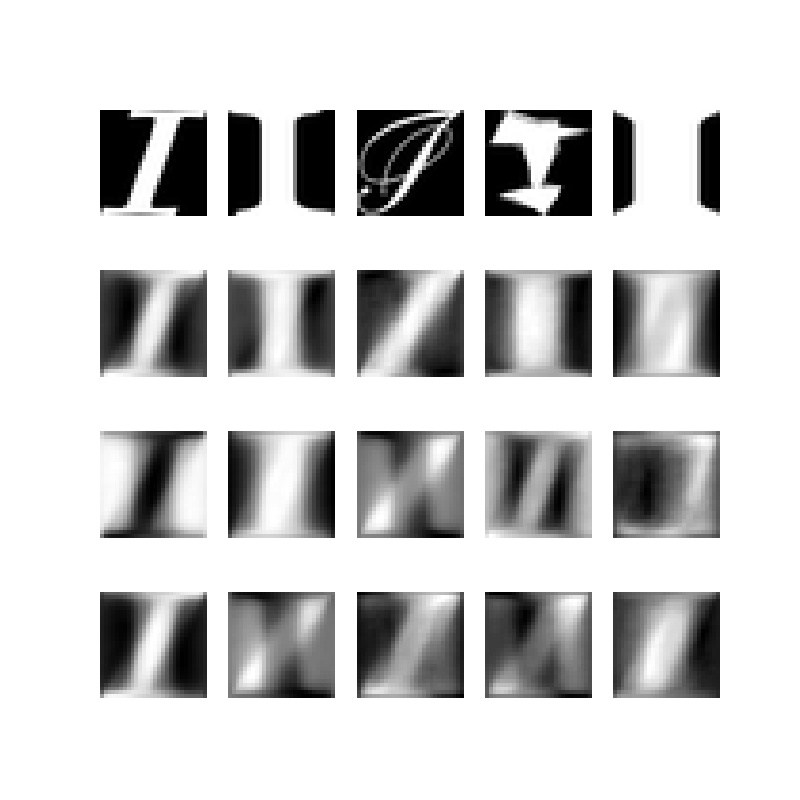}
    \caption{NOT MNIST, class 8}
    \label{fig:GEN82}
    \vspace*{2mm}
\end{subfigure}
\hfill
\begin{subfigure}{.4\textwidth}
    \includegraphics[width=1.0\textwidth]{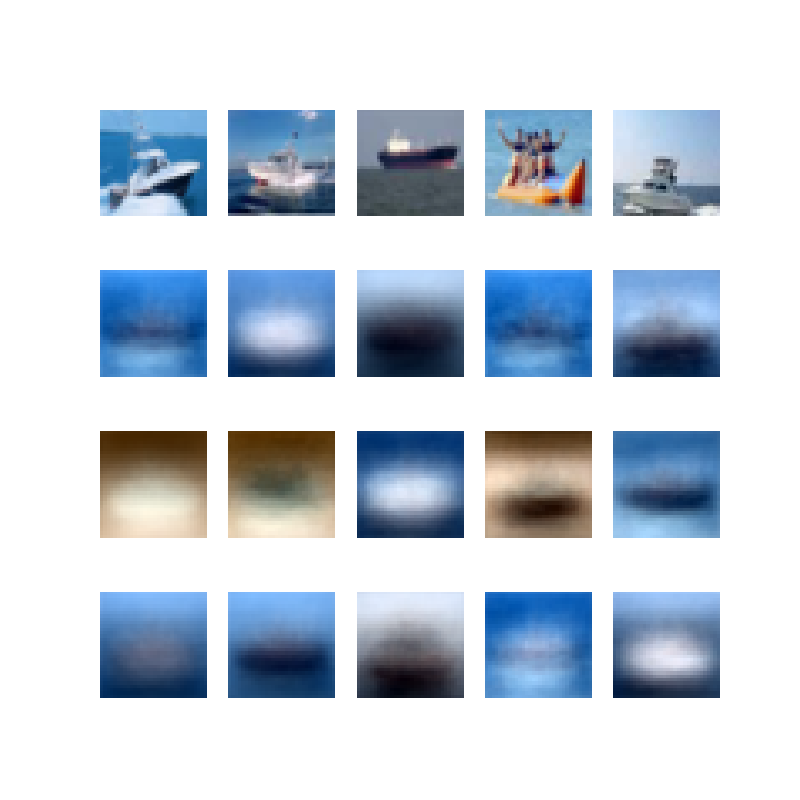}
    \caption{CIFAR10, class 8}
    \label{fig:GEN83}
    \vspace*{2mm}
\end{subfigure}\\
\begin{subfigure}{.4\textwidth}
    \includegraphics[width=1.0\textwidth]{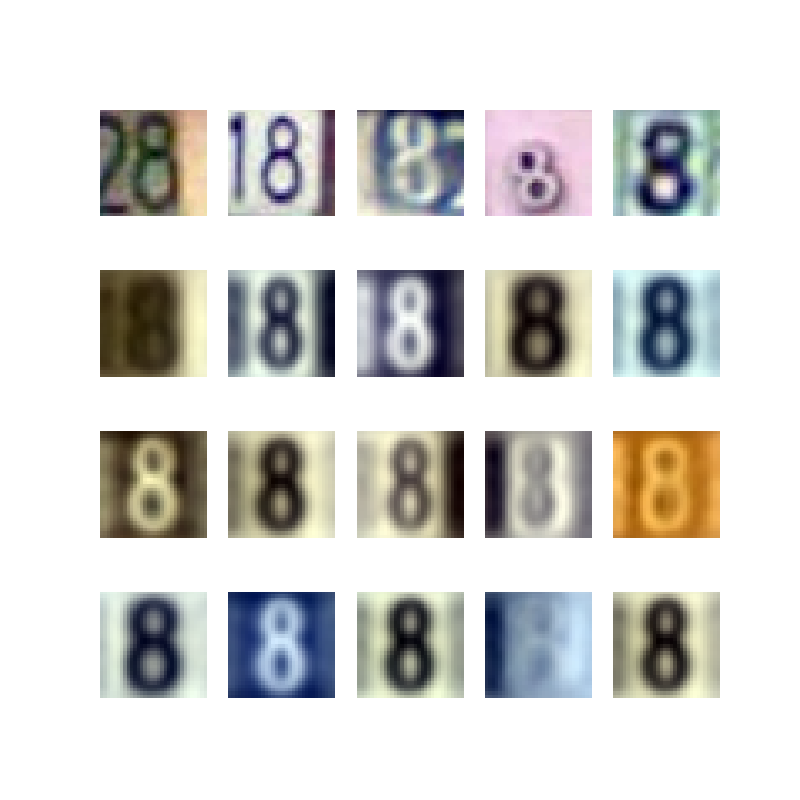}
    \caption{SVHN, class 8}
    \label{fig:GEN84}
    \vspace*{2mm}
\end{subfigure}
\caption{First row: Instances of one class. Second row: Recreation of instances using 5 components. Third row: The biggest 5 components, from big to small. Fourth row: Images synthesized by the lightweight generator, using 5 components.}
\label{fig:GEN8}
\end{figure*}

\begin{figure*}
\centering
\begin{subfigure}{.4\textwidth}
    \includegraphics[width=1.0\textwidth]{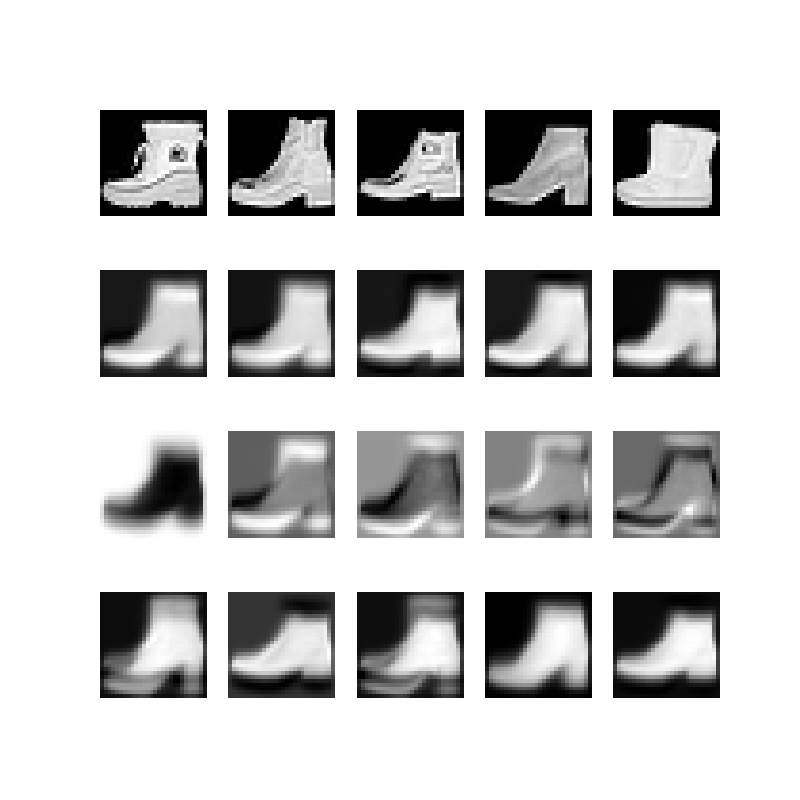}
    \caption{Fashion MNIST, class 9}
    \label{fig:GEN90}
    \vspace*{2mm}
\end{subfigure}
\hfill
\begin{subfigure}{.4\textwidth}
    \includegraphics[width=1.0\textwidth]{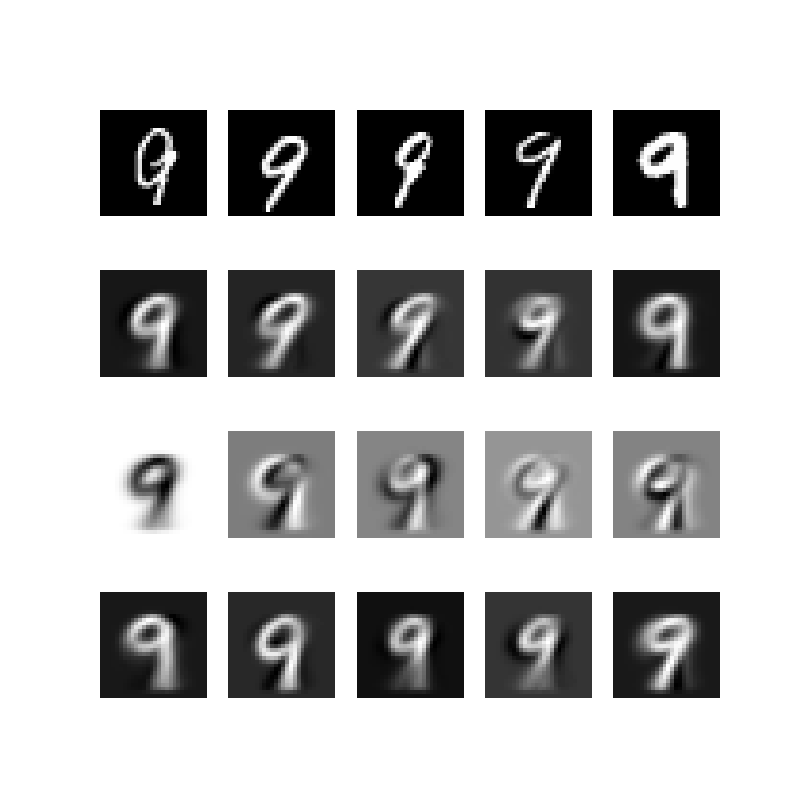}
    \caption{MNIST, class 9}
    \label{fig:GEN91}
    \vspace*{2mm}
\end{subfigure}\\
\begin{subfigure}{.4\textwidth}
    \includegraphics[width=1.0\textwidth]{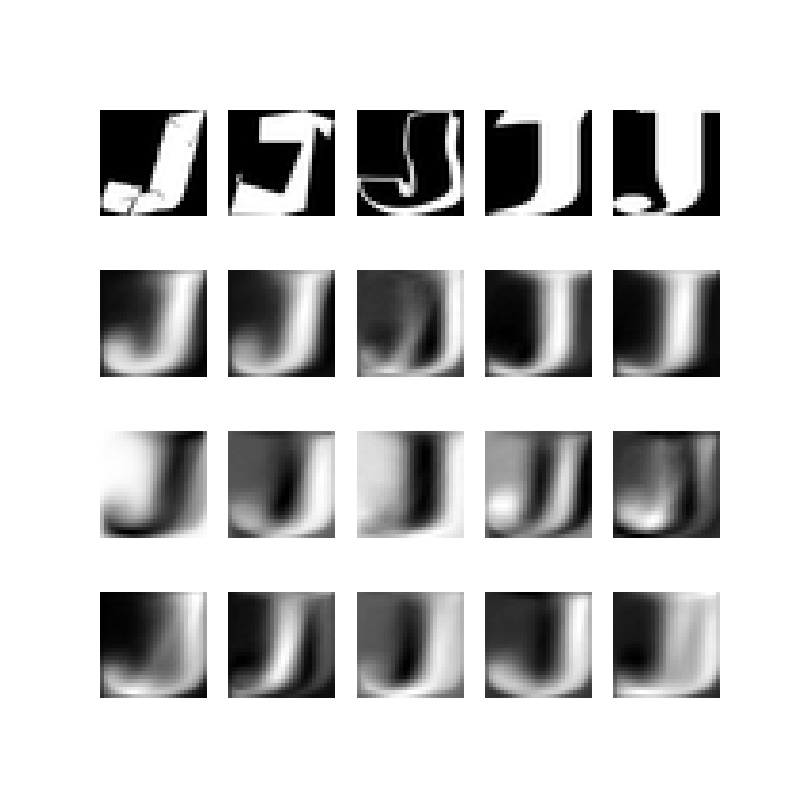}
    \caption{NOT MNIST, class 9}
    \label{fig:GEN92}
    \vspace*{2mm}
\end{subfigure}
\hfill
\begin{subfigure}{.4\textwidth}
    \includegraphics[width=1.0\textwidth]{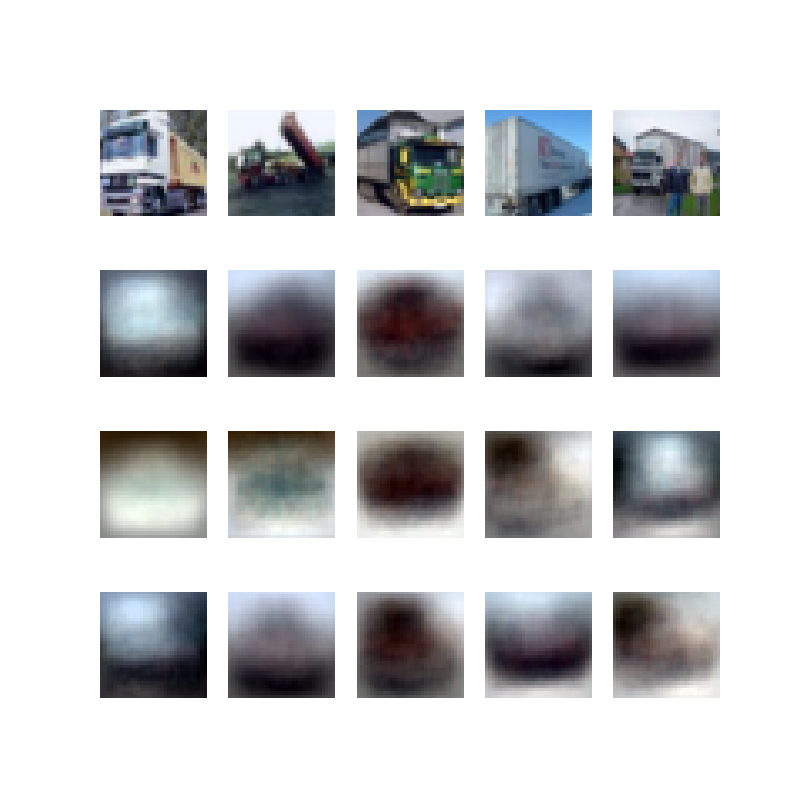}
    \caption{CIFAR10, class 9}
    \label{fig:GEN93}
    \vspace*{2mm}
\end{subfigure}\\
\begin{subfigure}{.4\textwidth}
    \includegraphics[width=1.0\textwidth]{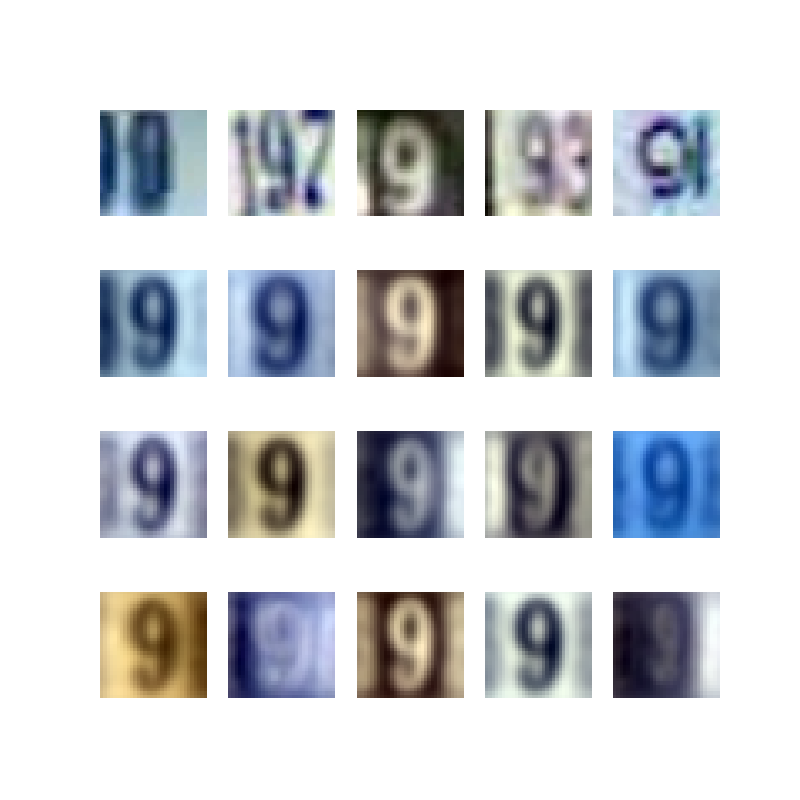}
    \caption{SVHN, class 9}
    \label{fig:GEN94}
    \vspace*{2mm}
\end{subfigure}
\caption{First row: Instances of one class. Second row: Recreation of instances using 5 components. Third row: The biggest 5 components, from big to small. Fourth row: Images synthesized by the lightweight generator, using 5 components.}
\label{fig:GEN9}
\end{figure*}

\begin{figure*}
\centering
\begin{subfigure}{.4\textwidth}
    \includegraphics[width=1.0\textwidth]{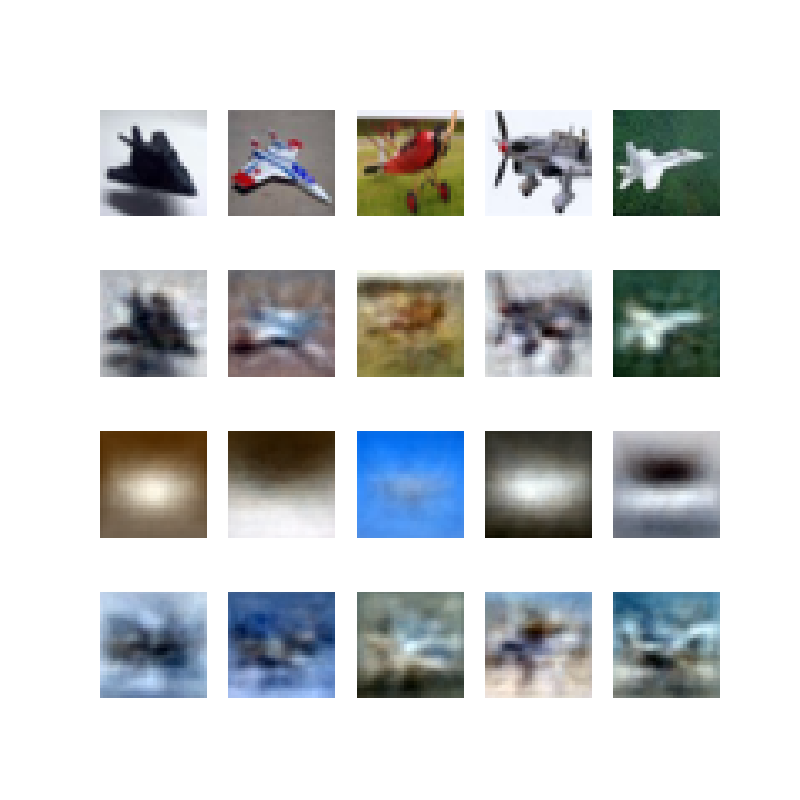}
    \caption{CIFAR10, class 0}
    \label{fig:CIFAR10_80_0}
    \vspace*{2mm}
\end{subfigure}
\hfill
\begin{subfigure}{.4\textwidth}
    \includegraphics[width=1.0\textwidth]{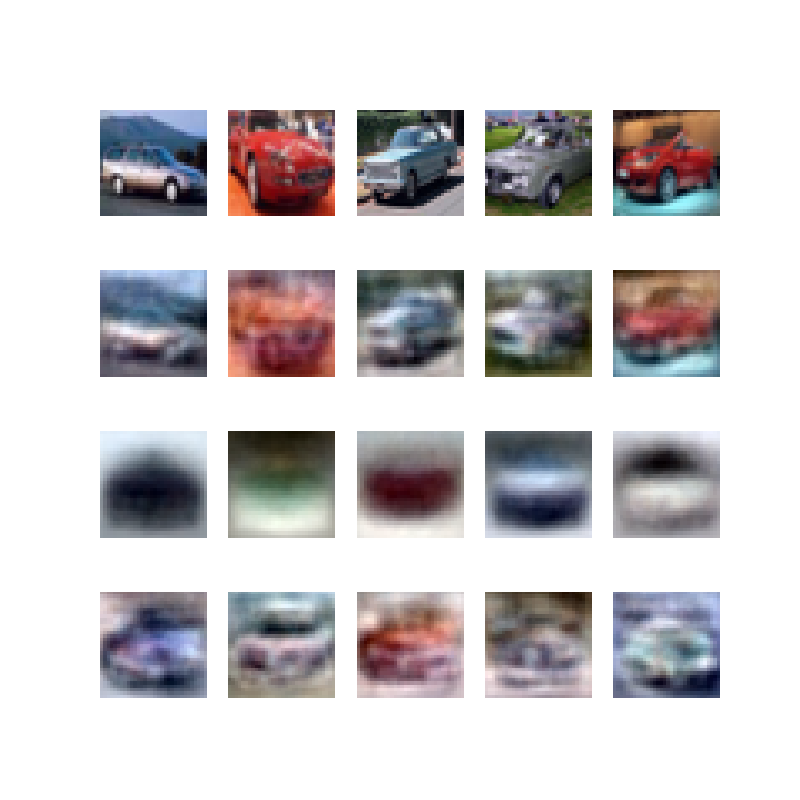}
    \caption{CIFAR10, class 1}
    \label{fig:CIFAR10_80_1}
    \vspace*{2mm}
\end{subfigure}\\
\begin{subfigure}{.4\textwidth}
    \includegraphics[width=1.0\textwidth]{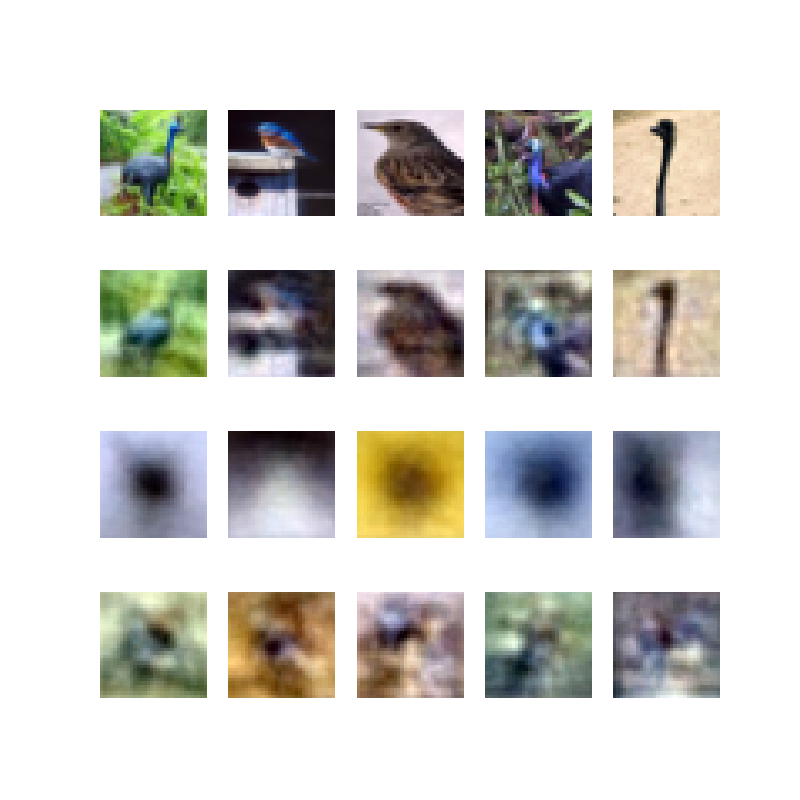}
    \caption{CIFAR10, class 2}
    \label{fig:CIFAR10_80_2}
    \vspace*{2mm}
\end{subfigure}
\hfill
\begin{subfigure}{.4\textwidth}
    \includegraphics[width=1.0\textwidth]{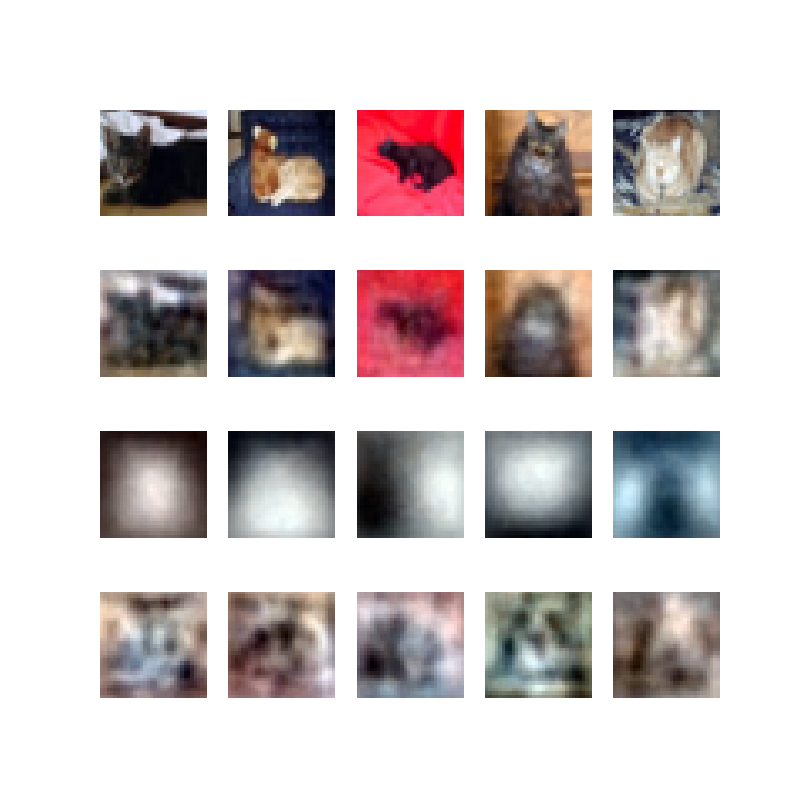}
    \caption{CIFAR10, class 3}
    \label{fig:CIFAR10_80_3}
    \vspace*{2mm}
\end{subfigure}\\
\begin{subfigure}{.4\textwidth}
    \includegraphics[width=1.0\textwidth]{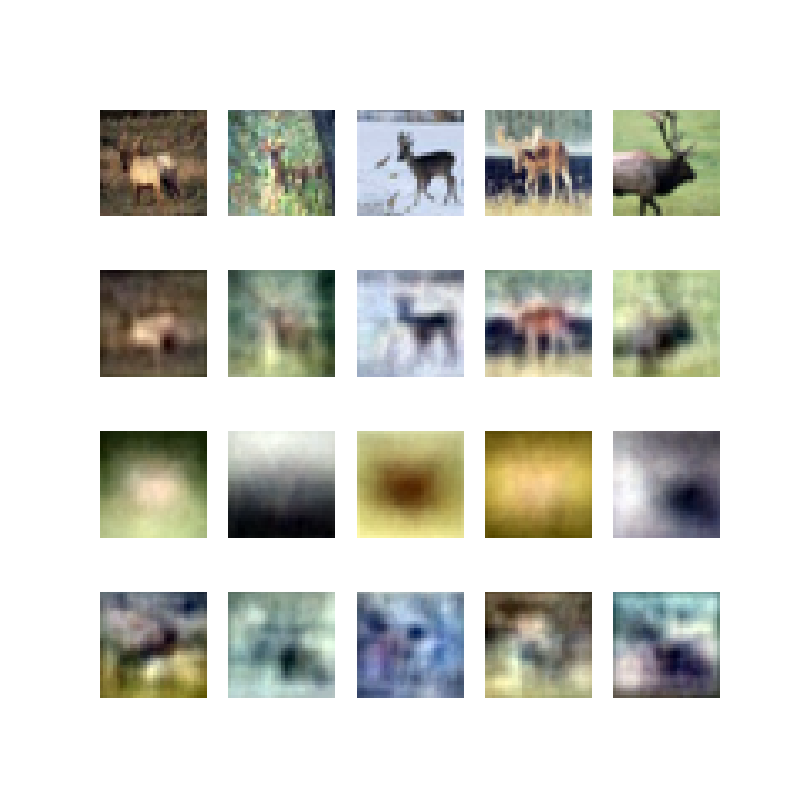}
    \caption{CIFAR10, class 4}
    \label{fig:CIFAR10_80_4}
    \vspace*{2mm}
\end{subfigure}
\caption{First row: Instances of one class. Second row: Recreation of instances using 80 components. Third row: The biggest 5 components, from big to small. Fourth row: Images synthesized by the lightweight generator, using 80 components.}
\label{fig:CIFAR10_80_0_4}
\end{figure*}

\begin{figure*}
\centering
\begin{subfigure}{.4\textwidth}
    \includegraphics[width=1.0\textwidth]{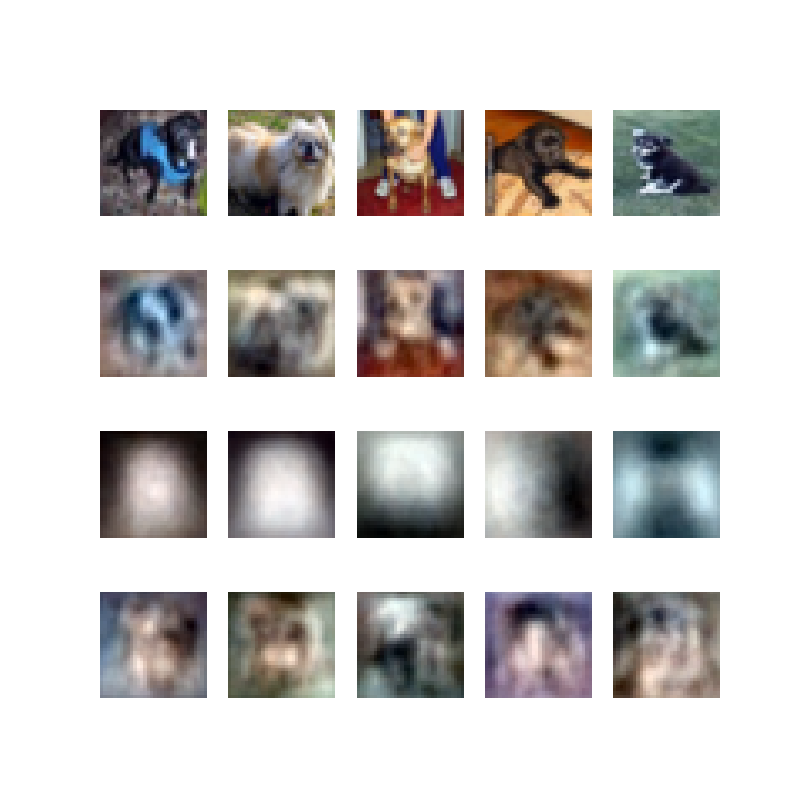}
    \caption{CIFAR10, class 5}
    \label{fig:CIFAR10_80_5}
    \vspace*{2mm}
\end{subfigure}
\hfill
\begin{subfigure}{.4\textwidth}
    \includegraphics[width=1.0\textwidth]{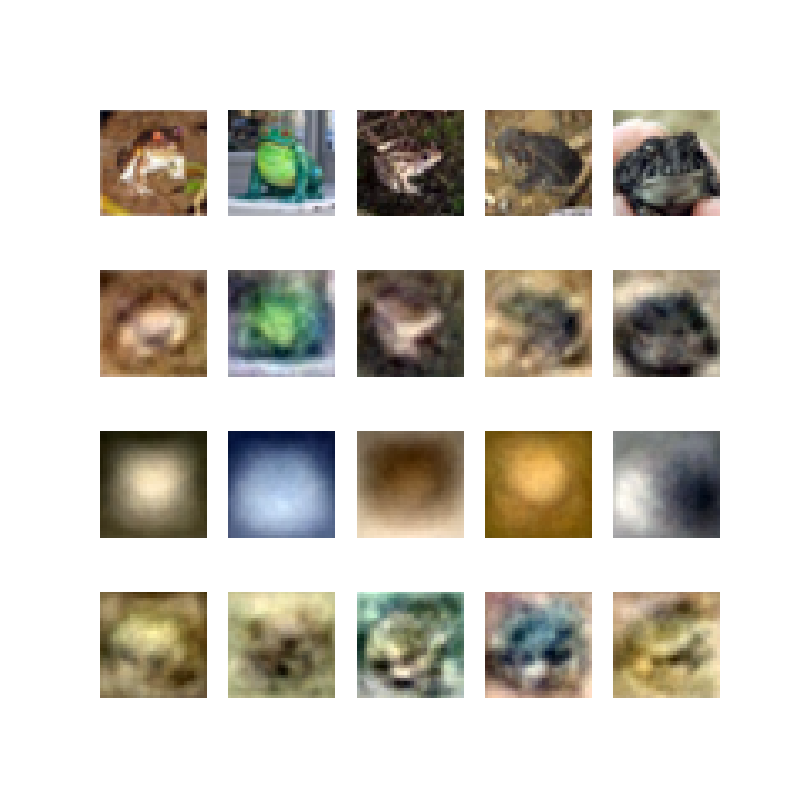}
    \caption{CIFAR10, class 6}
    \label{fig:CIFAR10_80_6}
    \vspace*{2mm}
\end{subfigure}\\
\begin{subfigure}{.4\textwidth}
    \includegraphics[width=1.0\textwidth]{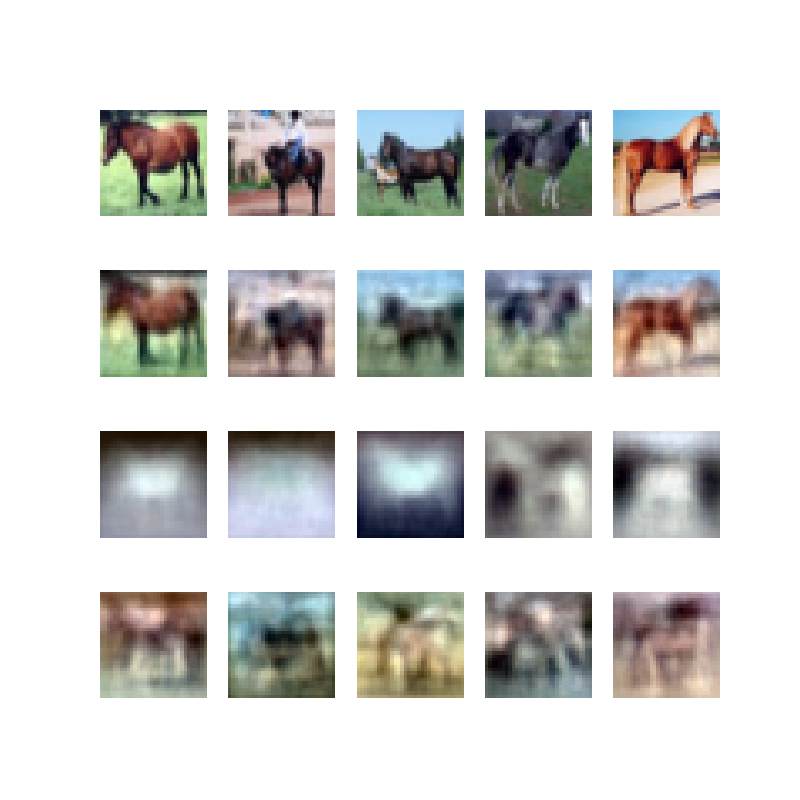}
    \caption{CIFAR10, class 7}
    \label{fig:CIFAR10_80_7}
    \vspace*{2mm}
\end{subfigure}
\hfill
\begin{subfigure}{.4\textwidth}
    \includegraphics[width=1.0\textwidth]{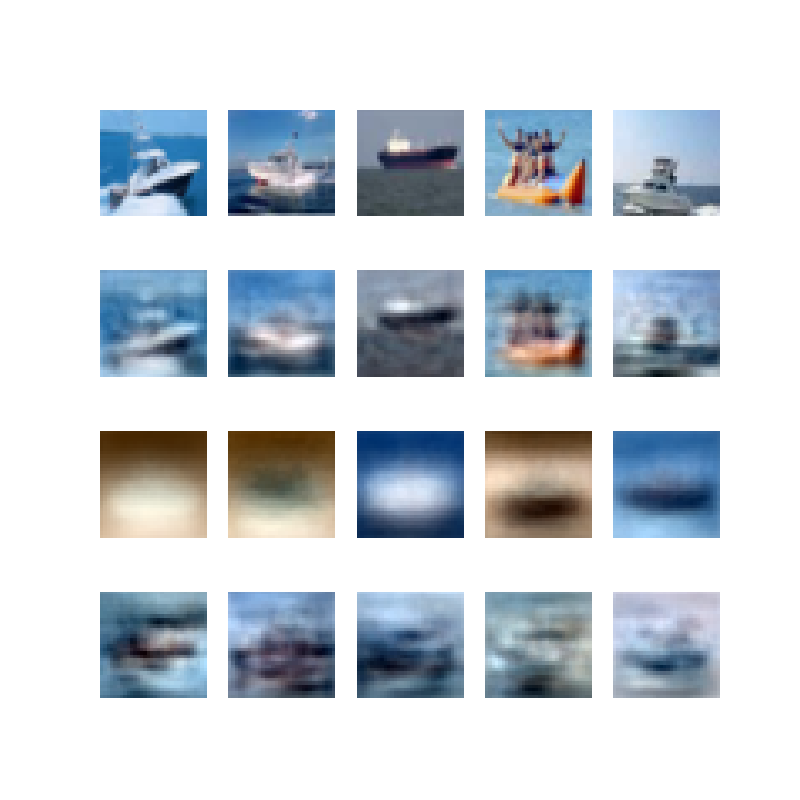}
    \caption{CIFAR10, class 8}
    \label{fig:CIFAR10_80_8}
    \vspace*{2mm}
\end{subfigure}\\
\begin{subfigure}{.4\textwidth}
    \includegraphics[width=1.0\textwidth]{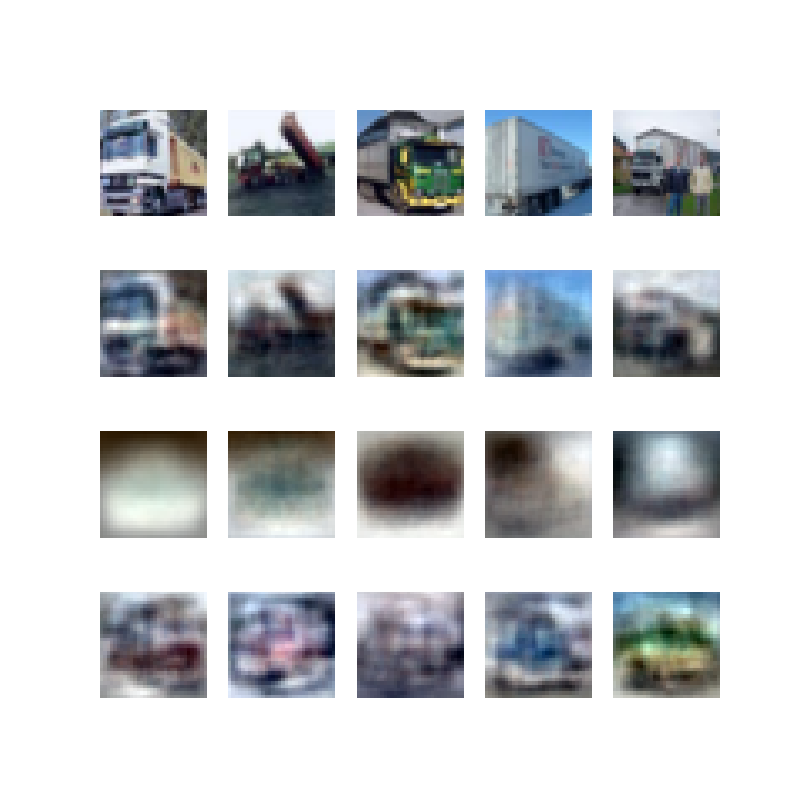}
    \caption{CIFAR10, class 9}
    \label{fig:CIFAR10_80_9}
    \vspace*{2mm}
\end{subfigure}
\caption{First row: Instances of one class. Second row: Recreation of instances using 80 components. Third row: The biggest 5 components, from big to small. Fourth row: Images synthesized by the lightweight generator, using 80 components.}
\label{fig:CIFAR10_80_5_9}
\end{figure*}

Figure \ref{fig:GEN0}, \ref{fig:GEN1}, \ref{fig:GEN2}, \ref{fig:GEN3}, \ref{fig:GEN4}, \ref{fig:GEN5}, \ref{fig:GEN6}, \ref{fig:GEN7}, \ref{fig:GEN8} and  \ref{fig:GEN9} give more insight in how the lightweight SVD generators work. They show instances of one class per dataset on row one, the recreation of these instances on row two, the biggest principle components from big to small on row three, and examples of generated samples on the fourth row. For figure \ref{fig:CIFAR10_80_0_4} and \ref{fig:CIFAR10_80_5_9}, the number of components is increased from five to $80$.

\section{Average Validation Accuracy}
\begin{figure*}
\centering
\begin{subfigure}{1.0\textwidth}
    \includegraphics[width=.45\textwidth]{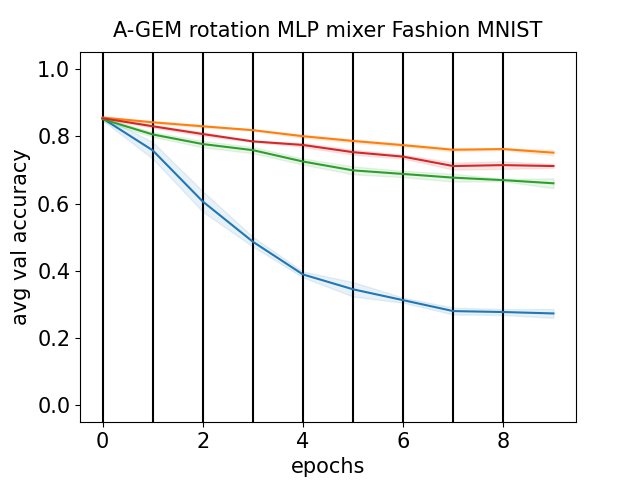}\hfill
    \includegraphics[width=.45\textwidth]{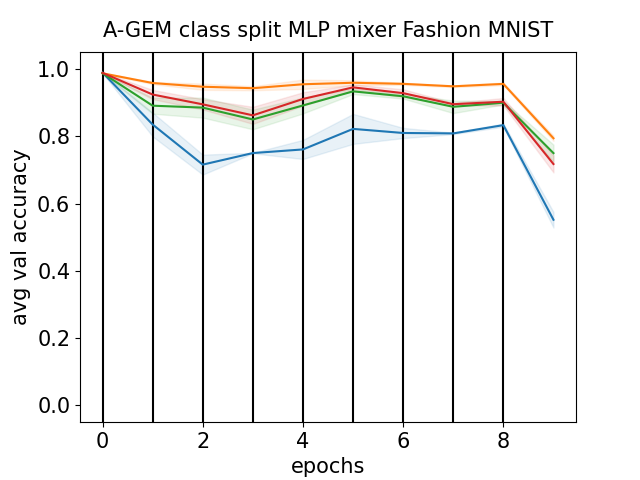}
    \caption{Average validation accuracy on Fashion MNIST}
    \label{fig:MLP_mixer_Fashion_MNIST}
    \vspace*{2mm}
\end{subfigure}\\
\begin{subfigure}{1.0\textwidth}
    \includegraphics[width=.45\textwidth]{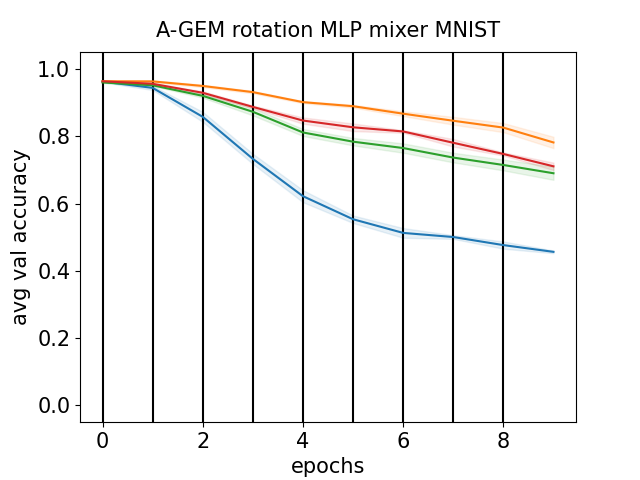}\hfill
    \includegraphics[width=.45\textwidth]{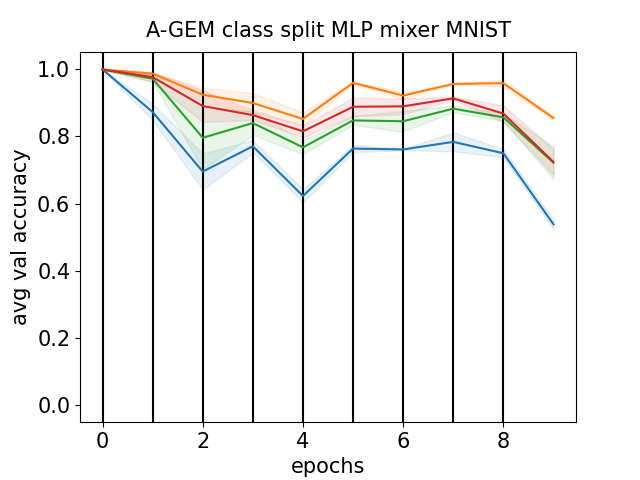}
    \caption{Average validation accuracy on MNIST}
    \label{fig:MLP_mixer_MNIST}
    \vspace*{2mm}
\end{subfigure}\\
\begin{subfigure}{1.0\textwidth}
    \includegraphics[width=.45\textwidth]{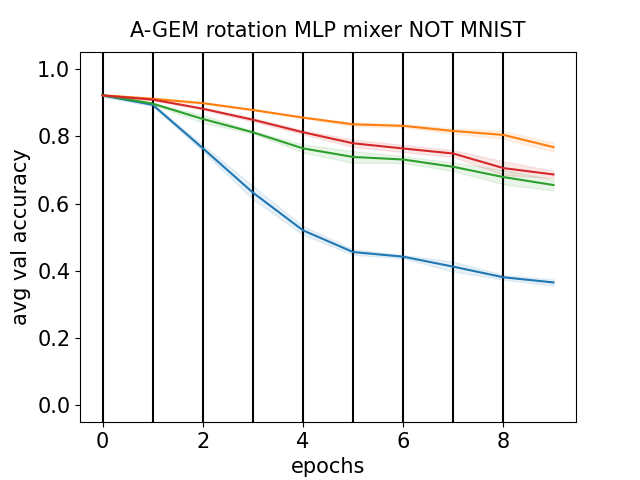}\hfill
    \includegraphics[width=.45\textwidth]{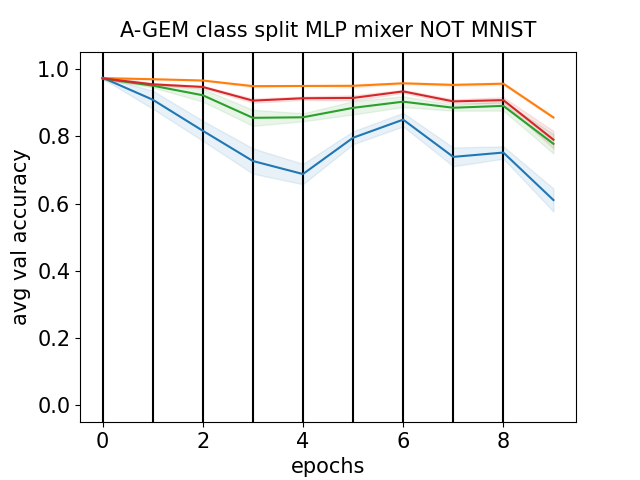}
    \caption{Average validation accuracy on NOT MNIST}
    \label{fig:MLP_mixer_NOT_MNIST}
    \vspace*{2mm}
\end{subfigure}\\
\begin{subfigure}{1.0\textwidth}
    \centering
    \includegraphics[width=1.0\textwidth]{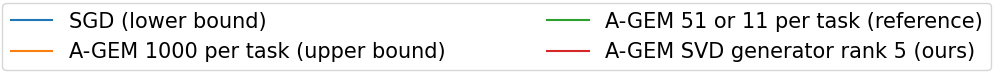}
    \label{fig:MLP_mixer_legend_1}
    \vspace*{2mm}
\end{subfigure}
\caption{These figures depict the average validation accuracy on all tasks trained on, using the \textbf{MLP mixer} architecture and \textbf{A-GEM} method. Subfigure a. shows the results for Fashion MNIST, subfigure b. those of MNIST and subfigure c. those of NOT MNIST. The left subfigures depict the results of a task creation by rotation, while the right subfigures depict the results of a task creating through class splitting. Each task is presented to the learner for 1 epoch. For a memory equivalent comparison of A-GEM and A-GEM with the SVD generator using 5 components, A-GEM uses 51 samples for the rotation experiments and 11 samples for the class split experiments. Each experiment was repeated five times. The average of these experiments is depicted as a line, surrounded by a shaded area of one standard deviation.}
\label{fig:MLP_mixer_1}
\end{figure*}

\begin{figure*}
\centering
\begin{subfigure}{1.0\textwidth}
    \includegraphics[width=.45\textwidth]{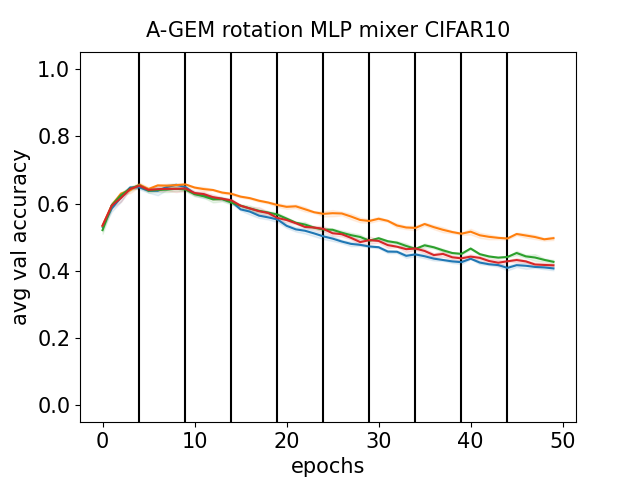}\hfill
    \includegraphics[width=.45\textwidth]{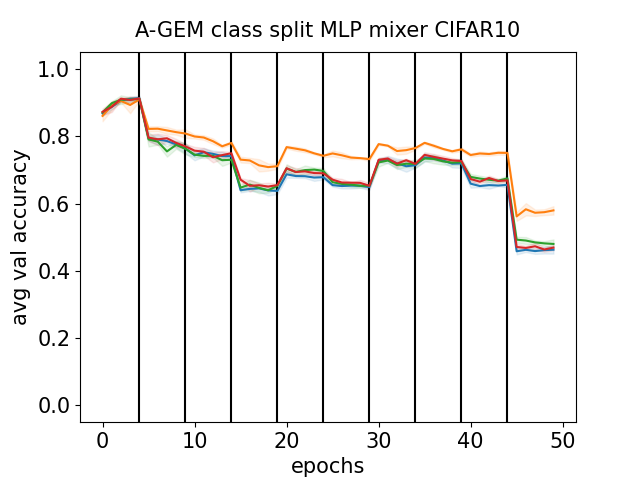}
    \caption{Average validation accuracy on CIFAR10}
    \label{fig:MLP_mixer_CIFAR10}
    \vspace*{2mm}
\end{subfigure}\\
\begin{subfigure}{1.0\textwidth}
    \includegraphics[width=.45\textwidth]{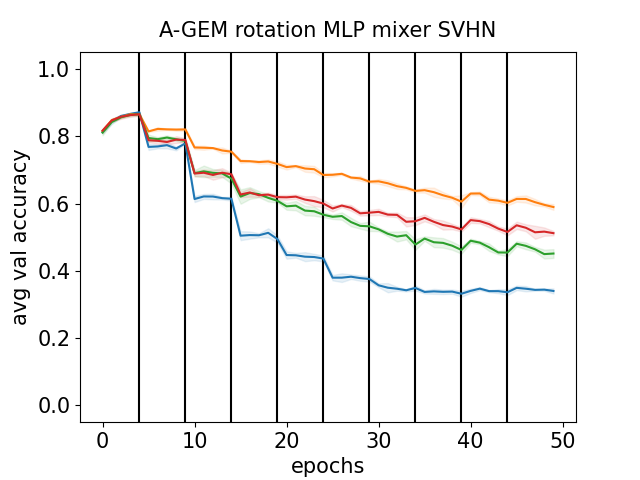}\hfill
    \includegraphics[width=.45\textwidth]{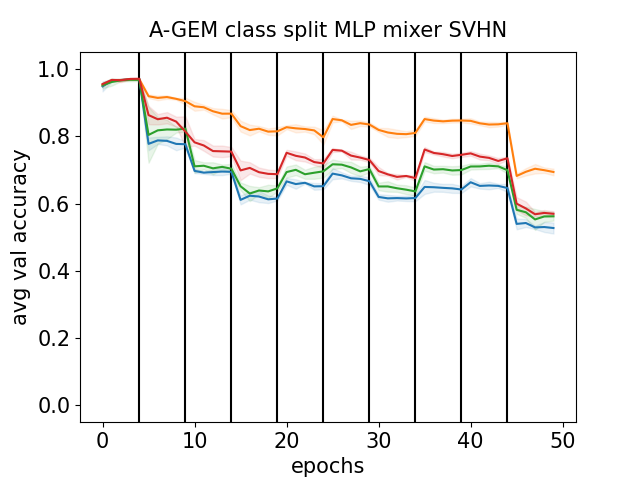}
    \caption{Average validation accuracy on SVHN}
    \label{fig:MLP_mixer_SVHN}
    \vspace*{2mm}
\end{subfigure}\\
\begin{subfigure}{1.0\textwidth}
    \centering
    \includegraphics[width=1.0\textwidth]{figures/MLP_mixer/legend_51_or_11.png}
    \label{fig:MLP_mixer_legend_2}
    \vspace*{2mm}
\end{subfigure}
\caption{These figures depict the average validation accuracy on all tasks trained on, using the \textbf{MLP mixer} architecture and \textbf{A-GEM} method. Subfigure a. shows the results for CIFAR10 and subfigure b. those of SVHN. The left subfigures depict the results of a task creation by rotation, while the right subfigures depict the results of a task creating through class splitting. Each task is presented to the learner for 5 epoch. For a memory equivalent comparison of A-GEM and A-GEM with the SVD generator using 5 components, A-GEM uses 51 samples for the rotation experiments and 11 samples for the class split experiments. Each experiment was repeated five times. The average of these experiments is depicted as a line, surrounded by a shaded area of one standard deviation.
}
\label{fig:MLP_mixer_2}
\end{figure*}

\begin{figure*}
\centering
\begin{subfigure}{1.0\textwidth}
    \includegraphics[width=.45\textwidth]{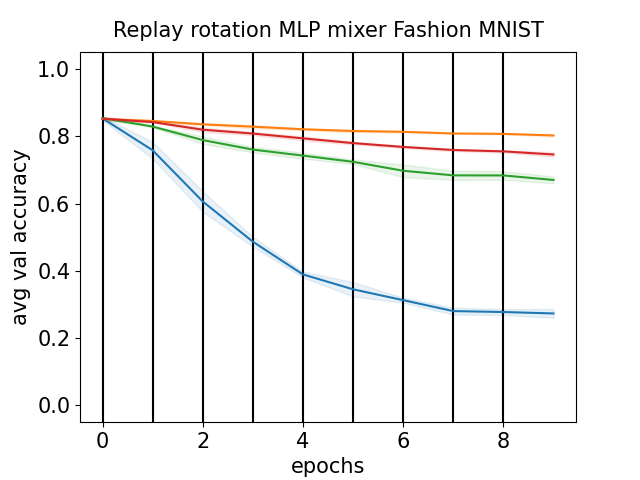}\hfill
    \includegraphics[width=.45\textwidth]{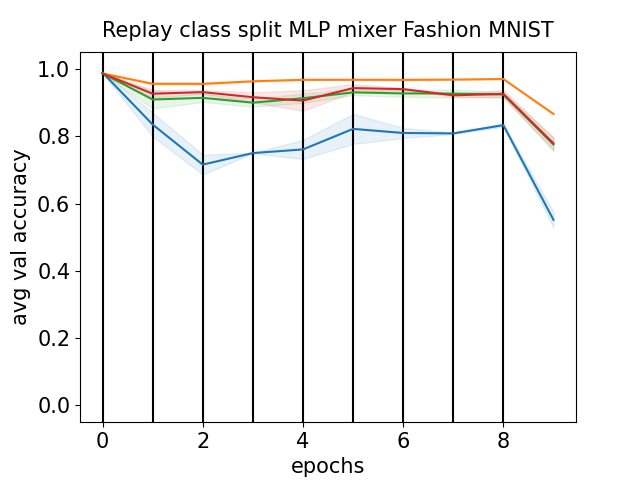}
    \caption{Average validation accuracy on Fashion MNIST}
    \label{fig:MLP_mixer_ER_Fashion_MNIST}
    \vspace*{2mm}
\end{subfigure}\\
\begin{subfigure}{1.0\textwidth}
    \includegraphics[width=.45\textwidth]{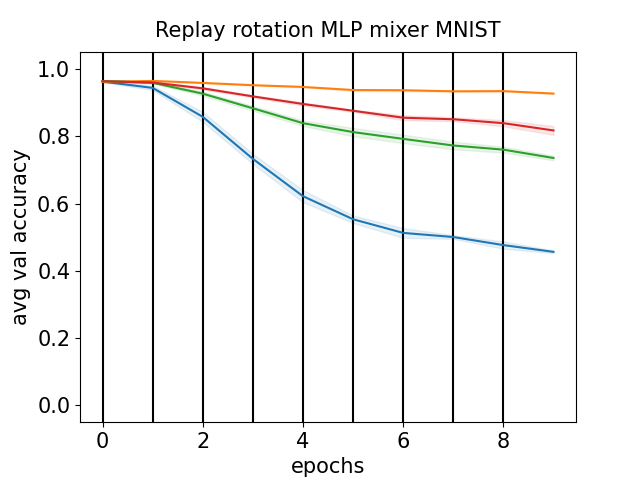}\hfill
    \includegraphics[width=.45\textwidth]{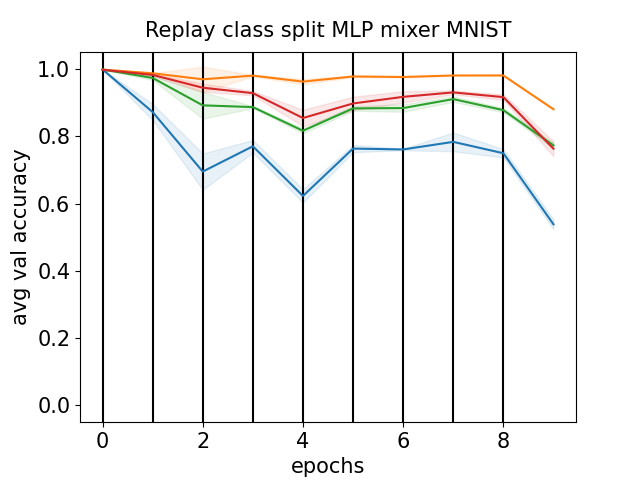}
    \caption{Average validation accuracy on MNIST}
    \label{fig:MLP_mixer_ER_MNIST}
    \vspace*{2mm}
\end{subfigure}\\
\begin{subfigure}{1.0\textwidth}
    \includegraphics[width=.45\textwidth]{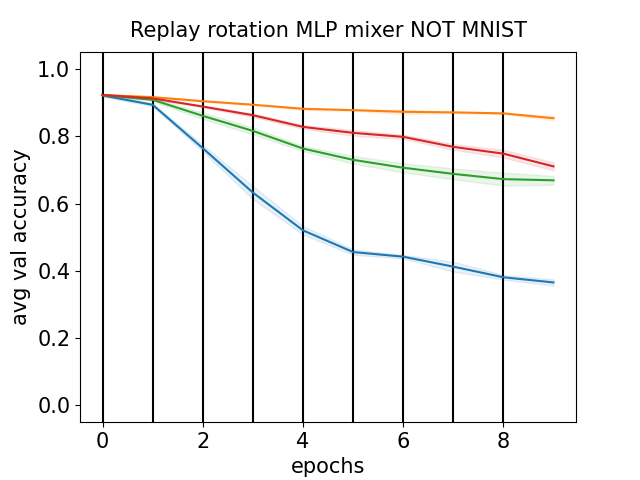}\hfill
    \includegraphics[width=.45\textwidth]{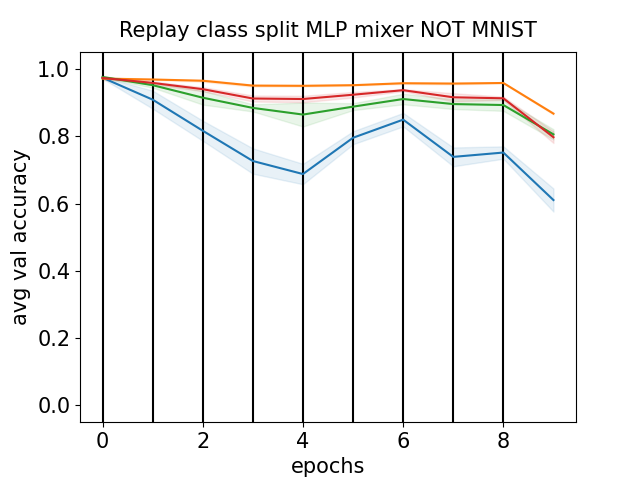}
    \caption{Average validation accuracy on NOT MNIST}
    \label{fig:MLP_mixer_ER_NOT_MNIST}
    \vspace*{2mm}
\end{subfigure}\\
\begin{subfigure}{1.0\textwidth}
    \centering
    \includegraphics[width=1.0\textwidth]{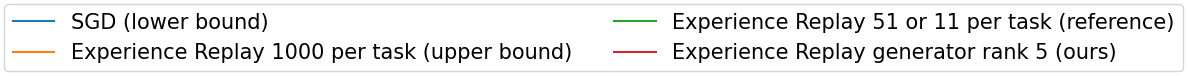}
    \label{fig:MLP_mixer_ER_legend_1}
    \vspace*{2mm}
\end{subfigure}
\caption{These figures depict the average validation accuracy on all tasks trained on, using the \textbf{MLP mixer} architecture and \textbf{ER} method. Subfigure a. shows the results for Fashion MNIST, subfigure b. those of MNIST and subfigure c. those of NOT MNIST. The left subfigures depict the results of a task creation by rotation, while the right subfigures depict the results of a task creating through class splitting. Each task is presented to the learner for 1 epoch. For a memory equivalent comparison of Experience Replay and Experience Replay with the SVD generator using 5 components, Experience Replay uses 51 samples for the rotation experiments and 11 samples for the class split experiments. Each experiment was repeated five times. The average of these experiments is depicted as a line, surrounded by a shaded area of one standard deviation.}
\label{fig:MLP_mixer_ER_1}
\end{figure*}

\begin{figure*}
\centering
\begin{subfigure}{1.0\textwidth}
    \includegraphics[width=.45\textwidth]{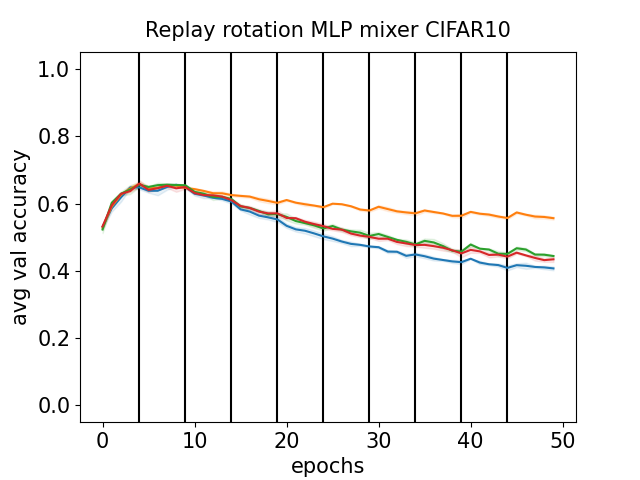}\hfill
    \includegraphics[width=.45\textwidth]{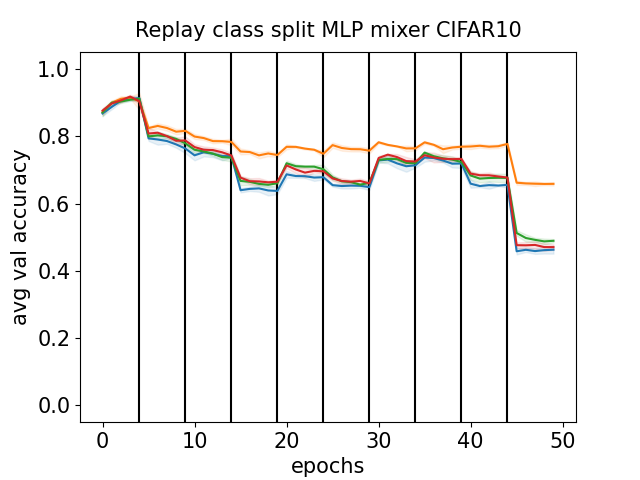}
    \caption{Average validation accuracy on CIFAR10}
    \label{fig:MLP_mixer_ER_CIFAR10}
    \vspace*{2mm}
\end{subfigure}\\
\begin{subfigure}{1.0\textwidth}
    \includegraphics[width=.45\textwidth]{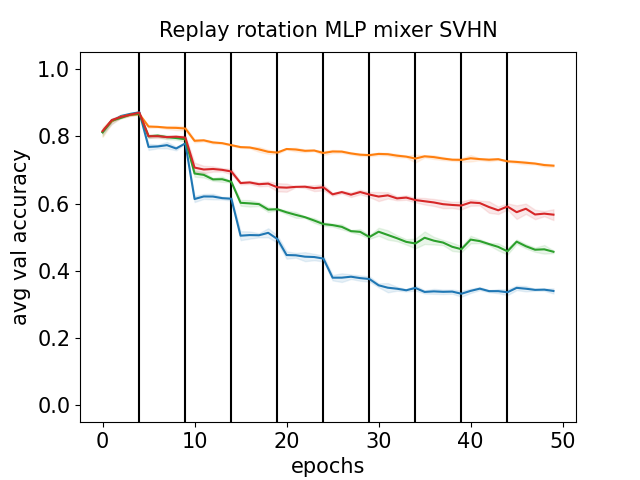}\hfill
    \includegraphics[width=.45\textwidth]{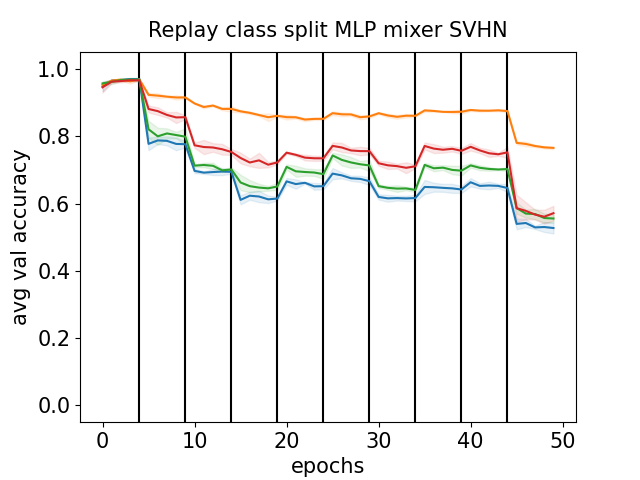}
    \caption{Average validation accuracy on SVHN}
    \label{fig:MLP_mixer_ER_SVHN}
    \vspace*{2mm}
\end{subfigure}\\
\begin{subfigure}{1.0\textwidth}
    \centering
    \includegraphics[width=1.0\textwidth]{figures/MLP_mixer/legend_51_or_11_replay.png}
    \label{fig:MLP_mixer_ER_legend_2}
    \vspace*{2mm}
\end{subfigure}
\caption{These figures depict the average validation accuracy on all tasks trained on, using the \textbf{MLP mixer} architecture and \textbf{ER} method. Subfigure a. shows the results for CIFAR10 and subfigure b. those of SVHN. The left subfigures depict the results of a task creation by rotation, while the right subfigures depict the results of a task creating through class splitting. Each task is presented to the learner for 5 epoch. For a memory equivalent comparison of Experience Replay and Experience Replay with the SVD generator using 5 components, Experience Replay uses 51 samples for the rotation experiments and 11 samples for the class split experiments. Each experiment was repeated five times. The average of these experiments is depicted as a line, surrounded by a shaded area of one standard deviation.
}
\label{fig:MLP_mixer_ER_2}
\end{figure*}

\begin{figure*}
\centering
\begin{subfigure}{1.0\textwidth}
    \includegraphics[width=.45\textwidth]{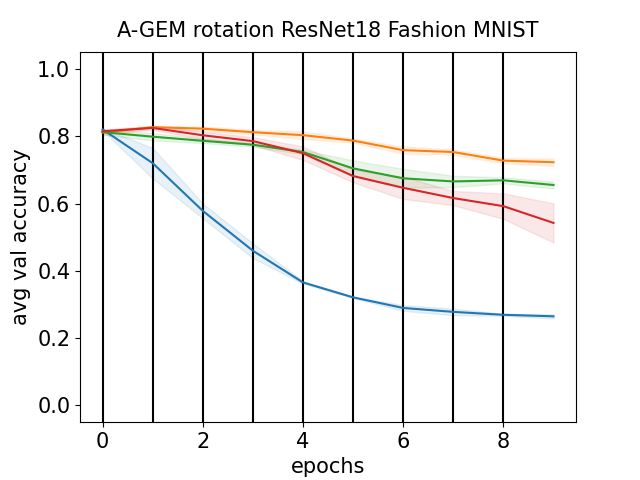}\hfill
    \includegraphics[width=.45\textwidth]{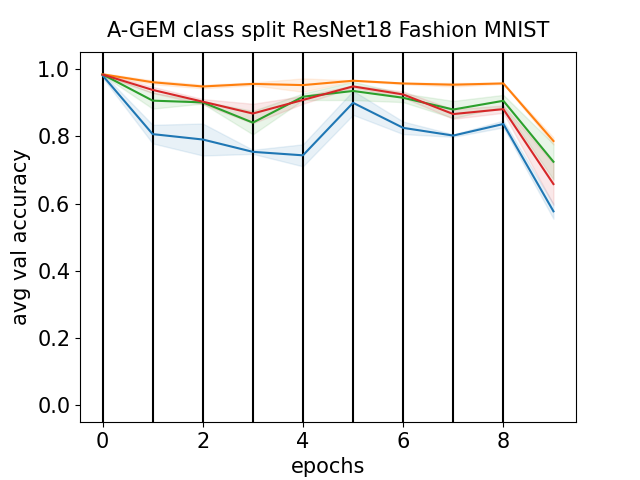}
    \caption{Average validation accuracy on Fashion MNIST}
    \label{fig:ResNet18_Fashion_MNIST}
    \vspace*{2mm}
\end{subfigure}\\
\begin{subfigure}{1.0\textwidth}
    \includegraphics[width=.45\textwidth]{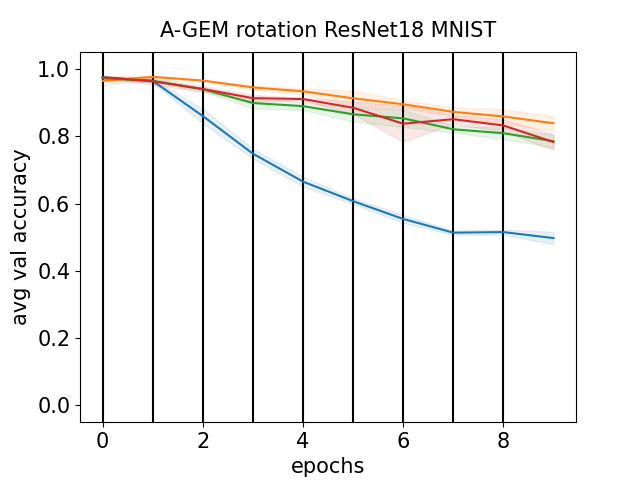}\hfill
    \includegraphics[width=.45\textwidth]{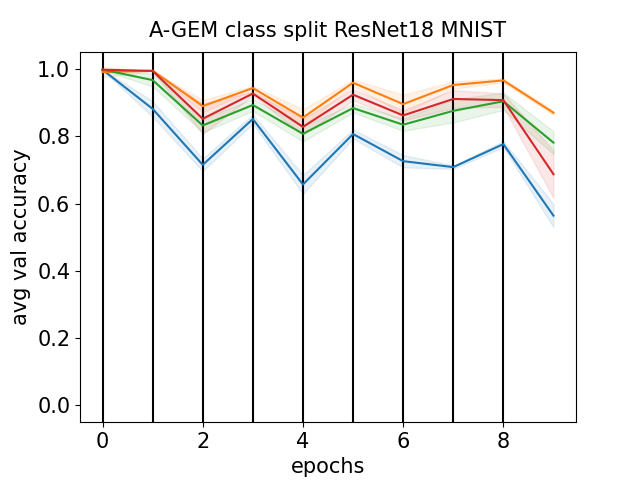}
    \caption{Average validation accuracy on MNIST}
    \label{fig:ResNet18_MNIST}
    \vspace*{2mm}
\end{subfigure}\\
\begin{subfigure}{1.0\textwidth}
    \includegraphics[width=.45\textwidth]{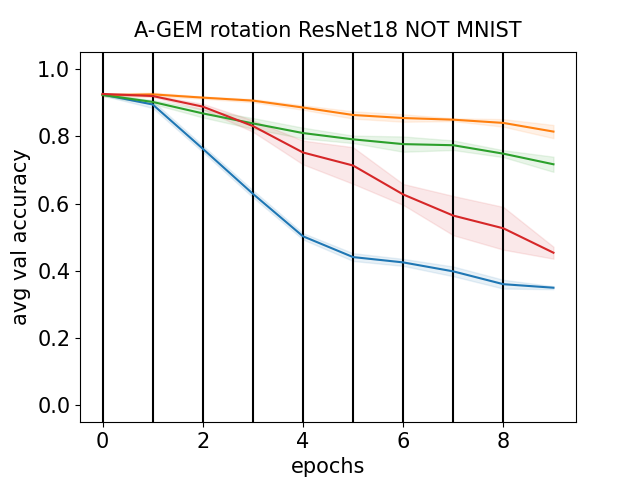}\hfill
    \includegraphics[width=.45\textwidth]{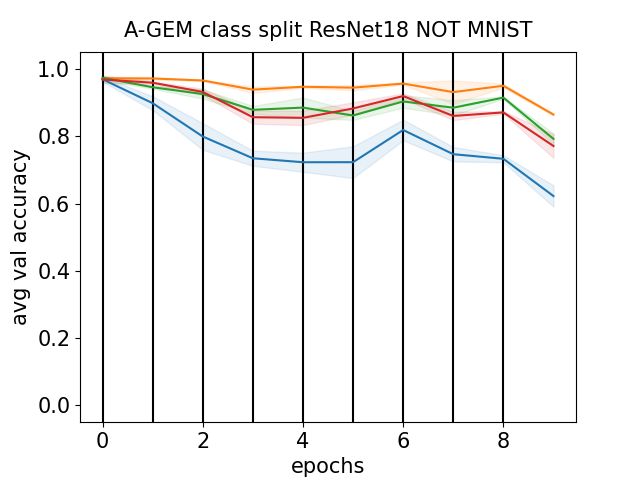}
    \caption{Average validation accuracy on NOT MNIST}
    \label{fig:ResNet18_NOT_MNIST}
    \vspace*{2mm}
\end{subfigure}\\
\begin{subfigure}{1.0\textwidth}
    \centering
    \includegraphics[width=1.0\textwidth]{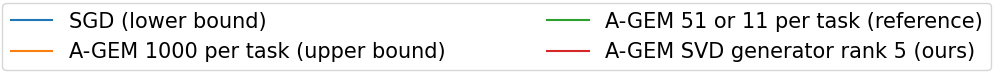}
    \label{fig:ResNet18_legend_1}
    \vspace*{2mm}
\end{subfigure}
\caption{These figures depict the average validation accuracy on all tasks trained on, using the \textbf{ResNet18} architecture and \textbf{A-GEM} method. Subfigure a. shows the results for Fashion MNIST, subfigure b. those of MNIST and subfigure c. those of NOT MNIST. The left subfigures depict the results of a task creation by rotation, while the right subfigures depict the results of a task creating through class splitting. Each task is presented to the learner for 1 epoch. For a memory equivalent comparison of A-GEM and A-GEM with the SVD generator using 5 components, A-GEM uses 51 samples for the rotation experiments and 11 samples for the class split experiments. Each experiment was repeated five times. The average of these experiments is depicted as a line, surrounded by a shaded area of one standard deviation.}
\label{fig:ResNet18_1}
\end{figure*}

\begin{figure*}
\centering
\begin{subfigure}{1.0\textwidth}
    \includegraphics[width=.45\textwidth]{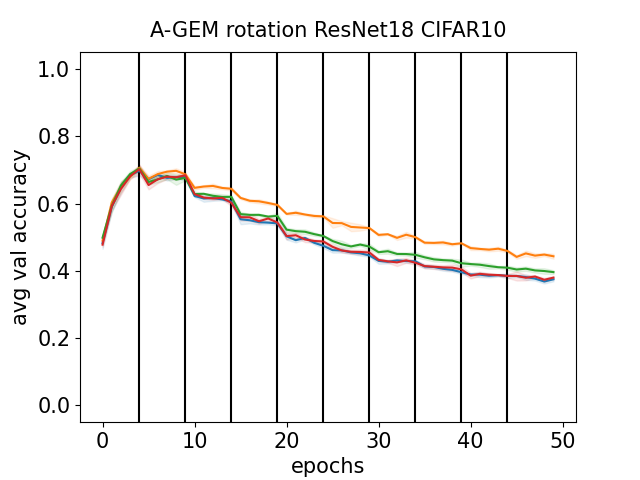}\hfill
    \includegraphics[width=.45\textwidth]{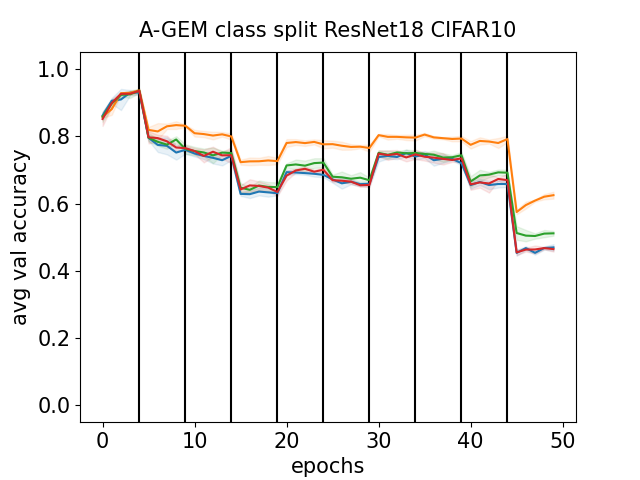}
    \caption{Average validation accuracy on CIFAR10}
    \label{fig:ResNet18_CIFAR10}
    \vspace*{2mm}
\end{subfigure}\\
\begin{subfigure}{1.0\textwidth}
    \includegraphics[width=.45\textwidth]{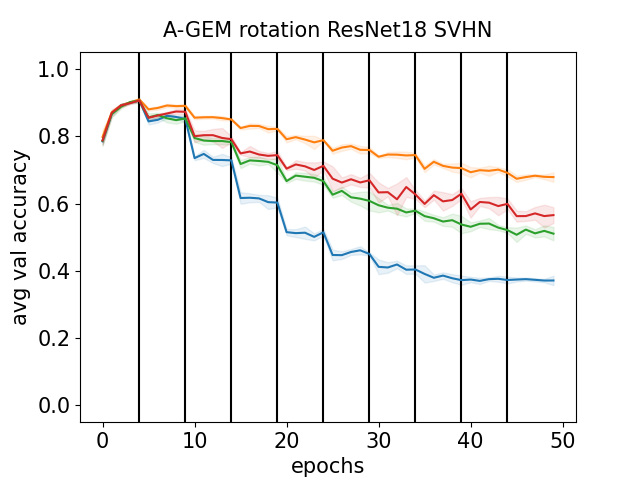}\hfill
    \includegraphics[width=.45\textwidth]{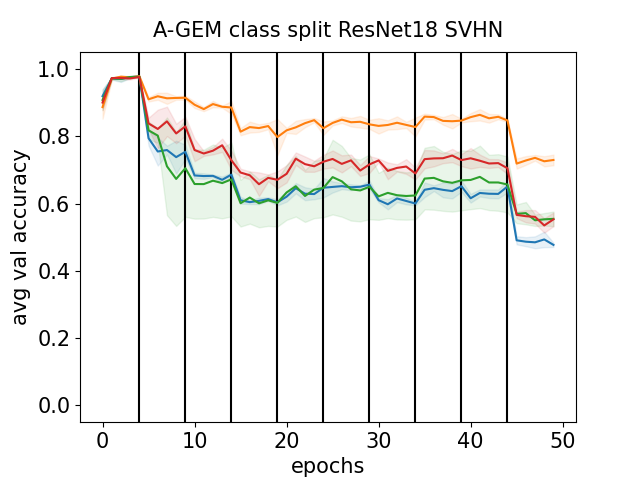}
    \caption{Average validation accuracy on SVHN}
    \label{fig:ResNet18_SVHN}
    \vspace*{2mm}
\end{subfigure}\\
\begin{subfigure}{1.0\textwidth}
    \centering
    \includegraphics[width=1.0\textwidth]{figures/ResNet18/legend_51_or_11.png}
    \label{fig:ResNet18_legend_2}
    \vspace*{2mm}
\end{subfigure}
\caption{These figures depict the average validation accuracy on all tasks trained on, using the \textbf{ResNet18} architecture and \textbf{A-GEM} method. Subfigure a. shows the results for CIFAR10 and subfigure b. those of SVHN. The left subfigures depict the results of a task creation by rotation, while the right subfigures depict the results of a task creating through class splitting. Each task is presented to the learner for 5 epoch. For a memory equivalent comparison of A-GEM and A-GEM with the SVD generator using 5 components, A-GEM uses 51 samples for the rotation experiments and 11 samples for the class split experiments. Each experiment was repeated five times. The average of these experiments is depicted as a line, surrounded by a shaded area of one standard deviation.}
\label{fig:ResNet18_2}
\end{figure*}

\begin{figure*}
\centering
\begin{subfigure}{1.0\textwidth}
    \includegraphics[width=.45\textwidth]{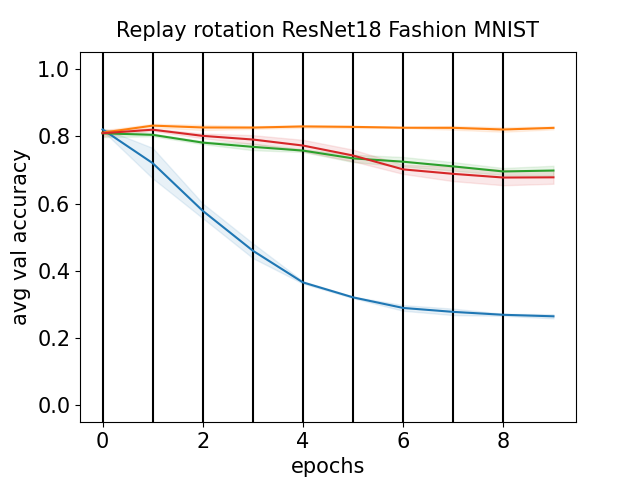}\hfill
    \includegraphics[width=.45\textwidth]{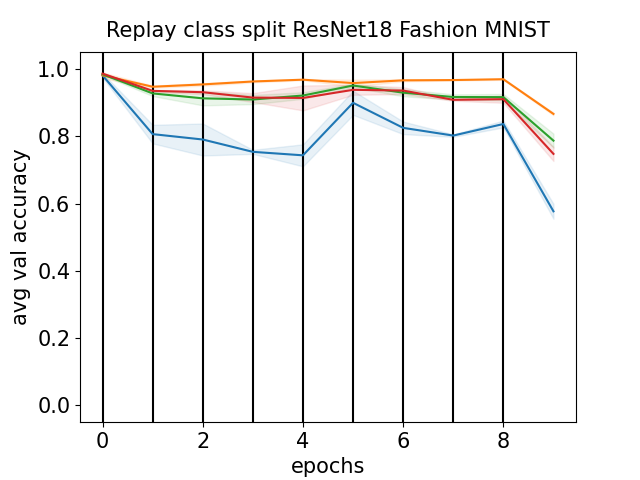}
    \caption{Average validation accuracy on Fashion MNIST}
    \label{fig:ResNet18_ER_Fashion_MNIST}
    \vspace*{2mm}
\end{subfigure}\\
\begin{subfigure}{1.0\textwidth}
    \includegraphics[width=.45\textwidth]{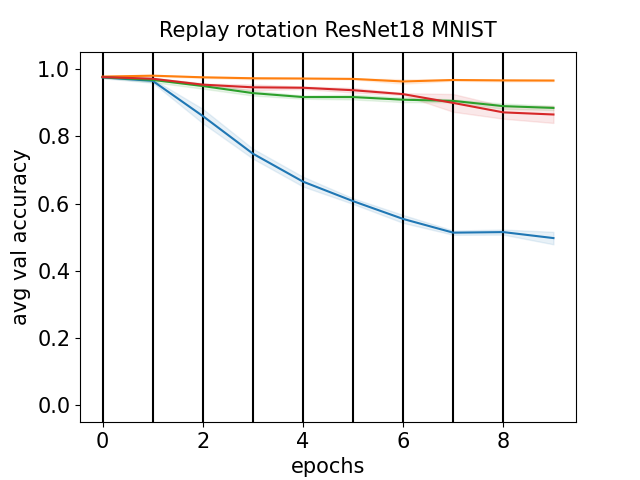}\hfill
    \includegraphics[width=.45\textwidth]{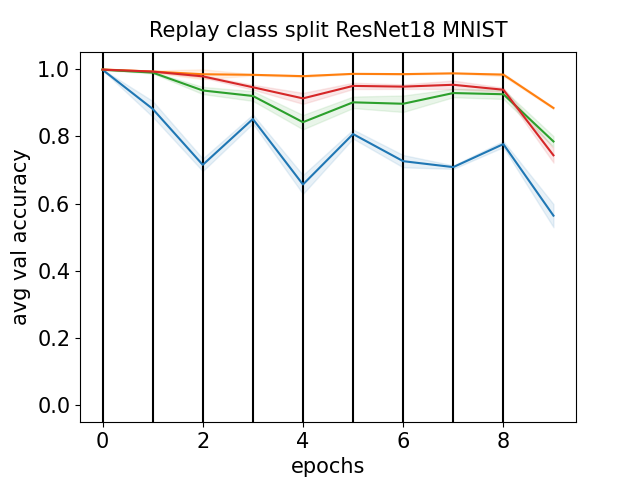}
    \caption{Average validation accuracy on MNIST}
    \label{fig:ResNet18_ER_MNIST}
    \vspace*{2mm}
\end{subfigure}\\
\begin{subfigure}{1.0\textwidth}
    \includegraphics[width=.45\textwidth]{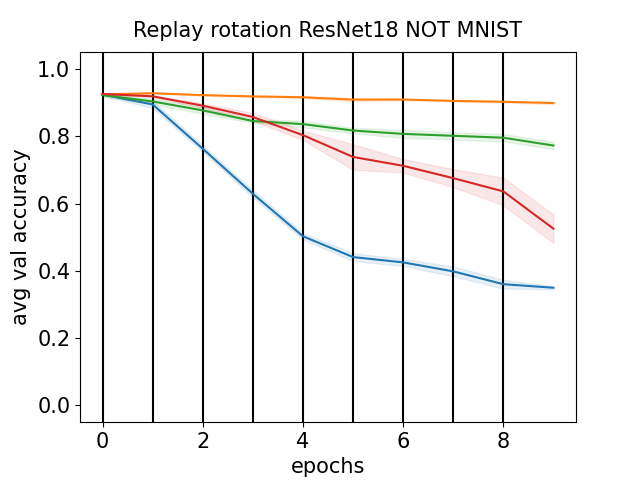}\hfill
    \includegraphics[width=.45\textwidth]{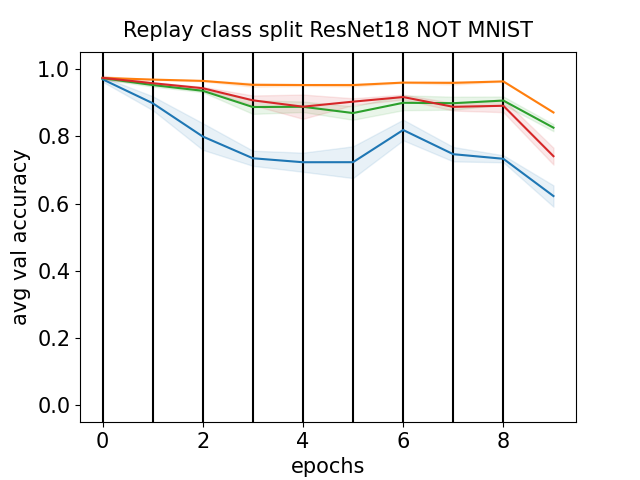}
    \caption{Average validation accuracy on NOT MNIST}
    \label{fig:ResNet18_ER_NOT_MNIST}
    \vspace*{2mm}
\end{subfigure}\\
\begin{subfigure}{1.0\textwidth}
    \centering
    \includegraphics[width=1.0\textwidth]{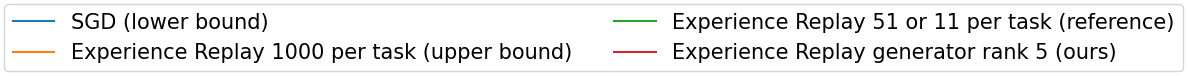}
    \label{fig:ResNet18_ER_legend_1}
    \vspace*{2mm}
\end{subfigure}
\caption{These figures depict the average validation accuracy on all tasks trained on, using the \textbf{ResNet18} architecture and \textbf{ER} method. Subfigure a. shows the results for Fashion MNIST, subfigure b. those of MNIST and subfigure c. those of NOT MNIST. The left subfigures depict the results of a task creation by rotation, while the right subfigures depict the results of a task creating through class splitting. Each task is presented to the learner for 1 epoch. For a memory equivalent comparison of Experience Replay and Experience Replay with the SVD generator using 5 components, Experience Replay uses 51 samples for the rotation experiments and 11 samples for the class split experiments. Each experiment was repeated five times. The average of these experiments is depicted as a line, surrounded by a shaded area of one standard deviation.}
\label{fig:ResNet18_ER_1}
\end{figure*}

\begin{figure*}
\centering
\begin{subfigure}{1.0\textwidth}
    \includegraphics[width=.45\textwidth]{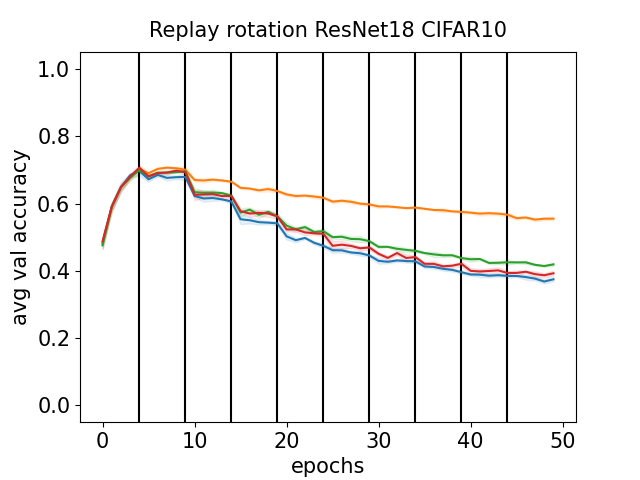}\hfill
    \includegraphics[width=.45\textwidth]{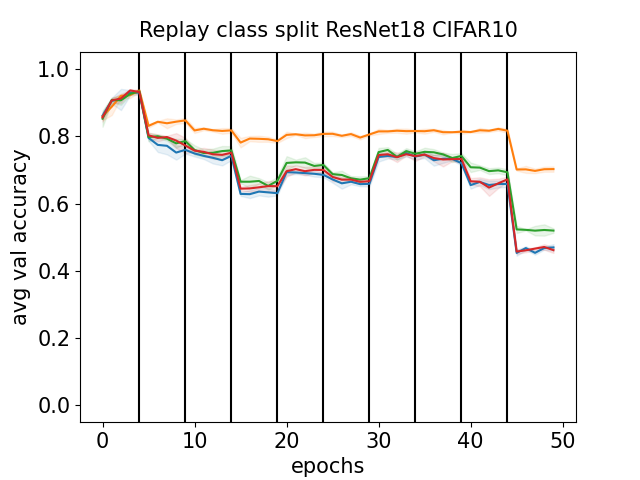}
    \caption{Average validation accuracy on CIFAR10}
    \label{fig:ResNet18_ER_CIFAR10}
    \vspace*{2mm}
\end{subfigure}\\
\begin{subfigure}{1.0\textwidth}
    \includegraphics[width=.45\textwidth]{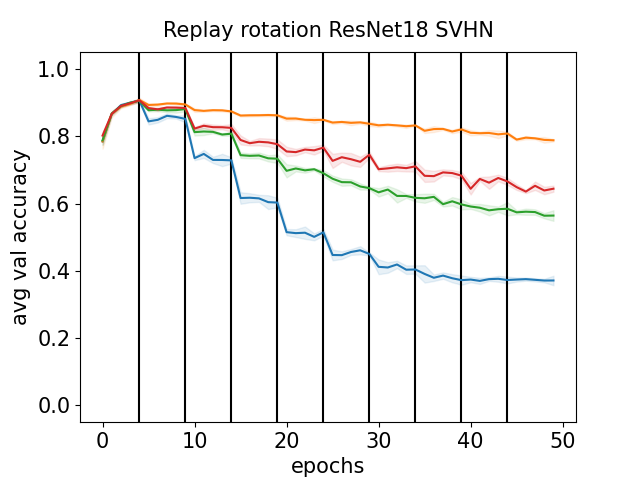}\hfill
    \includegraphics[width=.45\textwidth]{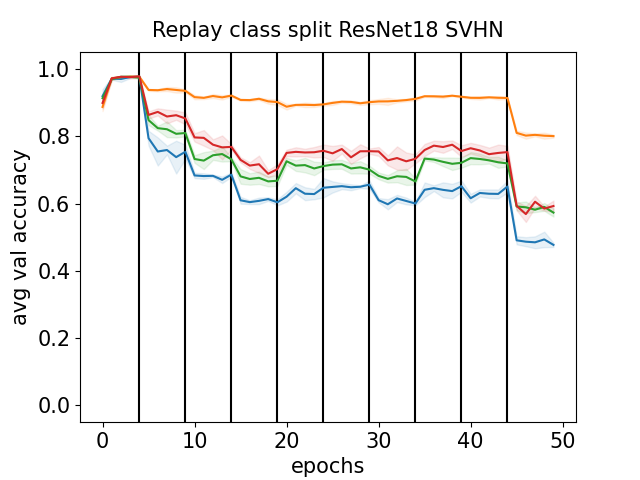}
    \caption{Average validation accuracy on SVHN}
    \label{fig:ResNet18_ER_SVHN}
    \vspace*{2mm}
\end{subfigure}\\
\begin{subfigure}{1.0\textwidth}
    \centering
    \includegraphics[width=1.0\textwidth]{figures/ResNet18/legend_51_or_11_replay.png}
    \label{fig:ResNet18_ER_legend_2}
    \vspace*{2mm}
\end{subfigure}
\caption{These figures depict the average validation accuracy on all tasks trained on, using the \textbf{ResNet18} architecture and \textbf{ER} method. Subfigure a. shows the results for CIFAR10 and subfigure b. those of SVHN. The left subfigures depict the results of a task creation by rotation, while the right subfigures depict the results of a task creating through class splitting. Each task is presented to the learner for 5 epoch. For a memory equivalent comparison of Experience Replay and Experience Replay with the SVD generator using 5 components, Experience Replay uses 51 samples for the rotation experiments and 11 samples for the class split experiments. Each experiment was repeated five times. The average of these experiments is depicted as a line, surrounded by a shaded area of one standard deviation.}
\label{fig:ResNet18_ER_2}
\end{figure*}

\begin{figure*}[!ht]
\centering
\begin{subfigure}{1.0\textwidth}
    \includegraphics[width=0.5\textwidth]{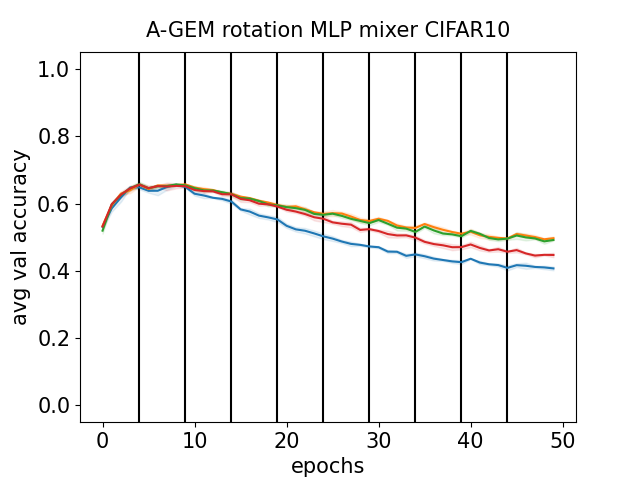}
    \includegraphics[width=0.5\textwidth]{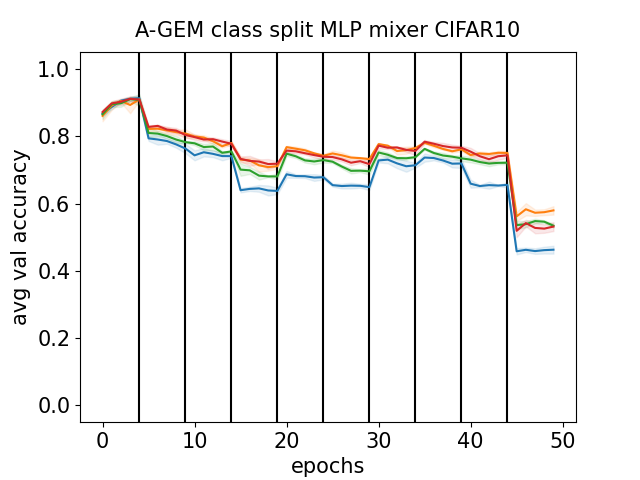}
    \caption{Average validation accuracy on CIFAR10, using A-GEM}
    \label{fig:MLP_mixer_CIFAR10_rank_80_AGEM}
    \vspace*{5mm}
\end{subfigure}\\
\begin{subfigure}{1.0\textwidth}
    \centering
    \includegraphics[width=1.0\textwidth]{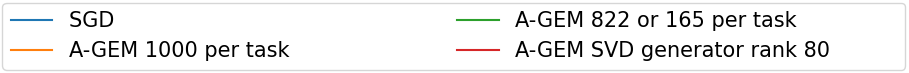}\hfill
    \label{fig:MLP_mixer_CIFAR10_rank_80_legend_1}
    \vspace*{4mm}
\end{subfigure}
\begin{subfigure}{0.95\textwidth}
    \includegraphics[width=0.5\textwidth]{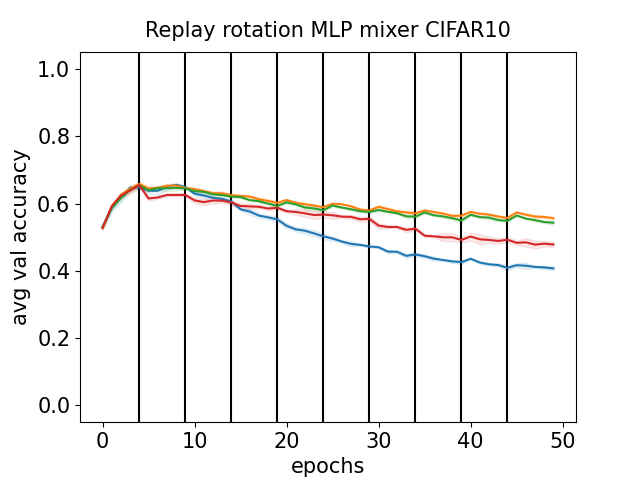}
    \includegraphics[width=0.5\textwidth]{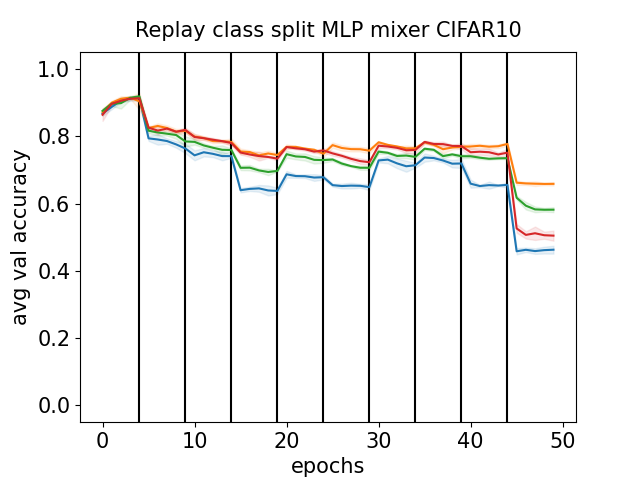}
    \caption{Average validation accuracy on CIFAR10, using Experience Replay (ER)}
    \label{fig:MLP_mixer_CIFAR10_rank_80_RE}
    \vspace*{5mm}
\end{subfigure}\\
\begin{subfigure}{1.0\textwidth}
    \centering
    \includegraphics[width=1.0\textwidth]{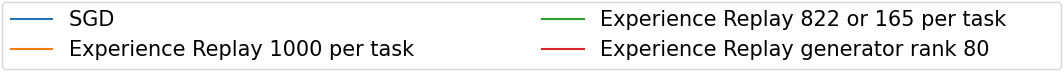}\hfill
    \label{fig:MLP_mixer_CIFAR10_rank_80_legend_2}
    \vspace*{4mm}
\end{subfigure}
\caption{These figures depict the average validation accuracy on all tasks trained on, using the \textbf{MLP mixer} architecture and CIFAR10. Subfigure a. shows the results when using \textbf{A-GEM} and subfigure b. those when using the \textbf{ER} method. The left subfigures depict the results of a task creation by rotation, while the right subfigures depict the results of a task creating through class splitting. Each task is presented to the learner for 5 epoch. For a memory equivalent comparison of raw sampling vs. the  SVD generator using 80 components, raw sampling uses 822 samples for the rotation experiments and 165 samples for the class split experiments. Each experiment was repeated five times. The average of these experiments is depicted as a line, surrounded by a shaded area of one standard deviation.}
\label{fig:MLP_mixer_rank_80}
\vspace*{4mm}
\end{figure*}

\begin{figure*}[!ht]
\centering
\begin{subfigure}{1.0\textwidth}
    \includegraphics[width=0.5\textwidth]{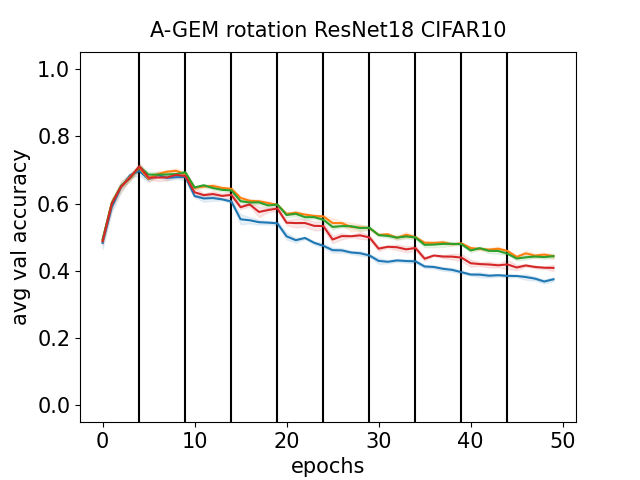}
    \includegraphics[width=0.5\textwidth]{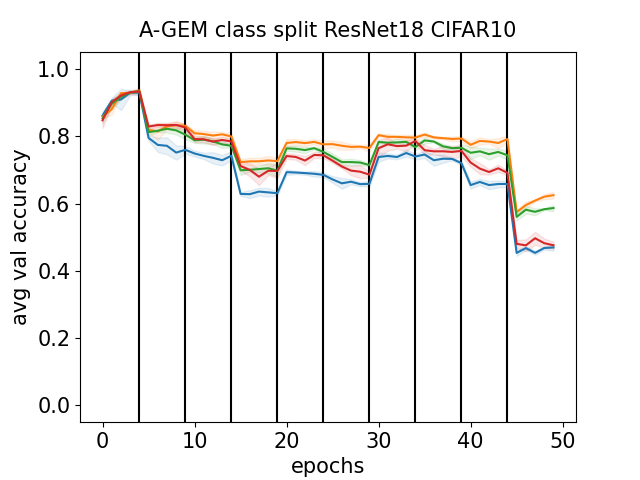}
    \caption{Average validation accuracy on CIFAR10, using A-GEM}
    \label{fig:ResNet18_CIFAR10_rank_80_AGEM}
    \vspace*{5mm}
\end{subfigure}\\
\begin{subfigure}{1.0\textwidth}
    \centering
    \includegraphics[width=1.0\textwidth]{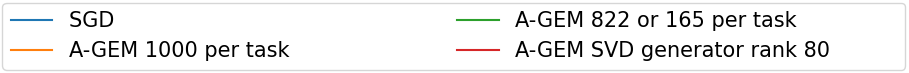}\hfill
    \label{fig:ResNet18_CIFAR10_rank_80_legend_1}
    \vspace*{4mm}
\end{subfigure}
\begin{subfigure}{0.95\textwidth}
    \includegraphics[width=0.5\textwidth]{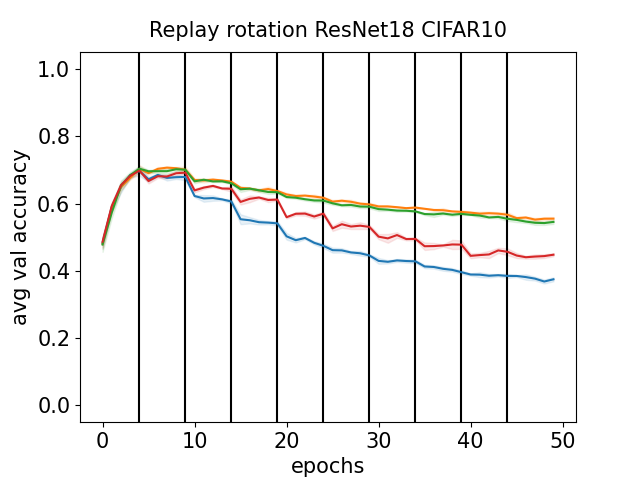}
    \includegraphics[width=0.5\textwidth]{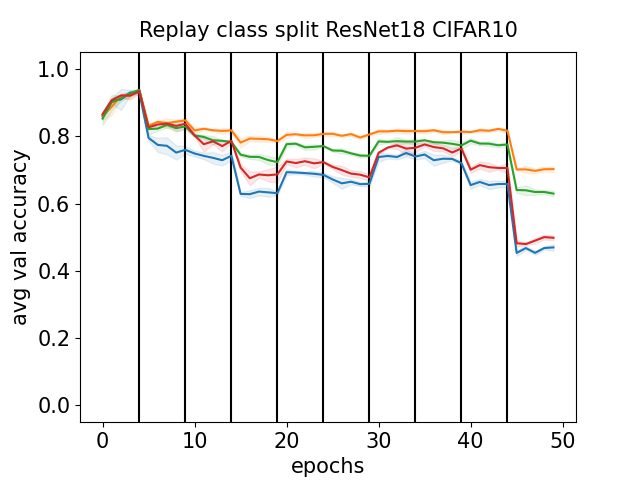}
    \caption{Average validation accuracy on CIFAR10, using Experience Replay (ER)}
    \label{fig:ResNet18_CIFAR10_rank_80_RE}
    \vspace*{5mm}
\end{subfigure}\\
\begin{subfigure}{1.0\textwidth}
    \centering
    \includegraphics[width=1.0\textwidth]{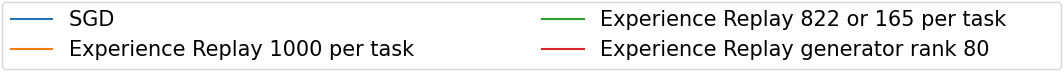}\hfill
    \label{fig:ResNet18_CIFAR10_rank_80_legend_2}
    \vspace*{4mm}
\end{subfigure}
\caption{These figures depict the average validation accuracy on all tasks trained on, using the \textbf{ResNet18} architecture and CIFAR10. Subfigure a. shows the results when using \textbf{A-GEM} and subfigure b. those when using the \textbf{ER} method. The left subfigures depict the results of a task creation by rotation, while the right subfigures depict the results of a task creating through class splitting. Each task is presented to the learner for 5 epoch. For a memory equivalent comparison of raw sampling vs. the  SVD generator using 80 components, raw sampling uses 822 samples for the rotation experiments and 165 samples for the class split experiments. Each experiment was repeated five times. The average of these experiments is depicted as a line, surrounded by a shaded area of one standard deviation.}
\label{fig:ResNet18_rank_80}
\vspace*{4mm}
\end{figure*}

Figure \ref{fig:MLP_mixer_1}, \ref{fig:MLP_mixer_2}, \ref{fig:MLP_mixer_ER_1}, \ref{fig:MLP_mixer_ER_2}, \ref{fig:ResNet18_1}, \ref{fig:ResNet18_2}, \ref{fig:ResNet18_ER_1} and \ref{fig:ResNet18_ER_2}  show the average validation accuracy on the MLP mixer and ResNet18 architectures. Figure \ref{fig:MLP_mixer_rank_80} and \ref{fig:ResNet18_rank_80} show the results on CIFAR10, when the number of generator components is increased from five to $80$.

%
%


\end{document}